%% file: main.tex
\documentclass[10pt,journal,compsoc]{IEEEtran}
\usepackage{colortbl}
\usepackage[utf8]{inputenc} % allow utf-8 input
\usepackage[T1]{fontenc}    % use 8-bit T1 fonts
\usepackage{url}            % simple URL typesetting
\usepackage{booktabs}       % professional-quality tables
\usepackage{amsfonts}       % blackboard math symbols
\usepackage{nicefrac}       % compact symbols for 1/2, etc.
\usepackage{microtype}      % microtypography
\usepackage[dvipsnames, svgnames, x11names,table]{xcolor}

\definecolor{darkamber}{RGB}{195, 95, 0}
\usepackage[pagebackref=true,breaklinks=true,colorlinks,citecolor=ForestGreen]{hyperref}
\usepackage{graphicx}
\usepackage{tabularx}
\usepackage{amsmath}
\usepackage{subfigure}
\usepackage{tcolorbox}
\usepackage{multirow}
\usepackage{pifont}
\usepackage{enumitem}
\usepackage{amssymb}

\usepackage{enumitem}
\usepackage[misc]{ifsym}
\usepackage{bm}
\usepackage{multirow}
\usepackage{bbding}
\usepackage{array}
\usepackage{hyperref}
\definecolor{LightGray}{gray}{0.95}
\usepackage{float}
\usepackage{wrapfig}
\usepackage{overpic}
\usepackage{stfloats}

\ifCLASSOPTIONcompsoc
  \usepackage[nocompress]{cite}
\else
  \usepackage{cite}
\fi

\ifCLASSINFOpdf
\else
\fi

\def\eg{\emph{e.g.}}
\def\ie{\emph{i.e.}}
\def\etc{\emph{etc}}
\def\vs{\emph{v.s.}}

\def\etal{\emph{et al.}}

\newtheorem{definition}{Definition}

\def\eg{\emph{e.g.}}
\def\vs{\emph{vs.}}
\def\ie{\emph{i.e.}}
\def\etc{\emph{etc}}

\def\etal{\emph{et al.}}

\usepackage{booktabs}
\usepackage{multirow}
\usepackage{tabularx}
\usepackage{wasysym}
\usepackage{colortbl}
\usepackage{xcolor}
\usepackage{hhline}
\usepackage{graphicx}
\usepackage{hyperref}
\usepackage{xcolor}
\definecolor{deepgreen}{RGB}{0,100,0}

\usepackage{ragged2e}
\usepackage[edges]{forest}
\usetikzlibrary{shadows.blur}
\usetikzlibrary{shapes.geometric}
\definecolor{hiddendraw}{RGB}{10,128,122}
\definecolor{hidden-orange}{RGB}{224,224,224}
\usepackage{subcaption}
\usepackage{svg}
\usepackage{hyperref}

\usepackage{pifont}
\usepackage[perpage,symbol*]{footmisc}
\DefineFNsymbols{circled}{{\ding{192}}{\ding{193}}{\ding{194}}
{\ding{195}}{\ding{196}}{\ding{197}}{\ding{198}}{\ding{199}}{\ding{200}}{\ding{201}}}
\setfnsymbol{circled}

\newcommand\myfootnotestyle[1]{\ifcase#1 \or \ding{182}\or \ding{183}\or
\ding{184}\or \ding{185}\or \ding{186}\or \ding{187}%
\or \ding{188}\or \ding{189}\or \ding{190}\or \ding{191}\else *\fi\relax}

\hypersetup{hidelinks,colorlinks=true}

\newif\ifsubmit
\submitfalse

\begin{document}
\title{Object Detectors in the Open Environment: Challenges, Solutions, and Outlook}%A Survey on Open Environment Object Detectors: Challenges, Solutions, and Outlook} %Robustness,\\ Elasticity, Generalization}

\author{%
  Siyuan Liang\footnote[1]{}, Wei Wang, 
  Ruoyu Chen, Aishan Liu, Boxi Wu, Ee-Chien Chang, \\ Xiaochun Cao,~\IEEEmembership{Senior Member,~IEEE}, Dacheng Tao,~\IEEEmembership{Fellow,~IEEE}
% \thanks{S. Liang is with the School of Computing, National University of Singapore, Singapore, 117417 (Email: \href{mailto:pandaliang521@gmail.com}{pandaliang521@gmail.com}).}
% \thanks{W. Wang and X. Cao are with the School of Cyber Science and Technology, Shenzhen Campus of Sun Yat-sen University, Shenzhen, 518107, China (Email: \href{mailto:wangwei29@mail.sysu.edu.cn}{wangwei29@mail.sysu.edu.cn}, \href{mailto:caoxiaochun@mail.sysu.edu.cn}{caoxiaochun@mail.sysu.edu.cn}).}
% \thanks{R. Chen is with the Institute of Information Engineering, Chinese Academy of Sciences, Beijing 100093, China, and also with the School of Cyber Security, University of Chinese Academy of Sciences, Beijing 100049, China (Email: \href{mailto:chenruoyu@iie.ac.cn}{chenruoyu@iie.ac.cn}).}
\thanks{S. Liang and E.-C. Chang are with the School of Computing, National University of Singapore, Singapore 117417.}
\thanks{W. Wang and X. Cao (corresponding author) are with the School of Cyber Science and Technology, Shenzhen Campus of Sun Yat-sen University, Shenzhen 518107, China.}
\thanks{R. Chen is with the Institute of Information Engineering, Chinese Academy of Sciences, Beijing 100093, China, and also with the School of Cyber Security, University of Chinese Academy of Sciences, Beijing 100049, China.}
\thanks{A. Liu is with the SCSE, Beihang University, Beijing 100191, China.}
\thanks{B. Wu is with the School of Software, Zhejiang University, Hangzhou 310027, China.}
\thanks{D. Tao is with the School of Computer Science and Engineering at Nanyang Technological University, \#32 Block N4 \#02a-014, 50 Nanyang Avenue, Singapore 639798.}
\thanks{(Author W. Wang and R. Chen contributed equally to this work.)}
}

\markboth{Submitted to IEEE TRANSACTIONS ON PATTERN ANALYSIS AND MACHINE INTELLIGENCE}%
{Shell \MakeLowercase{\textit{\etal}}: Bare Demo of IEEEtran.cls for Computer Society Journals}



\IEEEtitleabstractindextext{%
\begin{abstract}
\justifying
With the emergence of foundation models, deep learning-based object detectors have shown practical usability in closed set scenarios. However, for real-world tasks, object detectors often operate in open environments, where crucial factors (\eg, data distribution, objective) that influence model learning are often changing. The dynamic and intricate nature of the open environment poses novel and formidable challenges to object detectors. Unfortunately, current research on object detectors in open environments lacks a comprehensive analysis of their distinctive characteristics, challenges, and corresponding solutions, which hinders their secure deployment in critical real-world scenarios. This paper aims to bridge this gap by conducting a comprehensive review and analysis of object detectors in open environments. We initially identified limitations of key structural components within the existing detection pipeline and propose the open environment object detector challenge framework that includes four quadrants (\ie, out-of-domain, out-of-category, robust learning, and incremental learning) based on the dimensions of the data / target changes. For each quadrant of challenges in the proposed framework, we present a detailed description and systematic analysis of the overarching goals and core difficulties, systematically review the corresponding solutions, and benchmark their performance over multiple widely adopted datasets. In addition, we engage in a discussion of open problems and potential avenues for future research. This paper aims to provide a fresh, comprehensive, and systematic understanding of the challenges and solutions associated with open-environment object detectors, thus catalyzing the development of more solid applications in real-world scenarios. A project related to this survey can be found at \url{https://github.com/LiangSiyuan21/OEOD_Survey}.

\end{abstract}

\begin{IEEEkeywords}
Object Detection, Open Environments, Challenges and Solutions.%Discriminant Detection, Domain Adaptation, Incremental Learning, Adversarial Defense, Deep Learning.
\end{IEEEkeywords}}

\maketitle

\IEEEdisplaynontitleabstractindextext

% \fi
\IEEEpeerreviewmaketitle

\input{sections/1-introduction}
\input{sections/2-background}
\input{sections/3-object_detection_towards_open_environments}
\input{sections/4-out-of-domain}

\input{sections/5-out-of-category}

\input{sections/6-malicious_data}

\input{sections/7-incremental}

\input{sections/10-benchmark}

\input{sections/8-future_outlook}
\input{sections/9-conclusion}
\ifCLASSOPTIONcaptionsoff
  \newpage
\fi

{
\bibliography{reference}
\bibliographystyle{unsrt}
}
%\input{checklist}

%\newpage

% \input{bios}

\newpage
\newpage
\clearpage

\appendices
\input{suppl_sections/1-related_survey}

\input{suppl_sections/4-out-of-domain}
\input{suppl_sections/5-out-of-category}

\input{suppl_sections/6-malicious-data}
\input{suppl_sections/7-incremental}

\input{suppl_sections/8-benchmark}

\end{document}

%% file: sections/1-introduction.tex
\section{Introduction}
As one of the foundational tasks in computer vision, object detection identifies instances of visual objects by predicting their locations and categories within a given image. With the evolution of deep neural networks, especially the emergence of foundation models, deep object detection~\cite{zou2019object,zhao2019object,szegedy2013deep} has garnered extensive attention and has served as fundamental blocks in various application areas, including autonomous driving~\cite{yurtsever2020survey}, healthcare~\cite{esteva2019guide}, face detection~\cite{tang2004video, liu2006spatio}, \etc.

In particular, the efficacy of deep object detectors predominantly relies on the \emph{closed set} assumption with finite, known, and well-defined structures~\cite{zhou2022open}. For example, the data used for training and testing satisfy the same distribution, and the learning process is optimized toward invariant target tasks (label classification and location regression on benign examples). Driven by increasingly challenging tasks, the landscape for practical deep detector application scenarios gradually evolved from closed to open environments~\cite{zhou2022open}. In the \emph{open environment}, critical factors of the model learning process (\eg, data distribution, and target tasks) can undergo dynamic shifts over time and in the surroundings. For example, an open environment may introduce novel categories during testing that the model has not encountered during training, even with noises that can disrupt the model decision. The open environment setting closely mirrors real-world situations, demanding deep detectors exhibit heightened robustness and scalability to navigate unforeseen changes and challenges effectively.
\begin{figure}[]
\vspace{-0.1in}
  \centering
  \includegraphics[width=0.45\textwidth]{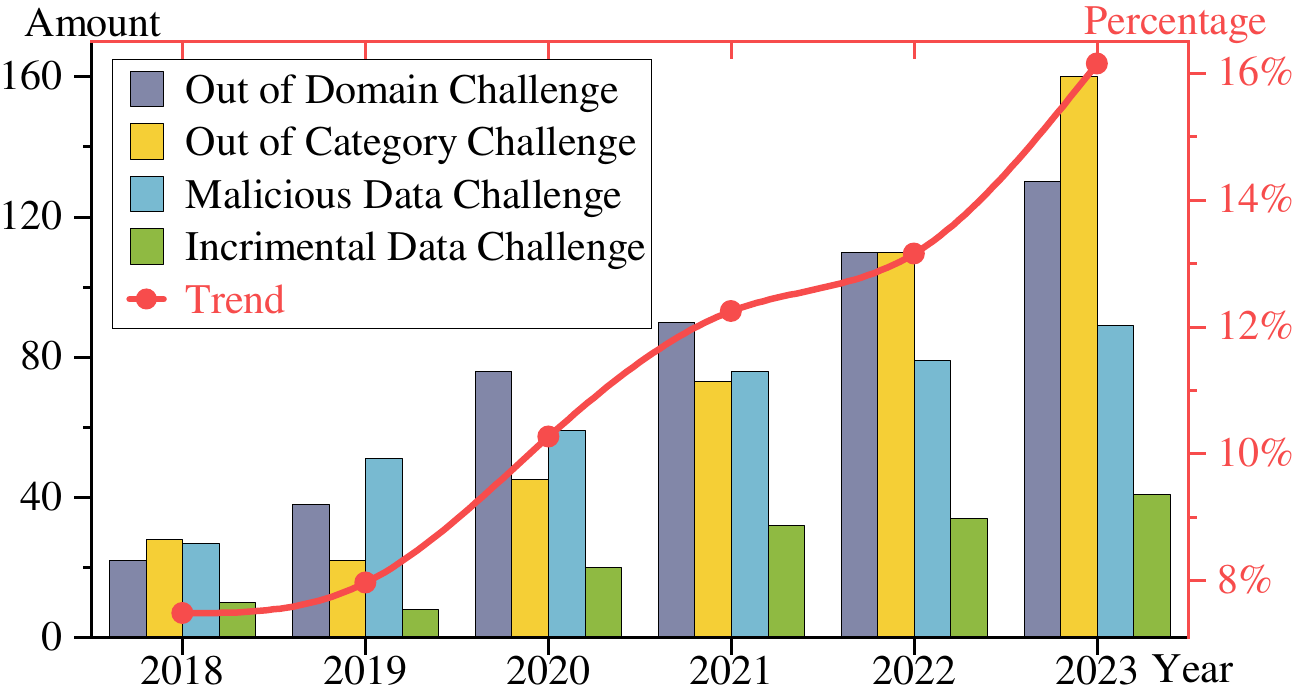}
  % \vspace{-8 pt}
  \caption{The escalating research interest and proportion of studies dedicated to the four major challenges encountered by open environment object detectors over the past six years (search with keywords from the \href{https://arxiv.org/}{arXiv} website).} %caption是图片的标题
  \label{numbers of paper in different years} %此处的label相当于一个图片的专属标志，目的是方便上下文的引用
  % \vspace{-0.15in}
  % \vspace{-12 pt}
\end{figure}

\begin{figure*}[t]
  \centering
  \includegraphics[width=1.0\textwidth]{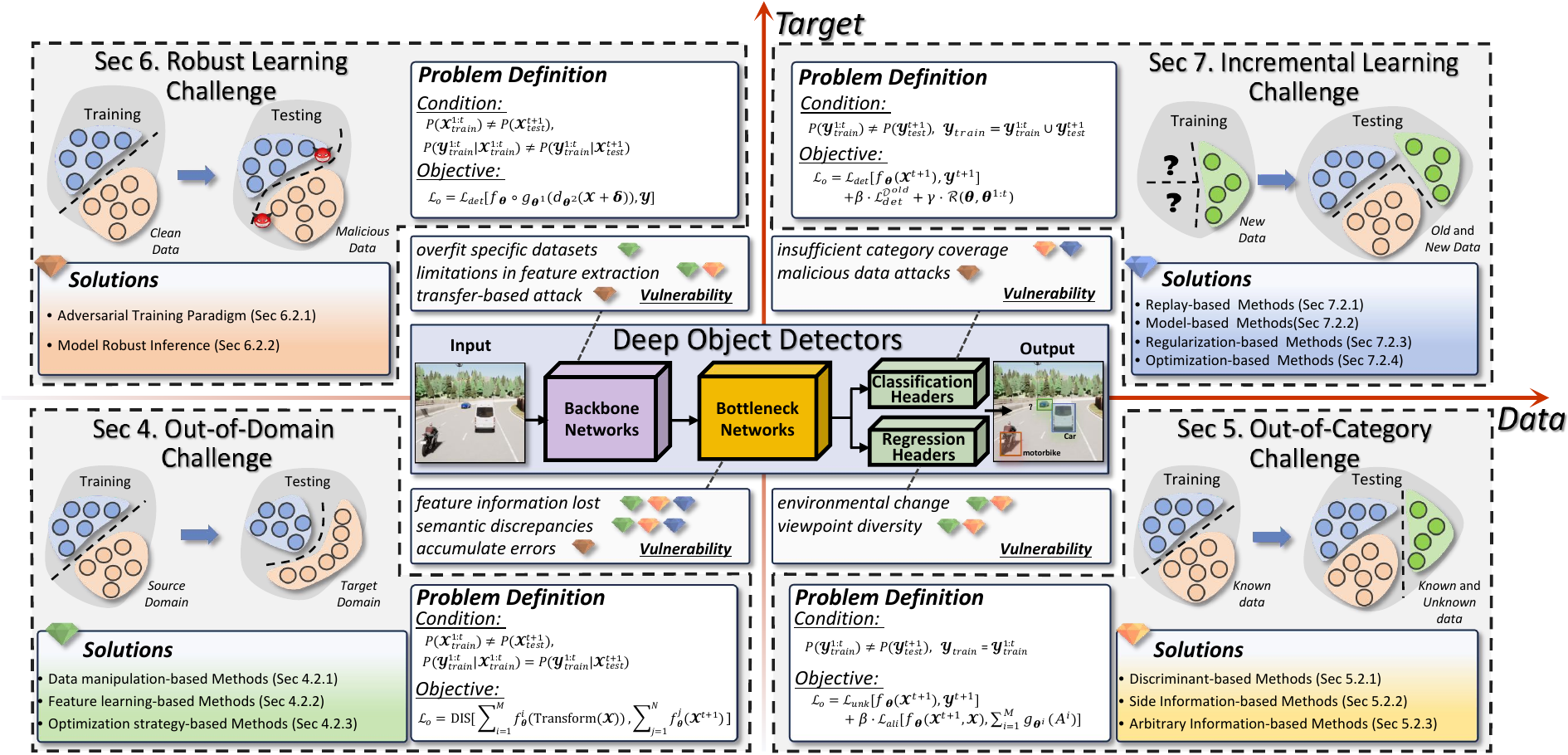}
  % \vspace{-15 pt}
  \caption{The open environment object detector challenge framework, encompasses four quadrants including out-of-domain, out-of-category, robust learning, and incremental learning based on the dimensions of data/target changes.}
  \label{fig:Overall framework}
  % \vspace{-0.10in}
\end{figure*}

Some surveys have focused specifically on the performance of deep object detectors when faced with certain challenges in open environments, such as domain adaptation issues~\cite{OzaSVP23}. More related surveys can be found in \emph{Supplementary Materials A}. However, there still exists no comprehensive and holistic analysis of the distinctive characteristics, challenges, and corresponding solutions of object detectors in open environments, which hinders their secure deployment in critical real-world scenarios. This survey bridges this gap by comprehensively examining object detectors in open environments, emphasizing the resilience of deep detection models to data variability and target alterations. In particular, we initially perform a structural analysis to identify vulnerabilities within the existing detection pipeline. Subsequently, we propose a four-quadrant categorization framework, which distinctly illustrates the interplay and distinctions among these challenges across two dimensions: data variation and target variation, represented on the horizontal and vertical axes, respectively. By delineating each quadrant, we provide a detailed analysis of the formulations, challenges, and potential solutions for deep object detectors in each specific context. We
also quantitatively benchmark and discuss the reviewed methods over multiple widely adopted detection datasets. Additionally, we highlight prospective research directions for each problem area and emphasize the importance of strengthening the connections between these challenges. The research trend illustrated in Fig.~\ref{numbers of paper in different years} highlights the growing interest and active participation of researchers in this field. This underlines the importance of our detailed review of object detection in open environments, aiming to clarify and address the challenges faced by object detectors and encourage innovative solutions in the real world. Our \textbf{contributions} can be summarized as follows:

%For instance, models in testing environments may face novel objects, elevating recognition challenges—the out-of-category data problem. Concurrently, the distribution of test data might shift, necessitating adaptability in models, such as an autonomous driving model trained on daytime conditions adapting to nighttime environments, highlighting the out-of-domain data problem. Furthermore, beyond accuracy and efficiency considerations, the model must exhibit scalability and adaptability when encountering new categories addressing the class incremental problem. Additionally, the model's robustness against malicious data problems, such as adversarial sample attacks on the perception layer of autonomous driving models, is crucial. By analyzing research papers from the past six years on the ~\href{https://arxiv.org}{arXiv} website, Fig.~\ref{numbers of paper in different years} delineates the attention these challenges have garnered in the object detection field and their prevalence in deep object detection research. The evident substantial interest and engagement of researchers in this area underscore the necessity and relevance of our review of prominent methodologies for object detection in open environment. This comprehensive examination aims to provide clarity and insight into object detection models' complexities and evolving nature, facilitating further innovation and advancement in the field.

\begin{itemize}
    \item This survey for the first time developed a four-quadrant taxonomy encompassing the primary challenges faced by object detectors in the open environment. Additionally, our analysis delves into the vulnerabilities of deep object detection architectures, elucidating the relationship between modules and the challenges.
    
    \item Based on the challenge framework, we emphasize four pivotal objectives for open environment object detectors and provide an in-depth examination of the specific issues and associated solutions.
    
    \item We also quantitatively benchmark the reviewed methods over multiple widely adopted detection datasets and explore potential directions for the future development of object detection in the open environment. %In contemplating the evolution of deep detection models in an open-world context, we propose a need to address under-explored challenges specific to certain problem areas. Moreover, we advocate for the integration of various objectives to forge more adaptable deep detection paradigms. For example, combining aspects of reliability and scalability could enhance the detection and recognition of unknown categories. This motivation aims to broaden the scope and applicability of deep detection models, ensuring their effectiveness and relevance in the dynamically evolving landscape of open environment detection.
\end{itemize}

The rest of the paper is organized as follows. Section \ref{sec:backgrounds} describes the development trend of deep object detectors and introduces the open environment problem. Section \ref{sec:framework} analyzes the vulnerability in each component of existing detectors and the overall goal in open environments. Sections \ref{sec:challenge1}, \ref{sec:challenge2}, \ref{sec:challenge3}, and \ref{sec:challenge4} detail the challenges, difficulties, and solutions associated with out-of-domain, out-of-category, robust learning, and incremental learning in the open environment, respectively. Section \ref{sec:benchmark} summarizes and benchmarks the performance of existing methods on multiple datasets. Section \ref{sec:outlook} summarizes the work and discusses several potential research directions. %Given the space limitations, a more expansive discussion of the solutions and a more detailed display of method properties are presented in the supplementary material. In addition, we also provide the datasets and benchmark for the above challenges to show the solutions's performance.

%% file: sections/2-background.tex
\section{Preliminaries and Backgrounds}
\label{sec:backgrounds}

In this section, we briefly review the evolution of deep object detectors and address the four primary challenges they face in an open environment.

\subsection{Deep Object Detector}
% Due to the advanced abstraction capabilities of convolutional neural networks (CNNs) and the convenience of automatic feature representation learning, researchers have shifted from hand-designed heuristic detection algorithms to a focus on deep object detectors. As a notable example of end-to-end detectors, Faster R-CNN~\cite{ren2015faster} revolutionizes the field by replacing selective search algorithms with Region Proposal Networks~\cite{ren2015faster} (RPNs). This advance, marked by the introduction of the novel anchor concept, signifies a milestone in deep object detection. The anchor technique employs a set of predefined boxes varying in scale and ratio, reframing the object detection task into two primary objectives: classifying objects within the anchors and measuring the distance from these anchor points to the actual objects. Thus, deep object detectors are generally divided into two main categories: \emph{anchor-based} detectors and \emph{anchor-free} detectors.
The shift from heuristic to deep object detectors has been catalyzed by CNNs' capability for abstraction and automatic feature learning. A pivotal development in this domain is Faster R-CNN~\cite{ren2015faster}, which was innovated with Region Proposal Networks and the concept of anchors, defining a new era in object detection. Anchors, with varied scales and ratios, bifurcate the detection task into object classification and localization. Consequently, this advancement categorizes deep detectors into two main types: \emph{anchor-based} detectors, which utilize this methodology, and \emph{anchor-free} detectors, delineating a fundamental classification in the field.

\textbf{Anchor-based detectors} enhance object detection by using anchors to generate candidate proposals. These systems either pre-defined or dynamically adjust critical hyperparameters of the anchors, such as quantity and scale, to accurately predict potential object proposals. Generally, anchor-based detectors are categorized into \emph{single-stage} and \emph{two-stage} models. Single-stage detectors, including SSD~\cite{liu2016ssd}, YOLO~\cite{redmon2016you}, and RetinaNet~\cite{lin2017focal}, directly execute category prediction and position estimation on the feature map using generated anchors, aiming for a compromise between detection speed and accuracy. In contrast, two-stage detectors, represented by Faster R-CNN, Mask R-CNN~\cite{he2020mask}, and Sparse R-CNN~\cite{sun2021sparse}, initially utilize RPNs to select anchors as preliminary proposals. These proposals are then refined in terms of their categories and locations in a subsequent stage. %In summary, single-stage detectors prioritize an optimal balance between speed and accuracy to enhance detection efficiency, whereas two-stage detectors stand out in terms of operational ease and high accuracy, markedly surpassing conventional detection algorithms.

The limitations of the anchor technique, such as its heavy reliance on training datasets and computational demands for label assignment, have spurred the development of \textbf{anchor-free detectors}. These new models redefine object detection tasks in terms of keypoint prediction or dense prediction. In keypoint prediction, a category of anchor-free detectors identifies the location and class of an object by defining and predicting representative key points of the object, such as CenterNet~\cite{duan2019centernet}. In contrast, within dense prediction, another branch of anchor-free detectors adopts concepts from semantic segmentation, using the object's location directly as a training sample, including FCOS~\cite{tian2019fcos}, FSAF~\cite{zhu2019feature}, and FoveaBox~\cite{kong2020foveabox}. Furthermore, researchers have innovated the structure of anchor-free detectors, such as employing the self-attention mechanism~\cite{carion2020end} to capture global dependencies in an image, which led to the design of DETR~\cite{carion2020end} and Deformable DETR~\cite{zhu2020deformable}.

%. For instance, CenterNet~\cite{} employs a convolutional neural network to extract image features, subsequently predicting the likelihood of each pixel point being the center of an object. 

 %The anchor-free detector demonstrates another developmental idea of deep object detection and shows unique advantages in terms of operation speed and scalability.

% From the history of depth detectors, the development of high-precision or high-efficiency object detectors has been the core pursuit of professionals in the field. As the real-world applications of depth detectors continue to expand, the research focus has gradually shifted to more complex and practical open environments, which has become the main focus of our survey.

\subsection{Open Environment Challenges}
\label{Open Environment Problems}
%Traditional machine learning approaches are designed to work within a closed set where training and testing datasets have similar distributions. 
Deep object detectors excel in controlled settings but face hurdles in open environments, like those encountered in autonomous driving, where conditions and objectives vary dynamically~\cite{zhou2022open}. To capture the evolution of the dataset, we introduce $\mathcal{D}^{1:t}$ for data up to time $t$ and $\mathcal{D}^{t+1}=\{\mathcal{X}^{t+1}, \mathcal{Y}^{t+1}\}$ for the subsequent time $t+1$. We identify four key challenges (Fig.~\ref{fig:Overall framework}) that affect detectors in open environments as follows:

\ding{182} \textbf{Out-of-domain challenge} deals with the model's ability to generalize across significantly different domains, such as adapting from urban to rural settings in autonomous driving systems. Using fundamental models can partially help with out-of-domain data challenges but can lead to overfitting and reduced performance in new domains. The condition lies in the discrepancy between the training set $\mathcal{X}^{1:t}_{train}$ and the test set $\mathcal{X}^{t+1}_{test}$, represented by $P(\mathcal{X}^{1:t}_{train}) \neq P(\mathcal{X}^{t+1}_{test})$. Despite differing statistical properties, the model is expected to maintain consistent predictions, hence $P(\mathcal{Y}^{1:t}_{train} \mid \mathcal{X}^{1:t}_{train})= P(\mathcal{Y}^{1:t}_{train} \mid \mathcal{X}^{t+1}_{test})$, underscoring the need for robust generalization.

\ding{183} \textbf{Out-of-category challenge} denotes a deep detector's capacity to recognize and process categories not present in its training data. For example, autonomous driving systems may encounter new vehicles or unexpected road obstacles that are absent from the training data set but emerging from the test data. This leads to a discrepancy between the label distributions of the training $\mathcal{Y}^{1:t}_{train}$ and test datasets $\mathcal{Y}^{t+1}_{test}$. The conditional assumption for this problem is denoted as $P(\mathcal{Y}^{1:t}_{train})\neq P(\mathcal{Y}^{t+1}_{test}), \mathcal{Y}_{train}=\mathcal{Y}^{1:t}_{train}$. The former condition highlights a notable disparity in label distribution between training and testing datasets, while the latter implies that typically only labels available up to time $1:t$ are used for training.

% \ding{184} \textbf{Robust learning challenge} aims to improve the robustness of the model to malicious data in an adversarial setting. For automated driving, adversaries could manipulate road sign data to impair the model's recognition capabilities or induce incorrect predictions. Despite their relatively low likelihood, such attacks are a serious concern for system robustness. Malicious data $\mathcal{Y}^{t+1}_{test}$ at time $t+1$ have distribution that may differ markedly from benign training data $\mathcal{Y}^{1:t}_{train}$ at time. The conditional premise for this issue can be denoted as $P(\mathcal{X}^{1:t}_{train}) \neq P(\mathcal{X}^{t+1}_{test}), P(\mathcal{Y}^{1:t}_{train} \mid \mathcal{X}^{1:t}_{train})\neq P(\mathcal{Y}^{1:t}_{train} \mid \mathcal{X}^{t+1}_{test})$. These malicious data not only result in a large difference between the data distribution of the test set and the training set, but can also change the label distribution of the test set by manipulating the categories of the malicious data.
\ding{184} \textbf{Robust learning challenge} focuses on enhancing model resilience against adversarial manipulation, such as altering road sign data in automated driving, which could lead to misrecognition or false predictions. Malicious data $\mathcal{X}^{1:t}_{train}$ at time $t+1$ may have a distribution significantly different from benign training data $\mathcal{X}^{t+1}_{test}$, as expressed by conditions $P(\mathcal{X}^{1:t}_{train}) \neq P(\mathcal{X}^{t+1}_{test})$. Moreover, a test dataset that not only differs in data distribution from the training dataset but may also have manipulated label distributions, is denoted as $P(\mathcal{Y}^{1:t}_{train} \mid \mathcal{X}^{1:t}_{train})\neq P(\mathcal{Y}^{1:t}_{train} \mid \mathcal{X}^{t+1}_{test})$.

\ding{185} \textbf{Incremental learning challenge} presents a significant challenge for deep detectors, requiring them to learn new classes without forgetting previously acquired knowledge despite the continuous emergence of new categories. In the context of autonomous driving systems, for instance, the detector must continuously recognize both new types of vehicles and existing ones. This dual objective requires a balance between learning plasticity and memory stability, which means that the detection model is tasked with effectively learning the new category $y_{new}$ while maintaining its accuracy for older object categories. Thus, the training data category $\mathcal{Y}^{t+1}_{train}$ encompasses both the new object category at the moment $t+1$ and the old object categories $\mathcal{Y}^{1:t}_{train}$. The conditional prerequisites of the problem are expressed as $P(\mathcal{Y}^{1:t}_{train})\neq P(\mathcal{Y}^{t+1}_{test}), \mathcal{Y}_{train}=\mathcal{Y}^{1:t}_{train} \cup \mathcal{Y}^{t+1}_{test}$.

%% file: sections/3-object_detection_towards_open_environments.tex
\section{Open Environment Object Detection}
\label{sec:framework}

An open environment object detector is a computer vision system that accurately recognizes and localizes a variety of objects in variable, unconstrained environments. Compared to traditional detection systems, open environment detectors need to deal with more complex and unknown scenarios.

\subsection{Limitations of Existing Detectors}
Firstly, we dissect the deep object detector, a system comprising four core components, as illustrated in Fig.~\ref{fig:Overall framework}. Our analysis delves into the structural limitations of these components and uses colored diamond icons to elucidate their connection to open environment challenges.
\begin{figure*}[t]
  \centering
  \includegraphics[width=0.96\textwidth]{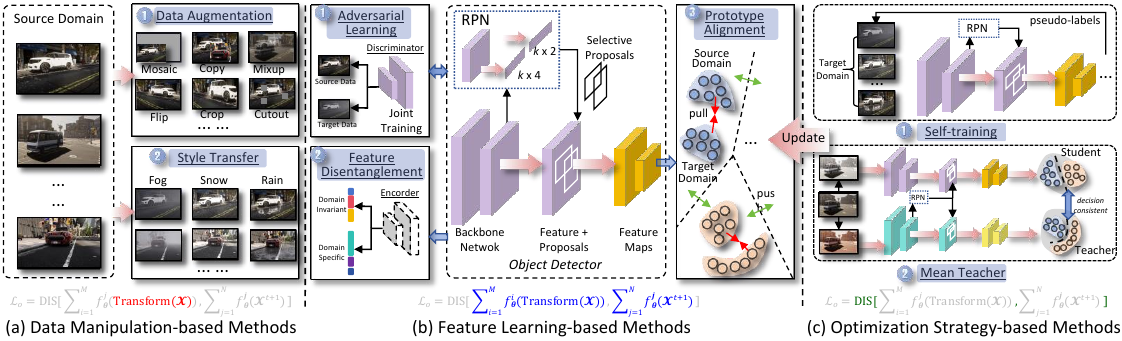}
  \caption{Illustration of solution categories for the out-of-domain challenge.}
  \label{out_of_domain}
  % \vspace{-0.15in}
\end{figure*}

\textbf{Backbone networks} are pivotal in advancing object detection. Liang \emph{et al.}~\cite{liang2022cbnet} emphasize the potential to significantly improve the performance of mainstream object detection systems (\eg, Faster R-CNN~\cite{ren2015faster}) by extending the width and depth of the backbone network. The development of backbone networks has also gone through three notable phases: traditional deep learning networks (ResNet50/101~\cite{he2016deep}), efficient compressed deep networks (MobileNet~\cite{sandler2018mobilenetv2}), and foundation models (CLIP~\cite{radford2021learning}) with richer semantic representations. These networks' tendencies to \emph{overfit specific datasets} and their \emph{limitations in feature extraction} may impede their ability to capture diverse and robust features, thus making feature extraction vulnerable to out-of-domain and out-of-category data challenges. Moreover, malicious attackers can exploit the use of the same backbone network in various detectors and enable \emph{transfer-based attacks}~\cite{wei2018transferable} without the need to access the model's internal information, challenging the model's robust learning.

\textbf{Bottleneck networks} in deep detectors are designed to streamline computational complexity and parameter count while retaining critical features. Their structure is intricately related to the core of the detection algorithm~\cite{tian2019fcos,duan2019centernet,he2020mask}, playing a crucial role in feature extraction and transformation. Commonly, these algorithms fall into three categories: multiscale hybrid feature detection~\cite{lin2017feature}, multilevel feature detection~\cite{cai2018cascade}, and combinations of both~\cite{tan2020efficientdet}. However, the inherent compression and simplification of the design could result in a feature \emph{information lost}. Furthermore, due to \emph{semantic discrepancies} across different scale features, direct fusion can adversely affect the expressiveness of the features~\cite{liu2018path}. Such limitations diminish the model's adaptability to novel scenarios and subtle features, resulting in subpar performance when encountering out-of-domain data, learning objects, and discriminating objects within or beyond the categories. Moreover, when the extracted feature maps of the backbone network include noise or defects, the neck network may \emph{accumulates errors} by utilizing multiscale features~\cite{lin2017feature}. This highlights a trade-off in neck networks: While they enhance computational efficiency, they might simultaneously reduce the model's robustness.

\textbf{Classification headers} in deep object detection primarily identify the predefined categories of detected objects within an image. In the open environment, the classification header encounters limitations similar to those in traditional image categorization, primarily concerning \emph{insufficient category coverage} and \emph{malicious data attacks}. The lack of sufficient category coverage is reflected mainly in the fact that the model needs to predefine the number of categorizing neurons, which can limit the extension of the model to out-of-category data ~\cite{VSGOSP21} and new category targets ~\cite{cui2023rt}. Malicious data attacks pose another significant threat to deep detectors. Attackers can deploy adversarial~\cite{athalye2018synthesizing} and backdoor attacks~\cite{lin2020composite} to trick the classification header into incorrectly classifying test samples, thus undermining the robustness of the model.

\textbf{Regression headers} aim to precisely determine the location of an object. It uses various strategies to define the bounding box, such as predicting the center point, width, and height of the bounding boxes~\cite{ren2015faster} or directly estimating the coordinates of the box~\cite{duan2019centernet} based on the features of the input image. Out-of-domain data (\eg, rain, snow) and malicious data can cause \emph{environmental changes} and \emph{viewpoint diversity} during the testing phase to jeopardize the generalization and robustness of the model. Specifically, complex physical environments and artificially constructed adversarial environments can result in varying lighting, weather conditions, viewpoints, and object sizes, which can significantly change the appearance of the object and affect the accuracy of positional regression.

% \textbf{Regression headers} aim to determine an object's location precisely. It uses various strategies to define the bounding box, like predicting the center point, width, and height of bounding boxes~\cite{ren2015faster} or directly estimating the box's coordinates~\cite{duan2019centernet} based on the input image's features. However, \emph{environmental changes} and \emph{viewpoint diversity} can compromise its generalization and reliability in the open-environment. Factors like varying lighting, weather conditions, viewpoints, and target sizes can significantly alter the target's appearance compared to the training data, impacting the accuracy of positional regression and posing a challenge to the regression network, \ie, out-of-domain data problem. %Moreover, the diversity in target sizes and viewpoints can extend beyond the training data's scope, posing a challenge to the regression network's predictive capability.

\subsection{Optimization Objectives}
%Standard loss functions for generic deep detection models include object classification loss and location regression loss. The overall objective of a deep detection model in an open environment setting needs to consider the various challenges that may be encountered in real-world applications, as described in Sec.~\ref{Open Environment Problems}. %Thus, in addition to the traditional loss functions, additional loss functions need to be introduced to cope with the aforementioned challenges.

This subsection symbolizes the overall optimization objective of the open environment object detector and discusses the characteristics and design of the objective function under different challenges.

Assuming that we have the training set $\mathcal{D} = \{\mathcal{X}, \mathcal{Y}\} = \{(\bm{x}_i, \bm{y}_i, \bm{b}_i)\}_{i=1}^{N}$, where $\bm{x}_i$ represents the image of the $i$-th samples $\mathcal{X}$ and $\mathcal{Y}$ signifies the positions $\bm{b}_i$ and categories $\bm{y}_i$ corresponding to multiple objects. $N$ denotes the total samples in the training set $\mathcal{D}$. The deep detection model $f_{\bm{\theta}}$ aims to optimize its performance by learning the parameter $\bm{\theta}$. Thus, the optimization objective can be expressed as:
\begin{equation}
    \min \sum_{i=1}^{N}[\mathcal{L}_{cls}(f_{\bm{\theta}}(\bm{x}_i), \bm{y}_i)+ \mathcal{L}_{loc}(f_{\bm{\theta}}(\bm{x}_i), \bm{b}_i)] + \lambda \cdot \mathcal{L}_{\text{o}}.
\end{equation}

Deep detection models typically focus on two primary types of losses: classification loss ($\mathcal{L}_{cls}$) and localization loss ($\mathcal{L}_{loc}$). Furthermore, to address the complexities of the open environment, these models integrate additional open loss components $\mathcal{L}_{\text{o}}$ with the hyperparameter $\lambda$. 

The open environment loss is tailored to handle the dynamic and unpredictable changes encountered in such settings as described in Sect.~\ref{Open Environment Problems}. Next, we will detail the specific loss functions adapted to the out-of-domain challenge (Section \ref{sec:challenge1}), the out-of-category challenge (Section \ref{sec:challenge2}), robust secure learning (Section \ref{sec:challenge3}), and adaptive incremental learning (Section \ref{sec:challenge4}), and discuss the core difficulties and solutions for each challenge.

%% file: sections/4-out-of-domain.tex
\section{Out-of-Domain Challenge}
\label{sec:challenge1}

% This section focuses on the out-of-domain challenge, discussing the difficulties and solutions of adapting deep detectors to diverse and cross-domain data.

% \subsection{Problem Definition}
% Generalization requires a model to effectively process and adapt to new data sources, types, or distributions that were not encountered during its training phase. This capability is especially crucial when dealing with ``out-of-domain'' data, where the testing data exhibits a different distribution to the training data, often due to varying environmental conditions, human influences, or differences in capture devices. In autonomous driving systems, a detector trained predominantly under sunny weather conditions might struggle to maintain performance in adverse weather conditions like rain or fog. This variance in data distribution necessitates the model to generalize beyond its initial training parameters.

%Therefore, in this paper, we define the problem faced by deep detectors in handling out-of-domain data.

This section focuses on the out-of-domain challenge, defining the fundamental goals for deep detectors, and discussing the difficulties and solutions of domain adaptive object detection (DAOD). 

Due to varying environmental conditions, human influences, or differences in capture devices, the testing data exhibits a different distribution from the training data. In autonomous driving systems, a detector trained predominantly under sunny weather conditions might struggle to maintain performance in adverse weather conditions like rain or fog. This variance in data distribution requires the model to generalize beyond its initial training parameters, \ie, the object detector is required to effectively process and adapt to new data sources, types, or distributions that were not encountered during its training phase. The open loss function can be specified as follows:

\begin{definition} [Domain Adaptive Object Detection]
Suppose the source domain data in the training phase is $ \mathcal{X} $. The data distribution during testing at time $t+1$ changes, \ie, the target domain data is $\mathcal{X}^{t+1}$. Facing the out-of-domain data, the goal of the detector is to maintain the same level of prediction as that of the source domain on the target domain data, which can be expressed as:
\begin{equation}
        % \mathcal{L}_{\text{o}}= \sum_{j=1}^M \textcolor{blue}{\underbrace{\text{DIS} [\textcolor{deepgreen}{ \underbrace{\textcolor{deepgreen}{f_{\bm{\theta}}^j}\textcolor{blue}{(}\textcolor{red}{\overbrace{\text{Transform}(\mathcal{X})}^{\text{Data Manipulation}}}\textcolor{blue}{)}, \textcolor{deepgreen}{f_{\bm{\theta}}^j}}_{\textcolor{deepgreen}{\text{Optimization Strategy}}} \textcolor{blue}{(\mathcal{X}^{t+1})} }]}_{\text{Feature Learning}}},
    % \mathcal{L}_{o} =  \text{DIS} [\sum_{i=1}^{M} f_{\bm{\theta}}^{i}(\text{Transform}(\mathcal{X})), \sum_{j=1}^{N} f_{\bm{\theta}}^{j}(\mathcal{X}^{t+1}) ],
    \mathcal{L}_{o} = \textcolor{deepgreen}{\overbrace{\text{DIS} [\textcolor{blue}{\underbrace{\sum_{i=1}^{M} f_{\bm{\theta}}^{i}(\textcolor{red}{\underbrace{\text{Transform}(\mathcal{X})}_{\text{Data Manipulation}}})\textcolor{deepgreen}{,} \sum_{j=1}^{N} f_{\bm{\theta}}^{j}(\mathcal{X}^{t+1})}_{\text{Feature Learning}}}]}^{\text{Optimization Strategy}}},
\label{ood}
\end{equation}
where $\mathcal{L}_{\text{o}}$ is the loss for out-of-domain data with a given parameter $\bm{\theta}$, $\text{DIS}$ is used to measure the difference between source and target domain data. $i$ and $j$ represent the network parameters of the different layers. $\text{Transform}$ is used to imitate data changes of the target domain.
\end{definition}

Specifically, for unsupervised domain adaptation, the model is trained without the labeled data of the target domain; whereas in semi-supervised or weakly supervised scenarios, the model may have some amount of target domain labeled data to assist in training. Next, we clarify the core difficulties for object detectors and divide the above Eq.~\eqref{ood} into three parts (Fig.~\ref{out_of_domain}) to correspond to the subsequent solution separately.

\subsection{Core Difficulties}
\ding{182} \textbf{Adaptation difficulty with unknown object location.} The difficulty lies in the data challenge. Standard object detectors typically involve multiple stages of feature extraction, including image-level global features, object-level instance features, and pixel-level local features. Due to the unknown location of each object, it becomes challenging to accurately realize adaption between corresponding objects of the same category in the source and target domains across these various feature extraction stages.

%The difficulty lies in how models can effectively manage and adapt to heterogeneous data with varied types and environmental characteristics during training and testing. Models must generalize across multiple domains, handling datasets from diverse contexts, capture conditions, and formats. This necessitates exploring detector's cross-domain invariant features, like feature representations and bounding box predictors, to find consistent elements across these datasets.

\ding{183} \textbf{Scalability and multi-task adaptation.} The difficulty arises from both architecture and task challenges. Due to the diverse range of object detection architectures, comprising multiple stages, determining where to place the domain adaptation module and designing a versatile plug-and-play module applicable to different architectures are inherently difficult. Additionally, since object detection involves both classification and localization tasks, designing a domain adaptation module that simultaneously addresses multiple tasks is also a challenging aspect.
\vspace{-10 pt}
\input{tables/solution/OOD_solution}

\subsection{Solutions}
We categorize the existing mainstream solutions into three distinct groups, as shown in Fig.~\ref{out-of-domain_solution}.

\subsubsection{Data Manipulation-based Methods}\label{4.3.1}

%\begin{figure}[!t]
 % \centering
  %\includegraphics[width=0.48\textwidth]{images/OOD/OOD_1.pdf}
 % \caption{Feature learning-based methods.}
 % \label{Feature learning-based methods}
%\end{figure}

Data manipulation-based methods manipulate the input directly to help in learning generalized detectors, as shown in Fig.~\ref{ood} (a). The above process can be represented by the `Transform' marked in red in Eq.~\eqref{out_of_domain}.  

\textbf{Data augmentation} improves the model generalization to domain variations by adding additional data. Kim \etal~\cite{KimJKCK19} diversity the distribution of labeled data by generating various distinctive shifted domains from the source domain, and then applying adversarial learning to encourage feature representation to be indistinguishable among the domains. Khirodkar \etal~\cite{KhirodkarYK19} propose to solve the domain gap through the use of appearance randomization to generate a wide range of synthetic objects to span the space of realistic
images for training. Prakash \etal~\cite{PrakashBBACSSB19} propose structured domain randomization to bridge the reality gap by generating images that enable the neural network to take the context around an object into consideration during detection. Wang \etal~\cite{WangLS21} propose a novel augmented feature alignment network that integrates intermediate domain image generation and domain-adversarial training into a unified framework. Hsu \etal~\cite{hsu2020progressive} propose to bridge the domain gap with an intermediate domain, which is constructed by translating the source images to mimic the ones in the target domain. Chen \etal~\cite{ChenZD0D20} generate synthetic samples from its counterpart domain to fill in the distributional gap between domains with CycleGAN~\cite{ZhuPIE17}. 
To deal with the few-shot problem in the target domain, Gao \etal identify target-similar samples in the source domain and utilize these samples to augment the target domain. Huang \etal~\cite{HuangLCWHL18} propose a structure-aware image-to-image translation network, which generates target-like images to train Faster-RCNN and YOLO~\cite{RedmonDGF16}.

\textbf{Style transfer} narrows the gap between two domains by adapting the style of one domain to match that of another domain. Rodriguez \etal~\cite{RodriguezM19} use a style transfer method for pixel-adaptation of source images to the target domain, and then enforce low distance in the high-level features of the object detector between the style transferred images and the source images to improve performance in the target domain. Yun \etal~\cite{YunHLKK21} transfer target-style information to source samples and simultaneously train the detection network with these target-stylized source samples in an end-to-end manner. Yu \etal~\cite{YuWCKSYLLS022} propose fine-grained domain style transfer with finer image details preserved for detecting small objects. Vidit \etal~\cite{ViditES23} leverage a pre-trained vision-language model to introduce semantic domain concepts via textual prompts, and then propose a semantic augmentation strategy acting on the features extracted by the detector backbone. 

\iffalse
\textbf{Style transfer.} Huang \etal~\cite{HuangLCWHL18} propose a structure-aware image-to-image translation network, which generates target-like images to train the Faster-RCNN and YOLO~\cite{RedmonDGF16}. Rodriguez \etal use a style transfer method for pixel adaptation of source images to the target domain, and then enforce low distance in the high-level features of the object detector between the style transferred images and the source images to improve the performance in the target domain~\cite{RodriguezM19}. Yun \etal transfer target-style information to source samples and simultaneously trains the detection network with these target-stylized source samples in an end-to-end manner~\cite{YunHLKK21}. Yu \etal propose fine-grained domain style transfer to reduce the style gaps with finer image details preserved for detecting small objects~\cite{YuWCKSYLLS022}. \textbf{Vidit \etal}%加点大模型？ 
\fi

\subsubsection{Feature Learning-based Methods}\label{4.3.2}
Feature learning-based methods operate at various feature levels within a basic object detection framework (\eg, Faster R-CNN), thus acquiring domain-invariant features from both the source/target domain, which can be shown in the blue part labeled by Eq.~\eqref{ood}, denoted as (b) in Fig.~\ref{out_of_domain}. %Moreover, these methods can be further categorized into three sub-categories.

\textbf{Adversarial learning-based strategy} is the most popular in DAOD. Specifically, the feature extractor is trained to generate features that deceive the domain discriminator, while the domain discriminator is trained to accurately categorize the features as belonging to either the source or target domains. As a result, the feature extractor can learn to produce domain-invariant features. Chen \etal~\cite{Chen0SDG18} are the first to use adversarial learning for DAOD, aligning both image-level features of the backbone and instance-level features from the ROI. Saito \etal~\cite{SaitoUHS19} divide the features extracted from the backbone into local features and global features for alignment. Some studies conduct adversarial learning on discriminative regions~\cite{ZhuPYSL19,XuZJW20,LiDZWLWZ20} or specific layers within the backbone~\cite{HeZ19,NguyenTS20,FuXLD20,HeZ20}. Some studies realize category-aware alignment with multi-label classifiers~\cite{ZhaoGSY20} or memory-guided attention module~\cite{VSGOSP21}. Rezaeianaran \etal~\cite{RezaeianaranSAR21} aggregate features into some groups and conduct group-level adversarial learning. Jiang \etal~\cite{JiangCWL22} conduct category adaptation and bounding box adaptation with adversarial learning. Wang \etal~\cite{Wang00HZ0T21} propose a sequence feature alignment method for the adaptation of detection transformers. Hou \etal~\cite{HouZFL21} propose to learn pixel-wise cross-domain correspondences for more accurate adversarial learning. Zhao \etal~\cite{Zhao022} introduce a task-specific inconsistency alignment with adversarial learning.

\textbf{Prototype-based strategy} constructs class prototypes from features in the source and target domains and then reduces the distribution gap by minimizing the difference between corresponding class prototypes in both domains~\cite{PanYLWNM19}. For example, Xu \etal~\cite{XuWNTZ20} obtain prototypes of each class on the instance-level features and then perform prototype alignment by enhancing intraclass compactness and interclass separability. Zheng \etal~\cite{Zheng0LW20} combine prototype alignment and adversarial learning in a unified framework to realize coarse-to-fine feature adaptation. Zhang \etal~\cite{ZhangWM21} construct one set of learnable RPN prototypes and enforce the RPN features to align with the prototypes for both source and target domains, thereby the RPN features in the two domains can be aligned automatically. 

\iffalse
\textbf{Prototype-based methods.} In domain adaptive recognition, prototype-based methods~\cite{PanYLWNM19} construct class prototypes from features in the source and target domains. Subsequently, they reduce the distribution gap between the two domains by minimizing the difference between the corresponding class prototypes in both domains. Motivated by this, Xu \etal obtain the prototype of each class on the instance-level features and then perform prototype alignment through enhancing intra-class compactness and inter-class separability~\cite{XuWNTZ20}. Zheng \etal combine prototype alignment and adversarial learning in a unified framework to realize coarse-to-fine feature adaptation~\cite{Zheng0LW20}. Zhang \etal construct one set of learnable RPN prototypes and enforce the RPN feature to align with the prototypes for both source and target domains, thereby the RPN features in the two domains can be aligned automatically~\cite{ZhangWM21}.
\fi

\textbf{Feature disentanglement-based strategy} aims to separate domain-specific and domain-invariant features of the source and target domains, respectively~\cite{BousmalisTSKE16}. Su \etal~\cite{SuWZTCQW20} disentangle the domain-specific attribute through a domain embedding module. Wu \etal~\cite{WuLHZ021} propose a vector-decomposition-based disentangled method that utilizes difference operations to enhance domain-invariant representations. Wu \etal~\cite{WuHZY22,WuD22} propose a progressive disentangled mechanism to decompose domain-invariant and domain-specific features and propose a cyclic-disentangled self-distillation to deal with single-domain generalized object detection.

\subsubsection{Optimization Strategy-based Methods}\label{4.3.3}
%间接学习领域不变特征
\begin{figure*}[t]
    \centering
    \includegraphics[width=0.96\textwidth]{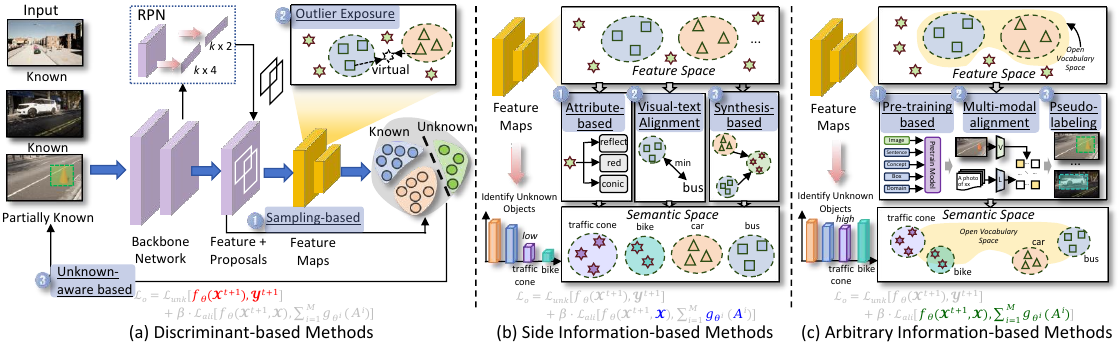}
    % \vspace{-8 pt}
    \caption{Illustration of solution categories for the out-of-category challenge.}
    \label{fig:out_of_category}
    % \vspace{-0.15in}
\end{figure*}

This type of method employs various optimization strategies to address the DAOD problem, such as self-training, mean teacher, \etc, as shown in Fig.~\ref{out_of_domain} (c). The `DIS' function (green part) in Eq.~\eqref{ood} is a key part of the optimization strategy, which directly affects the objective setting and evaluation criteria of the optimization process.

\textbf{Self-training-based strategy} utilizes the detector trained on the source domain to obtain pseudo-labels of the target domain, and then re-train the detector with the target domain and their pseudo-labels. As pseudo-labels may be noisy, RoyChowdhury \etal~\cite{RoyChowdhuryCSJ19} utilize high-confidence pseudo-labels to re-train the detector. Khodabandeh \etal~\cite{KhodabandehVRM19} formulate a self-training strategy as a robust learning paradigm that is robust to noisy pseudo-labels. Kim \etal~\cite{KimCKK19} propose a weak self-training that diminishes the adverse effects of noisy pseudo-labels to stabilize the learning procedure. Li \etal~\cite{LiCXYYPZ21} propose a metric of self-entropy descent to search for an appropriate confidence threshold for reliable pseudo-label generation and mine false negatives for pseudo-label denoising. To generate more informative pseudo-labels, Li \etal~\cite{LiH0Z21} propose a category dictionary-guided model, which learns category-specific dictionaries from the source domain to represent the candidate boxes in the target domain. Munir \etal~\cite{MunirKSA21} leverage the model’s predictive uncertainty to strike the right balance between adversarial learning and self-training. Lu \etal~\cite{0006ZS23} propose a robust self-training approach, where a tracking-based single-view augmentation is introduced to achieve weak-hard consistency learning.

\textbf{Mean teacher-based strategy} aims to encourage predictions of teacher and student consistent under small perturbations of inputs or network parameters~\cite{FrenchMF18}. Cai \etal~\cite{CaiPNTDY19} transform each target image into two perturbed samples with different augmentations and then input them into student and teacher models, respectively. They train the student model on the source domain and optimize the whole architecture with three consistency regularizations. Liu \etal~\cite{LiuMHKCZWKV21} extend the mean teacher framework to train an object detector in a semi-supervised fashion by exploiting large unlabeled data. Deng \etal~\cite{Deng0CD21} propose an unbiased mean teacher model with several simple yet highly effective strategies that could eliminate the model bias for the simple mean teacher in cross-domain scenarios. He \etal~\cite{HeWWWLLGWQ22} integrate detection branches of both source and target domains in a unified teacher-student learning scheme to reduce domain shift and generate reliable supervision effectively. Li \etal~\cite{LiDMLCWHKV22} propose a teacher-student framework named adaptive teacher which leverages domain adversarial learning and weak-strong data augmentation to address the domain gap. Kennerley \etal~\cite{KennerleyWVT23} propose a two-phase consistency training to deal with the problem of false-positive error propagation using a mean teacher framework. Moreover, Wu \etal~\cite{WuCHWLMGWW022} propose the target relevant knowledge preservation to deal with multi-source DAOD. Yao \etal~\cite{YaoZ0Y21} develop a novel pseudo subnet for the target domain and add consistency regularization between the pseudo subnet and other source subnets. More detailed classification scales can be found in \emph{Supplementary Materials B}.

%% file: tables/solution/OOD_solution.tex
\tikzstyle{leaf}=[draw=hiddendraw,
    rounded corners, minimum height=1em,
    fill=hidden-orange!40,text opacity=1, align=center,
    fill opacity=.5,  text=black,align=left,font=\tiny,
    inner xsep=3pt,
    inner ysep=1pt,
    ]
\begin{figure}[h]
\centering
\begin{forest}
  for tree={
  forked edges,
  grow=east,
  reversed=true,
  anchor=base west,
  parent anchor=east,
  child anchor=west,
  base=middle,
  font=\scriptsize,
  rectangle,
  draw=hiddendraw,
  rounded corners,align=left,
  minimum width=2em,
    s sep=5pt,
    inner xsep=3pt,
    inner ysep=1pt,
    text width=9em,
  },
  where level=1{text width=4.5em}{},
  where level=2{text width=4.5em,font=\scriptsize}{},
  where level=3{text width=9em,font=\tiny}{},
  where level=4{font=\tiny}{},
  where level=5{font=\tiny}{},
  [Out-of-domain Challenge,rotate=90,anchor=north,edge=hiddendraw
    [Data Manipu-\\lation (\S\ref{4.3.1})
        [Data Augm-\\entation
            [{Kim \etal ~\cite{KimJKCK19}, Khirodkar \etal~\cite{KhirodkarYK19},\\ Prakash \etal~\cite{PrakashBBACSSB19}, Wang \etal~\cite{WangLS21},\\ Hsu \etal~\cite{hsu2020progressive}, Chen \etal~\cite{ChenZD0D20},\\ CycleGAN~\cite{ZhuPIE17}, Gao \etal~\cite{GaoLYW023},\\ Huang \etal~\cite{HuangLCWHL18}},leaf,text width=9em, edge=hiddendraw]
        ]
        [Style Transfer
            [{Vidit \etal~\cite{ViditES23}, Rodriguez \etal~\cite{RodriguezM19},\\ Yun \etal~\cite{YunHLKK21}, Yu \etal~\cite{YuWCKSYLLS022}},leaf,text width=9em, edge=hiddendraw]
        ]
    ]
    [Feature Learn-\\ing (\S\ref{4.3.2})
     [Adversarial \\ Learning
        [{Chen \etal~\cite{Chen0SDG18}, Saito \etal~\cite{SaitoUHS19},\\ ROI-based grouping strategy~\cite{ZhuPYSL19},\\ categorical regularization~\cite{XuZJW20}, context \\ information~\cite{LiDZWLWZ20}, adversarial learning\\  on specific layers within the backbone\\ \cite{HeZ19, NguyenTS20, FuXLD20, HeZ20}, category-aware \\alignment~\cite{ZhaoGSY20}, memory-guided attent-\\ion module~\cite{VSGOSP21}, Rezaeianaran \etal\\~\cite{RezaeianaranSAR21}, Jiang \etal~\cite{JiangCWL22}, Wang \etal~\cite{Wang00HZ0T21},\\ Hou \etal~\cite{HouZFL21}, Zhao \etal~\cite{Zhao022}},leaf,text width=9em, edge=hiddendraw]
     ]
     [Prototype
        [{Xu \etal ~\cite{XuWNTZ20}, Zheng \etal ~\cite{Zheng0LW20}, Zhang \\ \etal ~\cite{ZhangWM21}},leaf,text width=9em, edge=hiddendraw
        ]
     ]
     [Feature Dise-\\ntanglement
        [{Su \etal~\cite{SuWZTCQW20}, Wu \etal ~\cite{WuLHZ021}, Wu \etal \\ \cite{WuHZY22, WuD22}},leaf,text width=9em, edge=hiddendraw]]
     ]
    [Optimization \\Strategy\\ (\S\ref{4.3.3})
      [Self-training
        [{RoyChowdhury \etal~\cite{RoyChowdhuryCSJ19}, Khodaban-\\deh \etal~\cite{KhodabandehVRM19}, Kim \etal~\cite{KimCKK19}, Li \etal\\~\cite{LiCXYYPZ21}, Li \etal~\cite{LiH0Z21}, Munir \etal~\cite{MunirKSA21}, Lu\\ \etal~\cite{0006ZS23}},leaf,text width=9em, edge=hiddendraw]]
    [Mean Teacher
      [{Cai \etal~\cite{CaiPNTDY19}, Liu \etal~\cite{LiuMHKCZWKV21}, Deng \etal\\ \cite{Deng0CD21}, He \etal~\cite{HeWWWLLGWQ22}, Li \etal~\cite{LiDMLCWHKV22}, Wu\\ \etal~\cite{WuCHWLMGWW022}, Yao \etal~\cite{YaoZ0Y21}, Kennerley\\ \etal~\cite{KennerleyWVT23}},leaf,text width=9em, edge=hiddendraw]]
     ]
]
\end{forest}
% \vspace{-8 pt}
\caption{Methods for addressing out-of-domain challenges.}
% \vspace{-10 pt}
\label{out-of-domain_solution}
\end{figure}

%% file: sections/5-out-of-category.tex
\section{Out-of-Category Challenge}
\label{sec:challenge2}
% This section focuses on the reliability of object detectors in handling out-of-category data. We discuss the core difficulties and category existing solutions as three groups in Fig.\ref{fig:out_of_category}.

% \subsection{Problem Definition}

% The reliability of deep detectors is mainly reflected in the recognition accuracy for out-of-category data in open environments, \ie, accurately recognizing known objects and effectively discriminating unknown objects (recognized as the background). Since novel and unexpected objects may appear in the real world, this requires the deep detector to have good adaptability and discrimination ability. In this paper, we define the problem of detecting out-of-category data as ``unknown category object detection.''%For example, incorrectly recognizing an unknown obstacle as a harmless object in an autonomous driving system may cause serious safety hazards.

This section focuses on the reliability of object detectors in handling out-of-category data. We firstly concretize the open loss function for out-of-category data, then discuss the core difficulties and existing solutions as three groups in Fig.~\ref{fig:out_of_category}.

The deep detectors for out-of-category data challenge is need to consider the recognition accuracy for unknown data in open environments, \ie, accurately recognizing known objects and effectively discriminating unknown objects (recognized as the background). Since novel and unexpected objects may appear in the real world~\cite{bao2021evidential,bao2022opental,bao2023latent}, this requires the deep detector to have good adaptability and discrimination ability. In this paper, we define the problem of detecting out-of-category data as ``unknown category object detection.'' Thus, the open loss function can be defined as follows:

\begin{definition}[Unknown Category Object Detection]
Suppose that the training images are $\mathcal{X}$ and the unknown image is $\mathcal{X}^{t+1} \in \mathcal{D}^{t+1}$ at moment $t+1$. In unknown category object detection, The model can directly discriminate or use additional information to indirectly separate unknown object with category $\mathcal{Y}^{t+1}$, as demonstrated by the following: 

\begin{equation}
\begin{split}
\mathcal{L}_{\text{o}} = & \mathcal{L}_{\text{unk}}[\textcolor{red}{ \overbrace{f_{\bm{\theta}}(\mathcal{X}^{t+1}), \mathcal{Y}^{t+1}}^{\text{Discriminant}}}] \\
& + \beta \cdot \mathcal{L}_{\text{ali}}[ \textcolor{deepgreen}{\underbrace{f_{\bm{\theta}}(\mathcal{X}^{t+1}, \textcolor{blue}{\overbrace{\mathcal{X}), \textcolor{deepgreen}{\sum_{i=1}^{M}}g_{\bm{\theta}^{\textcolor{deepgreen}{i}}}(A^{\textcolor{deepgreen}{i}}}^{\text{Side Information}})}}_{\text{Arbitrary Information}}}],
\label{ooc}
\end{split}
\end{equation}
where $\mathcal{L}_{\text{unk}}$ is the ability of the model to recognize unknown objects in the image as $\mathcal{Y}^{t+1}$, which can be binary (known \vs~unknown) or include specific categories of unknown objects (\eg, ``cat'').  The $\mathcal{L}_{\text{ali}}$ represents the auxiliary information function, which utilizes a set of auxiliary information $A^i$ (e.g., attribute labels or other modal information) through the function $g_{\bm{\theta}^i}$ and intends to provide context or guidance that can be helpful in identifying unknown categories. $\beta$ is a weighting parameter used to balance the relationship between the loss of unknown category recognition and the loss of auxiliary information, thus optimizing the model's ability to handle unknown objects.
\end{definition}

Next, we will describe the core difficulties faced by out-of-category data challenges and describe existing solutions in the context of the three parts (Fig.~\ref{fig:out_of_category}) of Eq.~\eqref{ooc}.

% For the computation of $\mathcal{L}_{ooc}$, 
% if the training data contains some samples marked as ''unknown``, 
% ``unknown'' samples $\mathcal{X}^{t+1}$ can be generated from training data $\mathcal{D}$ through outlier exposure or pseudo-label annotation, which can be optimized directly during the training process.
% If no such samples are included, other techniques, such as novelty detection or open-set detection, will be considered to optimize this loss.

\subsection{Core Difficulties}

\ding{182} \textbf{Data agnostic of unknown categories.} The absence of training data for unknown categories hinders the model to learn and identify these categories through parameter optimization. This leads to a scenario where the model cannot directly compute the loss function for unknown categories during training, complicating the distinction between known and unknown categories. The detector is required to recognize and process unknown categories without having been explicitly trained on them, necessitating a level of intuitive or inferential learning capability.

\ding{183} \textbf{Constructing robust representations from auxiliary information.} Leveraging additional information sources, such as textual descriptions or tags, can be helpful to build a deep semantic understanding of both known and unknown categories. It helps the model to make more informed predictions by utilizing the contextual cues provided by these auxiliary information sources. However, how to effectively utilize this additional information and align these auxiliary resources in the same context remains a difficulty to resolve.

\input{tables/solution/OOC_solution}

\subsection{Solutions}
We categorize the existing mainstream soluitons for out-of-category challenges into three distinct groups, as shown in Fig.~\ref{out-of-category_solution}.

\subsubsection{Discriminant-based Methods}\label{solution:discriminant}
The discriminant-based methods, also called open set object detection (OSOD) by the academic community, do not need to identify classes of unseen classes but instead mark them as unknown classes~\cite{han2022expanding, miller2018dropout}. This process is shown in Fig.~\ref{fig:out_of_category} (a). The red segment in Eq.~\eqref{ooc} is responsible for detecting and assigning the category of the unseen object as `unknown'. 

\textbf{Sampling-based strategy} is the initial method applied to OSOD, the core idea is Dropout Variational Inference, offering a practical way to measure model confidence uncertainty while remaining computationally feasible. Monte Carlo Dropout~\cite{miller2018dropout, miller2019evaluating} methods are proposed to detect unknown objects, but these are time-consuming due to the need for extensive detection box generation and clustering for class inference. To enhance efficiency, single-inference detection of unknown classes is preferable. 

\textbf{Outlier exposure strategy} creates virtual outliers from normal samples in the feature space, enhancing the model's ability to distinguish actual outliers. Du \etal~\cite{du2021vos} introduce VOS, a technique that creates virtual outliers from known class data to refine the decision boundary and assist in detecting unknown classes. Han \etal~\cite{han2022expanding} develop OpenDet, an uncertainty-guided method that mines OOD samples in low-density latent areas, facilitating network training to recognize unknown classes.

\textbf{Unknown-aware based strategy} estimates unknown objects in training data and co-trains a correction model, boosting uncertain object estimation performance. Joseph \etal~\cite{joseph2021towards} propose an energy-based method using predefined unknown classes for training, which, however, contradicts OSOD's principle of not using external class data during training. STUD~\cite{du2022unknown} effectively distills unknown objects from unstructured videos and meaningfully regularizes the model's decision boundary. Liu \etal~\cite{liu2022open} investigate the open-set problem within the context of semi-supervised object detection, focusing on detecting unknown classes in some unlabeled images and enhancing the accuracy of known class detection. ProposalCLIP~\cite{shi2022proposalclip} introduces a proposal regression mechanism that leverages CLIP cues to extract unknown objects. This process is designed to identify potential proposals belonging to unknown classes. OW-DETR~\cite{gupta2022ow} uses Deformable DETR as the detection baseline and proposes an attention-driven pseudo-labeling strategy to select candidate unknown queries. UC-OWOD~\cite{wu2022uc} discovers and generates unknown class objects and categorizes unknown classes into a limited number (less than four) of semantically distinct groups, allowing for the classification of detected unknown objects. Su \etal~\cite{su2023hsic} study the OSOD problem in the few-shot setting, which mines unknown-aware objects through deep evidential learning.

\subsubsection{Side Information-based Methods}\label{solution:side}

% The side information-based methods primarily address the issue of unknown category classification, which necessitates the use of side information to aid the detector during the test phase in locating and identifying objects that were not encountered during the training phase, and the number of identifiable unknown categories are usually limited~\cite{zhu2019zero}. The academic community has collectively termed these techniques zero-shot object detection (ZSOD)~\cite{huang2022robust}, which are trained on a given known dataset $\mathcal{D}_{k}$. Given a novel set $\mathcal{D}_n$, the corresponding novel category space is denoted as $\mathcal{Y}_n = \{K+1, K+2, \dots, K+K_n\}$, where $K_n$ denotes the number of the novel classes. During the testing phase, objects may belong to the classes of both $\mathcal{D}_{k}$ and $\mathcal{D}_{n}$. The detector needs to distinguish objects from the background and correctly recognize classes from $\mathcal{D}_{k}$ and $\mathcal{D}_{n}$ with side information.
Side information-based methods, termed Zero-Shot Object Detection (ZSOD), focus on classifying unknown categories using additional information~\cite{zhu2019zero,huang2022robust} as shown in Fig.~\ref{fig:out_of_category} (b). Side information, orthogonal to both input and output spaces and highlighted in blue in Eq.~\eqref{ooc}, pertains exclusively to the training inputs $\mathcal{X}$. These methods are trained on a known dataset with side information; during testing, the detector aims to differentiate objects from the background and accurately identifies classes using side information. ZSOD necessitates reasoning about unseen categories using side information like visual attributes~\cite{mao2020zero, zhu2019zero} or textual descriptions~\cite{li2019zero,yan2022semantics}. Side information is used to construct semantic information and is independent of the model input and output spaces~\cite{jonschkowski2015patterns}. 
% Most ZSOD methods rely on visual-semantic mapping alignment~\cite{bansal2018zero, yan2022semantics}, projecting visual features to a semantic feature space. 
This approach allows for flexible adjustment of the number of object categories to aid in identifying unknown classes~\cite{bansal2018zero,rahman2020improved}. 

\textbf{Attribute-based} methods initially convert image features into attribute space, subsequently deducing specific unknown classes, which use visual attributes as side information. Zhu \etal~\cite{zhu2019zero} propose the ZSOD method based on Yolo to jointly infer unknown classes by merging visual features and visual attribute prediction results Mao \etal~\cite{mao2020zero} utilize unsupervised learning methods to evaluate attribute tables and align visual features with attribute representations to reason about new classes. However, this paradigm struggles with generalizing attributes for unknown classes, impacting recognition~\cite{jayaraman2014zero}.
\begin{figure*}[!t]
\centering
\includegraphics[width=0.96\textwidth]{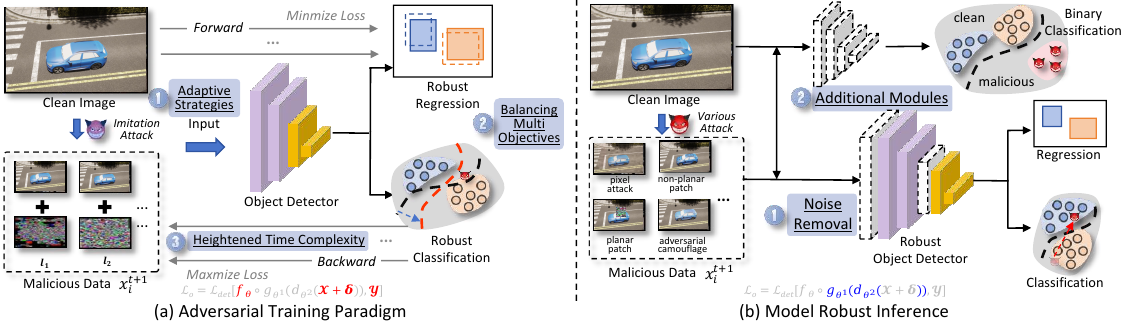}
\caption{Illustration of the solution categories for the robust learning challenge.}
\label{fig:malicious_data}
% \vspace{-0.15in}
\end{figure*}

\textbf{Visual-text alignment} methods leverage text as side information, aligning the spatial features of vision with text. It then deduces information about unknown classes through the established semantic relationships inherent in the text~\cite{pennington2014glove, rahman2020improved, li2019zero}.
A key challenge is the visual-semantic gap~\cite{yan2020semantics, rahman2020improved}. 
% Most studies prefer visual-text semantic alignment, using text and word vectors for strong, yet often noisy, semantic relationships~\cite{pennington2014glove, rahman2020improved, li2019zero}.
Sole reliance on categorical words leads to unstable visual-semantic alignment~\cite{rahman2020improved}, and fixed semantic information hinders accurate visual-semantic space alignment, \eg, SB~\cite{bansal2018zero}.
Several solutions have been proposed, including Yan \etal~\cite{yan2020semantics} with graph module for dynamic semantic information adjustment,  Rahman \etal~\cite{rahman2020improved} introduce polarity loss and vocabulary metric learning for word vector noise reduction, improving prediction distinction. Nie \etal~\cite{nie2022node} propose a graph-model-based approach to understanding relationships among multiple objects, addressing the complexity of object interactions in an image.  Yan \etal~\cite{yan2022semantics} with a semantic-guided contrastive network for unseen classes bias minimization.

\textbf{Synthesis-based} methods' core idea is to generate data distribution for unknown classes. Huang \etal~\cite{huang2022robust} employ a feature synthesizer for diverse visual feature generation from semantic features, reducing mis-classifications. Gupta \etal~\cite{gupta2023generative} achieve multi-label ZSOD by generating multi-label features. Li \etal~\cite{li2023zero} explore applying ZSOD to camouflaged object detection and propose a Camouflaged Visual Reasoning Generator (CVRG) to produce unknown class pseudo-features. 

\subsubsection{Arbitrary Information-based Methods} \label{solution:arbitrary}

% Arbitrary information-based methods focus on recognizing any class using diverse additional information, particularly utilizing multi-modal visual language models for identifying unknown categories. The concept of open-vocabulary object detection~\cite{zareian2021open}, emerging in this context, leverages these large-scale pre-trained models for versatile object detection in open-world scenes, aims to establish a feasible method for recognizing arbitrary object classes and allows the incorporation of any additional information, marking a significant development in object detection capabilities.

Arbitrary information-based methods aim to recognize any class using various additional data, especially through multimodal visual language models for detecting unknown categories, as shown in Fig.~\ref{fig:out_of_category} (c). This information can be related to both unknown images and known images, and is recorded as the green part in Eq.~\eqref{ooc}. Open-vocabulary object detection~\cite{zareian2021open, bravo2023open, chen2023ovarnet}, a concept in this realm, uses large-scale pre-trained models for flexible detection across open-world scenes. This approach enables the recognition of arbitrary object classes using diverse supplementary information, representing a notable advancement in object detection technology.

% \begin{definition}[Open-Vocabulary Object Detection]
%     Open-vocabulary object detection aims to establish a feasible method for recognizing arbitrary object classes and allows the incorporation of any additional information. It also has the function of zero-shot detection. 
% \end{definition}

% Zareian \etal~\cite{zareian2021open} trained a multimodal visual language model by grounding the image-caption pairs. They only use limited data with instance box annotations, cover a wider range of image captions, and can achieve performance beyond zero-shot detection. Novel class images may appear in the training set, but there is no accurate bounding box label. However, training such a visual language backbone is very computationally resource-intensive.

\textbf{Multi-modal alignment based} strategy aims to synchronize image features with a range of open vocabularies, encompassing elements like captions and concept descriptions.
OVR-CNN~\cite{zareian2021open} train a multimodal visual language model on limited instance box-annotated data and extensive image captions. This approach, while effective, demands substantial computational resources.
Current open-vocabulary object detection methods often employ multimodal pre-trained models like CLIP~\cite{radford2021learning}, trained on diverse image-text pairs but focusing more on image-level than region-level information. This leads to inaccuracies in object localization with CLIP in detection tasks. To address this, Gu \etal~\cite{guopen} propose ViLD, adapting CLIP for region-level feature extraction and integrating it with Faster R-CNN. Ma \etal~\cite{ma2022open} add image caption-based global knowledge distillation. Du \etal~\cite{du2022learning} develop DetPro, enhancing ViLD by adding a learnable token to the language encoder input. These knowledge distillation methods are time-intensive. To improve efficiency, Kuo \etal~\cite{kuo2023f} introduce F-VLM, using frozen CLIP directly as the detection backbone. Wu \etal~\cite{wu2023cora} present CORA, employing a region prompt with a frozen visual encoder for better unknown object localization. 

\textbf{Pseudo-labeling based} strategy employs CLIP to identify unknown class objects in unlabeled data. These pseudo-labeled data are then used to jointly train detectors, enhancing their capability to recognize unknown classes.
% Additionally, other approaches use CLIP to identify unknown classes in unlabeled data, create pseudo-labels, and train detectors with these labels. 
Gao\etal~\cite{gao2022open} develop a method for automatically generating pseudo-bounding box annotations for various objects using large-scale image-caption pairs. This approach involves training the model with these pseudo-labels to enhance the inference capabilities for unknown classes. PromptDet~\cite{feng2022promptdet} employs online resources to generate pseudo labels for unknown classes by utilizing a vast array of unlabeled images from the internet. This system is then jointly trained with these generated labels.

\textbf{Pre-training based} strategy focuses on using large, diverse datasets and targeted alignments to develop improved multi-modal pre-trained models to help identify unknown classes. In addition to CLIP, there are various innovative approaches to developing multi-modal pre-training models for object detection tasks. GLIP~\cite{li2022grounded} integrates object detection with phrase-based pre-training to learn object-level, language-aware visual representations. RegionCLIP~\cite{zhong2022regionclip} focuses on matching specific image regions with corresponding text in captions, pre-training the model to align these region-text pairs in the feature space. Wang \etal~\cite{wang2023detecting} propose UniDetector to recognize enormous categories in various scenes using RegionCLIP. DetCLIP~\cite{yao2022detclip} and DetCLIPv2~\cite{yao2023detclipv2} introduce a parallel concept formula, enabling the separate extraction of concepts. RO-ViT~\cite{kim2023region} is a contrastive image-text pre-training technique aimed at linking image-level pre-training with open-vocabulary target detection. More detailed classification scales can be found in \emph{Supplementary Materials C}.

%% file: tables/solution/OOC_solution.tex
\tikzstyle{leaf}=[draw=hiddendraw,
    rounded corners, minimum height=1em,
    fill=hidden-orange!40,text opacity=1, align=center,
    fill opacity=.5,  text=black,align=left,font=\tiny,
    inner xsep=3pt,
    inner ysep=1pt,
    ]
\begin{figure}[h]
\centering
\begin{forest}
  for tree={
  forked edges,
  grow=east,
  reversed=true,
  anchor=base west,
  parent anchor=east,
  child anchor=west,
  base=middle,
  font=\scriptsize,
  rectangle,
  draw=hiddendraw,
  rounded corners,align=left,
  minimum width=2em,
    s sep=5pt,
    inner xsep=3pt,
    inner ysep=1pt,
  },
  where level=1{text width=4.5em}{},
  where level=2{text width=6em,font=\scriptsize}{},
  where level=3{font=\tiny}{},
  where level=4{font=\tiny}{},
  where level=5{font=\tiny}{},
  [Out-of-category Challenge,rotate=90,anchor=north,edge=hiddendraw
    [Discriminant\\ (\S\ref{solution:discriminant}),edge=hiddendraw,text width=6em
        [Sampling, text width=5.8em, edge=hiddendraw
            [{Monte Carlo Dropout~\cite{miller2018dropout, miller2019evaluating}},leaf,text width=7.8em, edge=hiddendraw]
        ]
        [Outlier Exposure, text width=5.8em, edge=hiddendraw
            [{VOS~\cite{du2021vos}, OpenDet~\cite{han2022expanding}},leaf,text width=7.8em, edge=hiddendraw]
        ]
        [Unknown-aware, text width=5.8em, edge=hiddendraw
            [{Joseph \etal~\cite{joseph2021towards}, STUD~\cite{du2022unknown},\\Liu \etal~\cite{liu2022open}, ProposalCLIP~\cite{shi2022proposalclip},\\OW-DETR~\cite{gupta2022ow}, UC-OWOD~\cite{wu2022uc},\\Su \etal~\cite{su2023hsic}},leaf,text width=7.8em, edge=hiddendraw]
        ]
    ]
    [Side Information\\ (\S\ref{solution:side}),edge=hiddendraw,text width=6em
     [Attribute, text width=5.8em, edge=hiddendraw
        [{Zhu \etal~\cite{zhu2019zero}, Mao \etal~\cite{mao2020zero}},leaf,text width=7.8em, edge=hiddendraw]
     ]
     [Visual-text Align-\\ment, text width=5.8em, edge=hiddendraw
        [{SB~\cite{bansal2018zero}, Rahman \etal~\cite{rahman2020improved},\\Li \etal~\cite{li2019zero}, Yan \etal~\cite{yan2020semantics},\\Nie \etal~\cite{nie2022node}, Yan \etal~\cite{yan2022semantics}}
        ,leaf,text width=7.8em, edge=hiddendraw]
     ]
     [Synthesis, text width=5.8em, edge=hiddendraw
        [{Huang \etal~\cite{huang2022robust}, Li \etal~\cite{li2023zero},\\Gupta \etal~\cite{gupta2023generative}},leaf,text width=7.8em, edge=hiddendraw]]
     ]
    [Arbitrary Informa-\\tion (\S\ref{solution:arbitrary}), edge=hiddendraw,text width=6em
      [Multi-modal\\Alignment, text width=5.8em, edge=hiddendraw
        [{OVR-CNN~\cite{zareian2021open}, ViLD~\cite{guopen},\\Ma \etal~\cite{ma2022open}, DetPro~\cite{du2022learning},\\F-VLM~\cite{kuo2023f}, CORA~\cite{wu2023cora}},leaf,text width=7.8em, edge=hiddendraw]]
    [Pseudo-labeling, text width=5.8em, edge=hiddendraw
      [{Gao \etal~\cite{gao2022open}, PromptDet~\cite{feng2022promptdet}},leaf,text width=7.8em, edge=hiddendraw]]
    [Pre-training, text width=5.8em, edge=hiddendraw
      [{GLIP~\cite{li2022grounded}, RegionCLIP~\cite{zhong2022regionclip},\\ UniDetector~\cite{wang2023detecting}, Ro-ViT~\cite{kim2023region},\\DetCLIP~\cite{yao2022detclip}, DetCLIPv2~\cite{yao2023detclipv2}},leaf,text width=7.8em, edge=hiddendraw]]
    ]
  ]
\end{forest}
\caption{Methods for addressing out-of-category challenges.}
\label{out-of-category_solution}
\end{figure}

%% file: sections/6-malicious_data.tex
\section{Robust Learning Challenge}
\label{sec:challenge3}

This section focuses on the robust learning of object detectors in defending against malicious data such as adversarial attacks.

Existing malicious data in open environments focuses on two forms backdoor attacks and adversarial attacks. \emph{Backdoor samples}, embed specific triggers in the model during training primarily via data poisoning, leading to incorrect predictions during testing. \emph{Adversarial samples}, created during the testing phase, directly add perturbations to mislead the model. 

Robustness in the context of deep object detectors refers to their capability to maintain stability and accuracy despite malicious interference. Consequently, the modeling process of a deep object detector must thoroughly address these two types of threats, implementing strategies to improve robustness against adversarial examples and backdoor samples with triggers.  For critical real-world applications like autonomous driving, face recognition, and industrial inspection, enhancing the robustness of these detectors, particularly against malicious data, is highly important. We define the open loss function for robust learning as follows:

\begin{definition}[Robust Learning on Malicious Data] In this defense scenario, the objective is for the detector $f_{\bm{\theta}}$ to accurately identify annotations of malicious data by using the adversarial training or additional modules ($g_{\bm{\theta}^1}$ or $d_{\bm{\theta}^2}$), as outlined below: 
\begin{equation}
\begin{aligned}
% \min \sum_{i=1}^{N}[ \mathcal{L}_{det}(f_{\bm{\theta}}(\bm{x}_i, \bm{y}_i, \bm{b}_i)+ \lambda \mathcal{L}_{det}(f_{\bm{\theta}}(\hat{\bm{x}}_i, \bm{y}_i, \bm{b}_i)]. 
\mathcal{L}_{\text{o}}=\mathcal{L}_{det}[\textcolor{red}{\overbrace{\textcolor{red}{f_{\bm{\theta}}} \circ \textcolor{blue}{\underbrace{g_{\bm{\theta}^{1}} (d_{\bm{\theta}^{2}}(\textcolor{red}{\mathcal{X}+\bm{\delta}}))}_{\textcolor{blue}{\text{Model Robust Inference}}}}, \mathcal{Y}}^{\text{Adversarial Training Paradigm}}}],
\label{robust learning}
\end{aligned}
\end{equation}
where detection loss $\mathcal{L}_{det}$ includes both classification loss $\mathcal{L}_{cls}$ and localization loss $\mathcal{L}_{loc}$. A malicious image $\mathcal{X}+\bm{\delta}$ originates from a clean sample $\mathcal{X}$, modified to trick the model. These defenses create adversarial perturbations $\bm{\delta}$ to mimic malicious samples in the test phase $t+1$ and maintain accurate annotations $\mathcal{Y}=\mathcal{Y}^{1:t}$. 
\end{definition}

We will succinctly discuss the primary difficulties faced by object detectors and dissect Eq.~\eqref{robust learning} into two segments (Fig.~\ref{fig:malicious_data}) to align with the proposed solutions. Despite the availability of numerous defenses against backdoor attacks aimed at identifying poisoned models~\cite{lin2020composite, zhou2021multi}, these defenses are predominantly tailored for classifiers and not detectors, which falls outside the purview of this paper. Therefore, this section focuses on defense mechanisms against adversarial attacks: adversarial training paradigm and model robust inference.

\subsection{Core Difficulties}
%For malicious defense in object detection, the core difficulties can be summarized as follows.

\ding{182} \textbf{Diversity and complexity of adversarial attacks.} These attacks~\cite{liang2020efficient, wang2023diversifying, he2023generating, lou2024hide, muxue2023adversarial, he2023sa, liu2023improving, dong2023face, jia2020adv} make the adversarial defense of deep detectors extremely difficult. In the digital world, \emph{pixel attacks} focus on misleading the object detector through small perturbations~\cite{xie2017adversarial}, revealing the vulnerability and security risk of the detection system in the face of subtle changes including different detector types~\cite{liao2020category, im2022adversarial}, shared structure attacks~\cite{wei2018transferable, wang2021daedalus, shapira2023phantom}, and black-box attacks~\cite{yin2022adc, cai2022context, cai2022zero, nezami2021pick, xia2022ssmi, liang2021generate, liang2022large, liang2022parallel, huang2023t}. Besides, \emph{planar patch attack} circumvents defense algorithms against perturbation restrictions by generating neural network-sensitive rectangular patch attacks, such as PS-GAN~\cite{liu2019perceptual}, DPatch~\cite{liu2018dpatch}, and other patch shapes~\cite{zhang2021adversarial, zhao2020object, zhu2021you, bao2020sparse, wu2020dpattack, liu2020bias, pavlitskaya2022feasibility,liu2023x}. In the physical world, \emph{non-planar patch attacks} address the challenges of environmental transformations and material deformations by applying antagonistic patches to non-rigid or non-planar objects to confuse detectors~\cite{evtimov2017robust}. Such attacks use a series of transformation matrices, \ie, physical environment simulations~\cite{huang2020universal}, Thin Plate Spline (TPS) interpolation~\cite{wu2020making, xu2020adversarial}, nasted-AE~\cite{zhao2019seeing}, and 3D mesh generation~\cite{maesumi2021learning} to accommodate deformation. Besides, the \emph{adversarial camouflage} confuses object detectors in multiple viewpoints by generating full-coverage textures~\cite{chen2019shapeshifter,athalye2018synthesizing,wu2020physical,duan2021learning,wang2021dual,liu2020spatiotemporal} in real-world scenarios such as autonomous driving.
Thus, the ever-evolving nature of adversarial attack techniques necessitates a continuous adaptation and updating of defense solutions to effectively counter different types of unseen attacks. 

\ding{183} \textbf{Backdoor attacks are stealthy}. These attacks~\cite{chen2023universal, liang2024vl, liang2024poisoned, liu2023does, liu2023pre, liang2023badclip} embed specific triggers during the training phase with poisoned samples that are typically undetectable and invisible under normal conditions, yet, they can lead to erroneous model predictions under certain circumstances. Research in this domain spans both digital~\cite{lin2020composite, zhou2021multi} and physical attacks~\cite{ma2022dangerous}. These attacks in object detection include various types~\cite{chan2023baddet, luo2022untargeted}, such as object generation, regional misclassification, global misclassification, and object disappearance. Additionally, clean-annotation backdoors do not necessitate label modification~\cite{ma2022dangerous}. The inherent stealthiness of these backdoor triggers renders their detection~\cite{wang2022adaptive} and elimination exceedingly challenging.

\ding{184} \textbf{Defenses necessitate a balance among performance, robustness, and computational resources.} Firstly, \emph{balancing clean accuracy and robustness.} Enhancing robustness through complex data processing or additional defense layers can potentially decrease a model's clean prediction accuracy or increase inference time. It is essential to find an optimal balance where robustness enhancements do not significantly impair detection performance on clean samples. Secondly, \emph{considering data and resource constraints.} Training robust models require large and diverse datasets, which must adequately simulate potential malicious data that the model might encounter in the testing phase at time $t+1$. However, generating and sampling such extensive data sets demand significant resources and computational power, which needs to be factored into the overall solution.

\input{tables/solution/malicious_solution}

\subsection{Solutions}
We categorize the existing mainstream solutions for robust learning challenges into three distinct groups, as shown in Fig.~\ref{malicious_data_solution}. 
% To bolster the robustness of object detectors against malicious threats, current defense mechanisms can be categorized into two primary approaches: (1) Model structure design, where defenders enhance the robustness by making direct modifications to the detector architectures (Fig.~\ref{fig:malicous-data}(a)). (2) Adversarial training paradigm, where defenders implement novel training paradigms to train robust model parameters (Fig.~\ref{fig:malicous-data}(b)). This paradigm views adversarial attacks and defenses through the lens of game theory, treating the interaction between them as a strategic game.

% Despite the availability of numerous defenses against backdoor attacks aimed at identifying poisoned models~\cite{lin2020composite, zhou2021multi}, these defenses are predominantly tailored for classifiers and not detectors, which falls outside the purview of this paper. Therefore, this section focuses on defense mechanisms against adversarial attacks, which can be categorized in Fig.~\ref{fig:malicious_data} according to training and inference phases as shown in : adversarial training paradigm and model robust inference.
\subsubsection{Adversarial Training Paradigm}\label{6.3.1}
The adversarial training paradigm~\cite{liu2023exploring,sun2023improving, 10471619,jia2023improving,jia2022adversarial,jia2023revisiting} is widely recognized as a highly effective way of adversarial defense. This method involves the deliberate generation of adversarial examples during the training phase in Fig.~\ref{malicious_data_solution} (a), which are then used to train the deep network directly. The red part of Eq.~\eqref{robust learning} reveals the training process by which the model categorizes adversarial examples into correct labels.

\textbf{Adaptive strategies} are crucial in the algorithm development process, and researchers need to add novel insights when transferring existing adversarial training from image classification \cite{liu2021training,zhang2021interpreting,liu2023towards} to object detection. Choi \etal~\cite{im2022adversarial} conduct a study of adversarial weaknesses related to objects in the YOLO detector. They proposed a novel method of adversarial training that takes into account object-awareness. MTD~\cite{zhang2019towards} improves the robustness of the object detection model with the incorporation of various attack sources in the process of adversarial training. Xu \etal~\cite{xu2021using} propose that the utilization of feature alignment in the intermediate layer can enhance the detection performance for adversarial images. Li \etal~\cite{li2023importance} examine the significance of strengthening the robustness of the backbone network in the context of the object detector. Det-AdvProp~\cite{chen2021robust} applies the related ideas of AdvProp~\cite{xie2020adversarial} for the object detection task, and dynamically selects adversarial samples from different decision branches according to the model changes to strengthen the adversarial training. The UDFA~\cite{xu2022robust} presents a novel adversarial training framework that is built upon knowledge distillation. The research thoroughly investigates the interplay between self-knowledge distillation and adversarial training.

\textbf{Balancing multi-objectives} about model robustness and accuracy remains a central theme in the adversarial training process of object detectors, \ie, adversarial training leads to performance degradation of clean images~\cite{ wu2021wider}. Dong \etal~\cite{dong2022adversarially} demonstrate the presence of the above tradeoff in adversarial training for object detection. To address this concern, the authors propose the introduction of a novel framework called RobustDet. This framework is built upon the concept of adversarial perceptual convolution, which disentangles gradients for model learning on clean and adversarial images. AIAD~\cite{chengadversarial} considers the robustness of the detector under different iterations of adversarial examples and enhances the robustness of the detector in the face of multi-intensity attacks by utilizing the intensity-aware discriminator and the adversarial intensity information.

\begin{figure*}[t]
\centering
\includegraphics[width=0.96\textwidth]{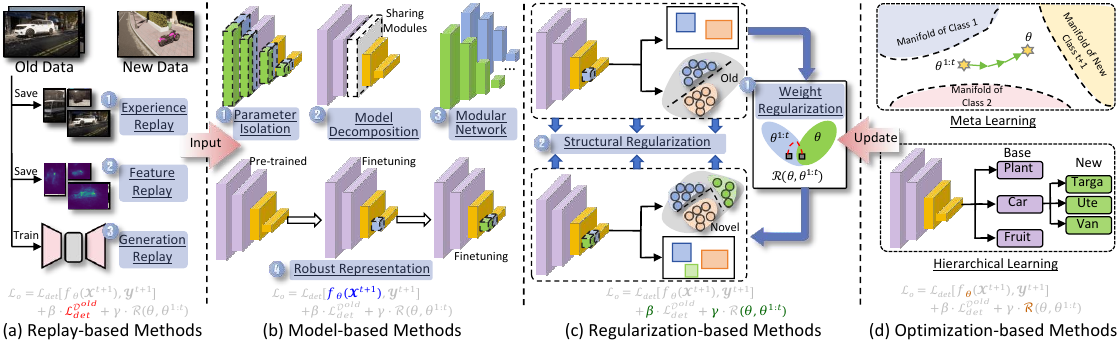}
% \vspace{-8pt}
\caption{Illustration of the solution categories for incremental learning challenges.}
\label{fig:incrimental}
% \vspace{-0.18in}
\end{figure*}

\textbf{Heightened time complexity} in the training process is another significant limitation. The process of adversarial training is often observed to exhibit a significantly slower pace compared to traditional training. This can be attributed to the generation of adversarial examples at different steps throughout the training process. Extensive research has been conducted on fast adversarial training for image classification, as represented by the works of Wu \etal~\cite{wu2022towards} and Wong \etal~\cite{wong2020fast}. Similarly, Chen \etal~\cite{chen2021class} study efficient adversarial training algorithms under the context of object detection and propose generating a universal adversarial perturbation to simultaneously attack all the occurring objects in the image.

\subsubsection{Model Robust Inference}\label{6.3.2}

In the robust inference phase of the detection model, as shown in Fig.~\ref{malicious_data_solution} (b), the defender can ensure robustness in the inference phase by removing malicious effects of perturbations or detecting and blocking adversarial examples. The $g_{\bm{\theta}^{1}}$ and $d_{\bm{\theta}^{2}}$ in the blue part of Eq.~\eqref{robust learning} denote the denoising method as well as the additional network modules used for defense, respectively.

\textbf{Noise removal} based methods can be designed at different stages of the model to destroy specific structures of perturbations to mitigate adversarial attacks. At the input stage, JPG~\cite{dziugaite2016study} processes the input samples to mitigate the attack of the adversarial examples through JPG compression technique; APM~\cite{chiang2021adversarial} learns a data pre-processing network based on a masking technique to remove the noise; The LGS method~\cite{naseer2019local} employs gradient normalization of the estimated region, aiming to mitigate the impact of high activation regions in an image that are induced by adversarial perturbations. SP~\cite{xu2017feature} reduces the available search space for the attack by combining different vectors in the original feature space into a smaller space. At the output stage, Chiang \etal~\cite{chiang2020detection} extend the robustness techniques in classification tasks to object detection by utilizing mean smoothing techniques to improve the accuracy and robustness of the detected objects.

Training \textbf{additional modules} can enhance the model's ability to recognize and protect against adversarial patches in images. DetectorGuard~\cite{xiang2021detectorguard} employs small receptive fields and a robust classification header for secure feature aggregation, determining the presence of objects. Han \etal~\cite{han2022real} note that adversarial patches show a highly localized surface feature importance in small regions, often leading to unreliable predictions. Building on this insight, they developed Themi, a combined hardware and software system designed to effectively remove adversarial regions, thereby enhancing defense performance in video target detection tasks. Similarly, SAC~\cite{liu2022segment} utilizes a patch segmenter trained to produce a patch mask, which aids in localizing and eliminating adversarial patches from images. Z-Mask~\cite{rossolini2023defending} employs a comparable approach. More detailed classification scales can be found in \emph{Supplementary Materials D}.

%% file: tables/solution/malicious_solution.tex
% \begin{table}[!t]
% \caption{Methods for addressing malicious data challenges.}
% \centering
% \footnotesize
% \renewcommand{\arraystretch}{1.1}

% \resizebox{\columnwidth}{!}{%
% \begin{tabular}{m{0.15\columnwidth}|m{0.31\columnwidth}|m{0.54\columnwidth}}

% \hline
% \rowcolor{gray!10} Category & Main Focus & Representative Method  \\ 
% \hline
% \multirow{3}{=}[-1.2em]{Adversarial Training} & Adaptive Strategies & Choi \etal~\cite{im2022adversarial}, MTD~\cite{zhang2019towards}, Xu \etal~\cite{xu2021using}, Li \etal~\cite{li2023importance}, Det-AdvProp~\cite{chen2021robust}, UDFA~\cite{xu2022robust} \\ 
% \cline{2-3}
%  & Balancing Multi Objectives & Dong \etal~\cite{dong2022adversarially}, AIAD~\cite{chengadversarial} \\ 
% \cline{2-3}
%  & Heightened Time Complexity & Wu \etal~\cite{wu2022towards}, Wong \etal~\cite{wong2020fast}, Chen \etal~\cite{chen2021class} \\ 
% \hline
% \multirow{2}{=}[-0.3em]{Model Robust Inference} & Noise Removal & JPG~\cite{dziugaite2016study}, APM~\cite{chiang2021adversarial}, LGS~\cite{naseer2019local}, SP ~\cite{xu2017feature}, Chiang \etal~\cite{chiang2020detection} \\ 
% \cline{2-3}
%  & Additional Modules & DetectorGuard~\cite{xiang2021detectorguard}, Han \etal~\cite{han2022real}, SAC~\cite{liu2022segment}, Z-Mask~\cite{rossolini2023defending} \\
% \hline

% \end{tabular}%
% }
% \end{table}

\tikzstyle{leaf}=[draw=hiddendraw,
    rounded corners, minimum height=1em,
    fill=hidden-orange!40,text opacity=1, align=center,
    fill opacity=.5,  text=black,align=left,font=\tiny,
    inner xsep=3pt,
    inner ysep=1pt,
    ]
\begin{figure}[ht]
\centering
\begin{forest}
  for tree={
  forked edges,
  grow=east,
  reversed=true,
  anchor=base west,
  parent anchor=east,
  child anchor=west,
  base=middle,
  font=\scriptsize,
  rectangle,
  draw=hiddendraw,
  rounded corners,align=left,
  minimum width=2em,
    s sep=5pt,
    inner xsep=3pt,
    inner ysep=1pt,
    text width=9em,
  },
  where level=1{text width=4.5em}{},
  where level=2{text width=5em,font=\scriptsize}{},
  where level=3{text width=9em,font=\tiny}{},
  where level=4{font=\tiny}{},
  where level=5{font=\tiny}{},
  [Robust Learning Challenge,rotate=90,anchor=north,edge=hiddendraw
    [Adversarial \\ Training \\ Paradigm \\(\S\ref{6.3.1})
        [Adaptive \\Strategies
            [{Choi \etal~\cite{im2022adversarial}, MTD~\cite{zhang2019towards}, Xu \etal\\ \cite{xu2021using}, Li \etal~\cite{li2023importance}, Det-AdvProp\\ \cite{chen2021robust}, UDFA~\cite{xu2022robust}}]
        ]
        [Balancing Mu-\\lti Objectives
            [{Dong \etal~\cite{dong2022adversarially}, AIAD~\cite{chengadversarial}}]
        ]
        [Heightened Ti-\\me Complexity
            [{Chen \etal~\cite{chen2021class}}]
        ]
    ]
    [Model Robust \\Inference \\(\S\ref{6.3.2})
     [Noise Removal
        [{JPG~\cite{dziugaite2016study}, APM~\cite{chiang2021adversarial}, LGS~\cite{naseer2019local}, SP \\ \cite{xu2017feature}, Chiang \etal~\cite{chiang2020detection}}]
     ]
     [Additional Mo-\\dules
        [{DetectorGuard~\cite{xiang2021detectorguard}, Han \etal~\cite{han2022real}, \\SAC~\cite{liu2022segment}, Z-Mask~\cite{rossolini2023defending}}
        ]
     ]
     % [Feature Disent-\\anglement
     %    [{Su \etal~\cite{SuWZTCQW20}, Wu \etal ~\cite{WuLHZ021}, Wu \etal \\ \cite{WuHZY22, WuD22}}]]
     % ]
  ]
]
\end{forest}
\caption{Solutions for robust learning challenges.}
\label{malicious_data_solution}
% \vspace{-20pt}
\end{figure}

%% file: sections/7-incremental.tex
\section{Incremental Learning Challenge}
\label{sec:challenge4}
% This section provides an in-depth discussion of object detectors for the class incremental challenges. We also provide a comprehensive overview of current solutions.

% \subsection{Problem Definition}
% Plasticity refers to the adaptability and flexibility of an object detector system in learning new information. Since mainstream object detectors trained following the traditional machine learning paradigm are viewed as \emph{static models}, they are not able to cope well with new categories of detection demands. Although deep detector owners can add new categories to the original training dataset to re-train the model, however, resource limitations as well as privacy concerns can limit the above ideas to a great extent. Therefore, model designers must focus on enhancing the plasticity of these systems so that target detection systems remain relevant and effective in dynamic environments where new detection categories and scenarios continually arise.
This section provides an in-depth discussion of object detectors for incremental learning challenges. Since mainstream object detectors trained following the traditional machine learning paradigm are viewed as \emph{static models}, they are not able to cope well with new categories of detection demands. Although deep detector owners can add new categories to the original training dataset to re-train the model, resource limitations as well as privacy concerns can limit the above ideas to a great extent.

Plasticity refers to the adaptability and flexibility of an object detector system in learning new objects. Therefore, model designers must focus on enhancing the plasticity of these systems so that object detection systems remain relevant and effective in dynamic environments where new detection categories and scenarios continually arise. In incremental learning, the open loss function of the model can be defined as follows:
\begin{definition}[Incremental Learning on New Objects] A continuous flow of information in the real world can be represented by the time series $\{1,2,... ,t+1\}$, with $\mathcal{D}^{1:t}$ and $\mathcal{D}^{t+1}=\{\mathcal{X}^{t+1}, \mathcal{Y}^{t+1}\}$ denoting the dataset sampled until moment $t$ and the new dataset at moment $t+1$, respectively. Incremental learning deals with changing detection datasets and learns new objects generated at time $t+1$ while retaining old knowledge $\mathcal{D}^{old} \subset \mathcal{D}^{1:t}$. The detection model with old parameter $\bm{\theta}^{1:t}$ continues to be trained on the new object dataset until it converges, yielding the new parameter $\bm{\theta}$ as formulated:
% A continuous flow of information in the real world can be represented by the time series $\{1,2,... ,t+1\}$, with $\mathcal{D}^{1:t}$ and $\mathcal{D}^{t+1}$ denoting the dataset sampled until moment $t$ and the new dataset at moment $t+1$, respectively. Class incremental object detection deals with changing detection datasets and learns new objects generated at time $t+1$ while retaining old knowledge. The detection model with parameter $\bm{\theta}^{1:t}$ continues to be trained on the new object dataset until it converges, yielding the new parameter $\bm{\theta}$. The objective function for this incremental problem is formulated as:
\begin{equation}
\label{ciod}
    \mathcal{L}_{\text{o}} = \mathcal{L}_{det}[\textcolor{darkamber}{\underbrace{\textcolor{blue}{\overbrace{f_{\textcolor{darkamber}{\bm{\theta}}}(\mathcal{X}^{t+1})}^{\text{Model}}}\textcolor{black}{,\mathcal{Y}^{t+1}] +} \textcolor{deepgreen}{\overbrace{\beta  \cdot \textcolor{red}{\overbrace{\mathcal{L}_{det}^{\mathcal{D}^{\text{old}}}}^{\text{Replay}}} + \textcolor{deepgreen}{\gamma} \cdot \textcolor{deepgreen}{\textcolor{darkamber}{\mathcal{R}}(\bm{\theta}, \bm{\theta}^{1:t})}}^{\textcolor{deepgreen}{\text{Regularization}}}}}_{\textcolor{darkamber}{\text{Optimization}}} },
\end{equation}
where $\mathcal{L}_{det}^{\mathcal{D}^{\text{old}}}$ represents the detection loss on a small set of old datasets. $\mathcal{R}$ is a regularization term to stabilize the model parameter. The weight coefficients $\lambda$ and $\beta$ are used to balance the coefficients for the loss terms.
\end{definition}
By optimizing the Eq.~\eqref{ciod}, the model can learn incrementally by absorbing new knowledge while retaining as much knowledge as possible about the old data. We decompose the components of Eq.~\eqref{ciod} into four parts according to different colors, corresponding to four solution strategies (Fig.~\ref{fig:incrimental}). We will elucidate the core challenges facing incremental learning and outline common solutions in detail.

\subsection{Core Difficulties}
%The core difficulties of incremental object detection can be summarized in the following two main areas.

\ding{182} \textbf{Stability-plasticity trade-off.} In this scenario, the detector must balance the learning of rapid adaptation to new data (plasticity) with retaining accuracy on existing data (stability). It is crucial for the detector to efficiently learn new categories without compromising its ability to recognize previously established categories. This is shown in the trade-off between $\mathcal{L}_{det}^{\mathcal{D}^{t+1}}$ and $\mathcal{L}_{det}^{\mathcal{D}^{1:t}}$ in Eq.~\eqref{ciod}. If the detector leans too much towards plasticity, it might forget older categories (catastrophic forgetting). Conversely, too much stability can hinder its ability to learn new categories effectively.

\ding{183} \textbf{Conflict between old and new objects.} Introduction of new category objects can lead to incorrect labeling in the old dataset. These new objects are often treated as the background in the old dataset and need to be recognized as new categories afterward. The model needs to adjust its understanding of both old and new objects to maintain accurate detection. This is shown as regular term $\mathcal{R}(\bm{\theta}, \bm{\theta}^{1:t})$ between new parameters and old parameters in Eq.~\eqref{ciod}. This conflict can lead to inconsistencies in the model performance, particularly in recognizing and differentiating between old and new category objects.

\input{tables/solution/incrimental_solution}

\subsection{Solutions}
Fig.~\ref{incremental_data_solution} summarizes four solutions for incremental learning.
% \input{tables/solutions for increment data}·
% Inspired by~\cite{wang2023comprehensive}, we categorize existing solutions that concentrate on class incremental object detection into four main groups as shown in Fig.~\ref{fig:class-incremental}.

% \begin{figure}[t!]
% \centering
% \subfigure[Replay-based method]{
% \includegraphics[width=0.99\linewidth]{images/incrimental/incrimen_1.pdf}
% }

% \subfigure[Model-based method]{
% \includegraphics[width=0.99\linewidth]{images/incrimental/incrimen_2.pdf}
% }

% \subfigure[Regularization-based method]{
% \includegraphics[width=0.99\linewidth]{images/incrimental/incrimen_3.pdf}
% }

% \subfigure[Optimization-based method]{
% \includegraphics[width=0.99\linewidth]{images/incrimental/incrimen_4.pdf}
% }
% \caption{Solutions for class incremental challenge.}
% \label{fig:class-incremental}
% %\includegraphics[width=0.48\textwidth]{images/OOD/OOD_2.pdf}
%   %\caption{Data manipulation-based methods.}
%   %\label{Data manipulation-based methods}
% \end{figure}

\subsubsection{Replay-based Methods}\label{7.3.1}
Replay-based methods~\cite{van2022three, zhu2021class, agarwal2022semantics}, inspired by the learning process from the hippocampus to neocortex representations in humans~\cite{mcclelland1995there}, relieve catastrophic forgetting when learning new objects~\cite{castro2018end, hayes2019memory, rebuffi2017icarl, wu2019large}, which is achieved by preserving relevant information about samples, characteristics, and distributions related to previous training objectives, as shown in Fig.~\ref{fig:incrimental} (a). The process can be represented as the red part in Eq.~\eqref{ciod}.

\textbf{Experience replay} emulates the capacity of biological systems to recall and revisit prior events. The ORE~\cite{joseph2021towards} employs a strategy of storing a set of old objects balanced in capacity. To address the insufficient storage of old samples, the ER~\cite{shieh2020continual}, ABR~\cite{yuyang2023augmented} and OSR~\cite{yang2023one} algorithm employ rich data augmentation~\cite{yun2019cutmix, zhang2017mixup, bochkovskiy2020yolov4} to enhance the previous training samples. The RODEO~\cite{acharya2020rodeo} and Shieh \etal~\cite{shieh2022utilizing} compress the feature representations of previous samples obtained from CNNs. CL-DETR~\cite{liu2023continual} improves the ER algorithm towards transformer-based detectors~\cite{carion2020end, zhu2020deformable}. On the contrary, \textbf{ feature replay} has the unique advantage of maintaining the privacy and efficiency of objects through the memory buffer. The CIFRCN-NP~\cite{hao2019end} proposes to assign labels to the region suggestion frames using the nearest prototype (NP) classifier~\cite{biehl2013distance}, which makes it easy to add new categories.  iMTFA~\cite{ganea2021incremental} solves the memory overhead problem by storing embedding vectors instead of images. In addition, \textbf{generation replay} trains additional generative models to generate replayed old category data. Based on the above idea, RT-Net~\cite{cui2023rt} uses conditional GAN networks~\cite{creswell2018generative} to replay the old category features after RoIs extraction to address the old and new data imbalance. 

\subsubsection{Model-based Methods}\label{7.3.2}

From the perspective of model architecture, the reason for catastrophic forgetting is interference in the model parameter space that arises from the simultaneous learning of old and new tasks~\cite{wang2023comprehensive}. Therefore, developing a parameter space structure (Fig.~\ref{fig:incrimental} (b)) that helps to strengthen the generalized representation applicable to new objects can be viewed as a type of solution, which can be represented as a treatment of model features (Eq.~\eqref{ciod} in blue).

\textbf{Parameter isolation} allows the model designer to allocate distinct subspaces for individual tasks, thereby mitigating potential conflicts. MMN~\cite{li2018incremental} employs weight-based memory neurons to retain the original class knowledge across all feature layers and updates the remaining neurons to acquire novel object knowledge. DIODE~\cite{peng2023diode} introduces a scalable architecture at the classification layer to relieve the network parameter capacity, which is inadequate. Besides, \textbf{model decomposition} offers a way to mitigate parameter space conflicts by dividing the object detection model architecture into two distinct modules, \ie, a task-sharing fundamental module and a task-specific incremental module. For example, LSTD~\cite{chen2018lstd} introduces incremental object detection in few-sample scenarios, focusing on knowledge transfer and model generalization in data-constrained environments. Kang \etal~\cite{kang2019few} first learns generalizable features using the Meta feature extractor and then generates activation coefficients using few-shot new samples. Other similar methods encompass iFSOD~\cite{cheng2021meta} and Sylph~\cite{yin2022sylph}. Based on the strong adaptability of the CenterNet detector structure, ONCE~\cite{perez2020incremental} separates the detectors into class-general and class-specific components.  DualFusion~\cite{tambwekar2021few} introduces an incremental framework that builds upon the FewX model~\cite{fan2020few} and suggests the utilization of a fusion network for classifying both old and novel objects. Another line of solution called \textbf{Modular network} suggests the adoption of parallel sub-networks or sub-modules to acquire incremental categories. In particular, context-Transformer~\cite{yang2020context} introduces an incremental module employing contextual associations that can be seamlessly integrated into SSD models.  Expert Detector~\cite{zhang2021incremental} employs an expert strategy~\cite{aljundi2017expert,collier2020routing,jacobs1991adaptive}  to train each task individually. OW-DETR~\cite{gupta2022ow} proposes a novel classification branch to differentiate between known and unknown targets. Motivated by the selection of fundamental architecture~\cite{srivastava2014dropout}, Rosetta~\cite{yang2022continual} determines the appropriate model branching for a given task by utilizing a collection of stored gate lists and feature prototypes.

In addition, \textbf{Representation-based} methods~\cite{shon2022dlcft,zhang2020side} are of significant importance in addressing the issue of model forgetting of previously learned objects and facilitating the learning of new objects. Rahman \etal~\cite{rahman2020any} associate visual features with a semantic space and refines the feature representation through the utilization of semantic representations of new classes during the pre-training phase~\cite{ostapenko2022foundational,ramasesh2021effect}. Incremental-DETR~\cite{dong2023incremental} utilizes a self-supervised learning strategy~\cite{purushwalkam2022challenges,gallardo2021self} to acquire comprehensive base class data. Subsequently, the classification layer is fine-tuned throughout the incremental phase to improve incremental learning.

\subsubsection{Regularization-based Methods}\label{7.3.3}

Regularization-based methods mitigate catastrophic forgetting by adding rules (Fig.~\ref{fig:incrimental} (c)) to detectors, which can be classified as weighted regularization and structural regularization, as shown in the green part of Eq.~\eqref{ciod}. 

\textbf{Weight regularization} applies selective regularization to network parameters, enhancing the model flexibility to novel objects by constraining the undue bias. For example, BRS~\cite{cui2022balanced} effectively addresses parameter overfitting by employing a task weighting strategy that facilitates the aggregation of diverse task instances throughout the iterative classification process. DIODE~\cite{peng2023diode} experimentally explores the performance of many parameter regularization methods on the classification branch, including EWC\cite{kirkpatrick2017overcoming}, Online EWC\cite{schwarz2018progress}, and MAS\cite{aljundi2018memory}. IncDet~\cite{liu2020incdet} proposes Huber regularization replace Elastic Weight Consolidation~\cite{kirkpatrick2017overcoming} to overcome the gradient explosion during training. Moreover, iFS-RCNN~\cite{nguyen2022ifs} achieves better accuracy under few-shot samples by learning the weight distribution of the novel classification head and penalizing the prediction error of high-certainty bounding boxes. 

Besides, \textbf{structural regularization} aims to impose constraints on both the intermediate features of the model and its final output. Specifically, ILOD~\cite{shmelkov2017incremental} employs the old detector as a teacher model, while using the incremental detectors to distill suggestion regions and category losses from the old detector. Discussions of knowledge distillation~\cite{liang2023exploring} usually focus on three main areas: the input level, the feature level, and the output level of the teacher model. At the \emph{input level}, Dong \etal~\cite{dong2021bridging} utilize a blind sampling strategy to extract heatmaps-based instance knowledge from a lot of irrelevant images. LEAST~\cite{li2021class} proposes a clustering-based sample selection with the expectation of capturing the base class distribution using a small number of images. At the \emph{feature-level}, SID~\cite{peng2021sid} selectively distills knowledge from feature layers, and instance interrelationships; MVCD~\cite{yang2022multi} distills discriminative features from three views (channel-wise, point-wise and instance-wise); PPAS~\cite{zhou2020lifelong}, Dong \etal~\cite{dong2021bridging} and Faster ILOD~\cite{peng2020faster} discuss the impact of knowledge distillation on the RPN network. To mitigate old and new object conflicts, PPAS~\cite{zhou2020lifelong} propose a pseudo-positive-aware sampling strategy for calculating RoI regions that favor the preservation of old knowledge. At the \emph{output-layer}, Chen \etal~\cite{chen2019new} propose a hard loss for the foreground suggestion box and a soft loss for the background region; ERD~\cite{feng2022overcoming} uses an output-response-based distillation framework focusing on learning the incremental output; Verwimp \etal~\cite{verwimp2022re} analyze the over-regularization problem of the MSE loss function and replaces with Huber loss.

\subsubsection{Optimization-based Methods}\label{7.3.4}

The last type is the optimization-based method (Fig.~\ref{fig:incrimental} (d)), which directly manipulates the model optimization process~\cite{li2022learning} to achieve a trade-off between stability and plasticity and ultimately affects the brown part in Eq.~\eqref{ciod}. Based on the stability-plasticity theory~\cite{mirzadeh2020understanding}, Teo \etal~\cite{li2021towards} propose a stable moment matching structure based on decoupling the base class and new class representations, limiting the excessive variation of parameters. Joseph \etal~\cite{joseph2021incremental} introduce a novel approach based on meta-learning~\cite{vanschoren2019meta, javed2019meta,riemer2018learning,rajasegaran2020itaml} to enhance the model gradient updating process. To address the sluggish convergence encountered by stochastic gradient descent, HDA~\cite{she2022fast} proposes a gradient optimization approach based on the hierarchical detection algorithm. More detailed classification scales can be found in \emph{Supplementary Materials E}.

%% file: tables/solution/incrimental_solution.tex
\tikzstyle{leaf}=[draw=hiddendraw,
    rounded corners, minimum height=1em,
    fill=hidden-orange!40,text opacity=1, align=center,
    fill opacity=.5,  text=black,align=left,font=\tiny,
    inner xsep=3pt,
    inner ysep=1pt,
    ]
\begin{figure}[ht]
\centering
\begin{forest}
  for tree={
  forked edges,
  grow=east,
  reversed=true,
  anchor=base west,
  parent anchor=east,
  child anchor=west,
  base=middle,
  font=\scriptsize,
  rectangle,
  draw=hiddendraw,
  rounded corners,align=left,
  minimum width=2em,
    s sep=5pt,
    inner xsep=3pt,
    inner ysep=1pt,
  },
  where level=1{text width=4.5em}{},
  where level=2{font=\scriptsize}{},
  where level=3{font=\tiny}{},
  where level=4{font=\tiny}{},
  where level=5{font=\tiny}{},
  [Incremental Learning Challenge,rotate=90,anchor=north,edge=hiddendraw
    [Replay (\S\ref{7.3.1}),edge=hiddendraw,text width=4.8em
        [Experience \\Replay, text width=5em, edge=hiddendraw
            [{ORE~\cite{joseph2021towards}, ER~\cite{shieh2020continual}, ABR~\cite{yuyang2023augmented}, OSR\\ \cite{yang2023one}, RODEO~\cite{acharya2020rodeo}, Shieh \etal~\cite{shieh2022utilizing},\\ CL-DETR~\cite{liu2023continual}},leaf,text width=9em, edge=hiddendraw]
        ]
        [Feature Replay, text width=5em, edge=hiddendraw
            [{CIFRCN-NP~\cite{hao2019end}, iMTFA~\cite{ganea2021incremental}},leaf,text width=9em, edge=hiddendraw]
        ]
        [Generation \\Replay, text width=5em, edge=hiddendraw
            [{RT-Net~\cite{cui2023rt}},leaf,text width=9em, edge=hiddendraw]
        ]
    ]
    [Model (\S\ref{7.3.2}),edge=hiddendraw,text width=4.8em
     [Parameter \\Isolation, text width=5em, edge=hiddendraw
        [{MMN~\cite{li2018incremental}, DIODE\cite{peng2023diode}},leaf,text width=9em, edge=hiddendraw]
     ]
     [Model \\Decomposition, text width=5em, edge=hiddendraw
        [{LSTD~\cite{chen2018lstd}, Kang \etal~\cite{kang2019few}, iFSOD\\ \cite{cheng2021meta}, Sylph~\cite{yin2022sylph}, ONCE~\cite{perez2020incremental}, Dual-\\Fusion~\cite{tambwekar2021few}}
        ,leaf,text width=9em, edge=hiddendraw]
     ]
     [Modular \\Network, text width=5em, edge=hiddendraw
        [{context-Transformer~\cite{yang2020context}, Expert Det-\\ector~\cite{zhang2021incremental}, OW-DETR~\cite{gupta2022ow}, Rosetta\\ \cite{yang2022continual}},leaf,text width=9em, edge=hiddendraw]
     ]
     [Representation, text width=5em, edge=hiddendraw
        [{Rahman \etal~\cite{rahman2020any}, Incremental\\ DETR~\cite{dong2023incremental}},leaf,text width=9em, edge=hiddendraw]
     ]
    ]
    [Regularization \\(\S\ref{7.3.3}),edge=hiddendraw,text width=4.8em
      [Weight\\ Regularization, text width=5em, edge=hiddendraw
        [{BRS~\cite{cui2022balanced}, DIODE~\cite{peng2023diode}, IncDet~\cite{liu2020incdet}, \\ iFS-RCNN~\cite{nguyen2022ifs}},leaf,text width=9em, edge=hiddendraw]]
    [Structural\\ Regularization, text width=5em, edge=hiddendraw
      [{ILOD~\cite{shmelkov2017incremental}, Dong \etal~\cite{dong2021bridging}, LEAST \\ \cite{li2021class}, SID~\cite{peng2021sid}, MVCD~\cite{yang2022multi}, Faster \\ ILOD~\cite{peng2020faster}, PPAS~\cite{zhou2020lifelong}, Chen \etal \\ \cite{chen2019new}, ERD~\cite{feng2022overcoming}, Verwimp \etal~\cite{verwimp2022re}},leaf,text width=9em, edge=hiddendraw]]
    ]
    [Optimization (\S\ref{7.3.4}),edge=hiddendraw,text width=7em
     [{Teo \etal~\cite{li2021towards}, Joseph \etal~\cite{joseph2021incremental}, HDA~\cite{she2022fast}},leaf,text width=13.55em, edge=hiddendraw
     ]
    ]
 ]
\end{forest}
% \vspace{-6pt}
\caption{Solutions for incremental learning challenges.}
\label{incremental_data_solution}
% \vspace{-20pt}
\end{figure}
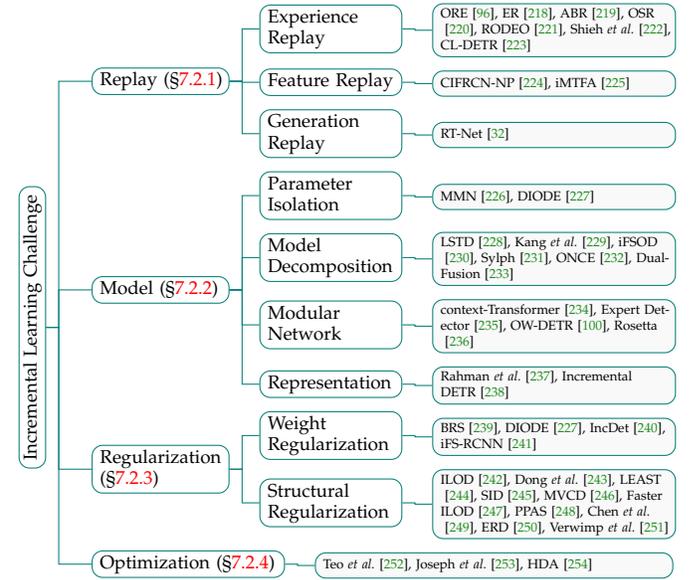

%% file: sections/10-benchmark.tex
\section{Benchmarks and Evaluation}
\label{sec:benchmark}
In this section, we first illustrate related datasets and evaluation metrics, then provide a detailed comparison and benchmark of existing methods in each specific challenge quadrant we discussed above. To facilitate meaningful comparisons, our analysis is confined to experimental settings that have been employed by two or more papers, where results are extracted from the corresponding papers. 

\subsection{Datasets and Metrics}

Pascal VOC (VOC)~\cite{everingham2010pascal} and MS COCO (COCO)~\cite{lin2014microsoft} are the most classical and fundamental datasets for object detection evaluation. In particular, VOC covers 20 common categories such as people and vehicles; COCO offers a broader range of categories and more complex scenarios, including small and dense objects. Besides the common benchmarks, there are also some datasets available for each type of challenge. For the evaluation of out-of-domain data, a variety of datasets such as Sim10k (Sim)~\cite{johnson2016driving}, Cityscapes (City)~\cite{cordts2016cityscapes}, Foggy Cityscapes (FCity)~\cite{sakaridis2018semantic}, KITTI (Kit)~\cite{geiger2013vision}, Berkeley DeepDrive (BDD)~\cite{xu2017end}, Clipart1k (Cli)~\cite{inoue2018cross}, and Watercolor2k (Wac) ~\cite{inoue2018cross} are utilized to ensure comprehensive coverage across multiple domains. To better explore the model's ability to identify out-of-category data, the large vocabulary dataset LVIS, which contains more than 1,000 object categories, is also used for the evaluation of open vocabulary detection. For assessing the robustness of detectors against malicious data, different attack methods are applied using common datasets and diverse satellite image datasets xView~\cite{lam2018xview} are also introduced for evaluation. To tackle class incremental data challenges, these common datasets are divided into two or more groups of categories, enabling the evaluation of a model capability for continuous learning and adaptation to new categories. 

In evaluating the performance of deep detectors, Mean Average Precision (mAP) is commonly used as it merges precision and recall aspects with multiple objects. However, the meaning of mAP is somewhat different for different challenges. For out-of-domain data, the evaluation focuses on the change in the model performance from the training domain to the target domain, \ie an mAP on the target domain. Meanwhile, for out-of-category data, we use mAP to measure whether the model can detect novel categories with open vocabulary. In evaluations of malicious and incremental data, mAP is mainly utilized to assess the robustness of malicious attacks and their effectiveness in learning new categories, respectively. 

\subsection{Out-of-Domain Solution Benchmarks}
\input{tables/out_of_domain_benchmark}
In this part, we benchmark the methods for out-of-domain challenges. We use Faster R-CNN as the object detector and ResNet-101 (R101) and VGG-16 (V16) as the backbone. All selected methods are implemented using their original codes with default settings. The evaluation encompasses diverse scenarios: \ding{182} Synthetic data (Sim) to real data (City) adaptation, where Sim is rendered from the Grand Theft Auto (GTAV) game engine, and City dataset represents urban driving scenes. \ding{183} Normal Weather (City) to Foggy Weather (FCity) Adaptation, with Fcity being a synthetic fog dataset based on City. \ding{184} Cross-sensor Adaptation. Comparing Kit, which utilizes various sensors like LIDAR, to City which primarily uses high-resolution camera images. \ding{185} Cross-scene Adaptation. City focuses on European urban streets, and BDD includes diverse driving conditions in the U.S. \ding{186} Cross-domain adaptation from VOC to abstract domains like Cli and Wac.

Tab.~\ref{out_of_domain_benchmark} delineates the detection efficacy of classical out-of-domain data methodologies across different target domains, using the metric mAP with an IoU threshold of 0.5. From the analysis, it is discernible that: \ding{182} Feature-based learning, especially with adversarial learning, tends to yield superior results, forming the crux of most current methodologies. \ding{183} We also introduce a new category of methods for DAOD named graph-based relational reasoning methods~\cite{LiLY22,TianZWXP21,LiuWHWX23,ChenLZHHDY23}, which is gaining attention and is poised to become a significant research trend in the future. \ding{184} Scene adaptation poses a significant challenge, highlighting a critical area for advancement. \ding{185} Predominantly, current methods are primarily built and designed for CNN-based backbones, with limited exploration of Transformers and visual language models, indicating potential avenues for future research. This benchmark underscores the necessity for continued innovation in adapting deep detectors to varied out-of-domain scenarios, particularly in scene adaptation and the incorporation of emerging network architectures. More evaluation results can be found in \emph{Supplementary Materials F}.

\input{tables/open_vocabulary_object_detection}

\subsection{Out-of-Category Solution Benchmarks}

In this part, we benchmark the methods for out-of-category challenges. Owing to the extensive application of visual language pre-training models and their effectiveness in addressing out-of-category issues, our focus is primarily on reporting the latest developments in open vocabulary object detection (OVOD).
% We use Mask R-CNN, DETR, and ATSS as the object detector and xxx as the backbone. All selected methods are implemented using their original codes with default settings. 
The common backbones used for the OVOD tasks include BERT, CLIP, and Swin-Transformer. CLIP includes two structures: ResNet-50 (denoted as $\mathbb{C}$) and Vision Transformer ($\mathbb{T}$). ResNet-50x4 and ResNet-50x64 are variations of the ResNet-50 structure, denoted $\mathbb{C}$x4 and $\mathbb{C}$x64. The Vision Transformer offers two backbones: ViT-B ($\mathbb{T}_{B}$) and ViT-L ($\mathbb{T}_{L}$), which have 12 and 24 network layers, respectively. Both versions use an input patch size of $16\times 16$. The Swin Transformer offers two size variants: Tiny (Swin-T) and Large (Swin-L). The Swin-L variant features a greater number of network layers, and its hidden layers possess more channels.
% ResNet (RN), ViT, and SwinTransformer (Swin)~\cite{liu2021swin} as backbones. 
% All selected methods are implemented using their original codes with default settings. 
The evaluation spans two challenging datasets: \ding{182}
the $mAP$ of 48 known classes and 17 unknown classes in the MS COCO~\cite{lin2014microsoft} dataset. \ding{183} the mAP of rare categories (mAP$_{r}$), common categories (mAP$_{c}$), and frequent categories (mAP$_{f}$) on the LVIS~\cite{gupta2019lvis} dataset. 

According to the results in Tab.~\ref{tab:open_vocabulary_object_detection}, we can identify: \ding{182} By extracting pseudo-labels from unlabeled data and refining the detector, the ability to identify unknown classes improves. For example, the PB-OVD method's introduction of pseudo-label training increased the mAP for unknown classes in the COCO dataset by 5.0. \ding{183} As visual language pre-training model parameters increase, detectors improve in identifying unknown categories. For instance, when the backbone of DetCLIPv2 transits from Swin-Transformer Tiny to Large, its ability to recognize unknown classes increases the mAP$_r$ in the LVIS dataset by 7.1. \ding{184} Superior pre-trained models like DetCLIPv2 exhibit stronger recognition in both known and unknown classes, indicating that advanced visual language pre-training models can improve unknown category recognition performance, \eg, under the same backbone (Swin-L), in the LVIS dataset, DetCLIPv2 outperforms GLIP by 26.0 mAP$_r$ in identifying unknown classes. More evaluation results can be found in \emph{Supplementary Materials F}.

\subsection{Robust Learning Solution Benchmarks}
\input{tables/benchmark_Adversarial_Training}
In this part, we report the results of each type of solution to protect object detectors against adversarial attacks.

Tab.~\ref{benchmark_Adversarial_Training} (a) evaluates the adversarial training with the SSD model on the VOC and COCO datasets. This evaluation considers clean examples and three classical adversarial attacks, \ie, $A_{cls}$, $A_{loc}$, and CWA. Specifically, adversarial training is conducted with 20 iterations for PGD attacks and 10 steps for CWAT, with a perturbation budget of 8. ``SSD'' represents the model detection effect without adversarial training. ``SSD-AT (cls)'' and ``SSD-AT (loc)'' represent classification and regression component training, respectively. Based on the results, we can observe: \ding{182} adversarial training enhances the SSD model performance against adversarial samples but reduces its efficacy in clean data, highlighting the typical accuracy-robustness trade-off. \ding{183} ATs like MTD, CWAT, RobustDet, and AIAD show certain performance against attacks on VOC but exhibit limited robustness on the complex COCO dataset, indicating a need for further optimization of adversarial defense strategies for real-world threats.

%Tab.~\ref{benchmark Adversarial Training} offers a detailed comparison of two primary classes of defense algorithms: model structure design-based methods and adversarial training paradigms. These are evaluated for their effectiveness in defending object detectors against various types of adversarial samples, with a specific focus on distinguishing between adversarial patch attacks and pixel attacks.

In addition, Tab.~\ref{benchmark_Adversarial_Training} (b) evaluates the defense performance of the model structure design method against adversarial patch attacks (with patch sizes 75x75, 100x100, and 125x125) on COCO~\cite{lin2014microsoft} and xView~\cite{lam2018xview} datasets using FPN detector. Here, we compare SAC with classical preprocessing defenses like JPG, SP, and LGS, where we can identify: \ding{182} the limited effectiveness of adversarial training methods against patch attacks, reflecting the design limitation that adversarial training mainly based on pixel attacks. \ding{183} existing preprocessing and adversarial training strategies significantly affect the prediction accuracy of clean images, highlighting the need for defense methods specifically designed for object detection. \ding{184} SAC performs best in maintaining clean image accuracy and shows a significant advantage with relatively small performance degradation as the size of the adversarial patch increases.

\subsection{ Benchmarks on incremental learning solutions.}
Depending on the amount of data used for incremental learning of new categories, this section outlines one experimental setting: multistep incremental object detection. Except that OW-DETR uses DETR as the detector, other methods use Faster R-CNN as the detector. All selected methods are implemented using their original codes with default settings.

\begin{figure}[!t]
 \centering
  \includegraphics[width=0.46\textwidth]{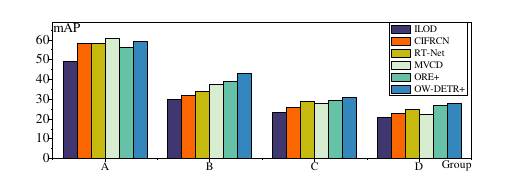}
 \caption{Multi-step incremental setting benchmarks.}
 \label{multi-step_incremental}
 % \vspace{-0.15in}
\end{figure}

Fig.~\ref{multi-step_incremental} shows the multi-step incremental object detection results in the COCO dataset. In this setup, the 80 categories of COCO are equally divided into four groups (A, B, C, and D), each containing 20 distinct categories. Specifically, ``+'' indicates the adherence to the experimental setup of the ORE for the incremental object grouping, which differs slightly from other methods. In this multi-part incremental process, the detection model sequentially incorporates a new set of object categories at each stage. We can observe: \ding{182} The complexity of multi-step incremental learning escalates with each additional step, as the model must continuously adapt to new sets of categories. \ding{183} The effectiveness of the model in assimilating new categories is heavily influenced by the specific combination of those categories. Varied sets can result in differing levels of learning difficulty, as observed in comparisons like MCVD vs. OW-DETR+. \ding{184} Structural enhancements in the model, such as the integration of the Transformer architecture in OW-DETR+, not only heighten detection accuracy but also bolster the model capability to successively learn new categories. More evaluation results can be found in \emph{Supplementary Materials F}.

%% file: tables/out_of_domain_benchmark.tex
\begin{table}[]
\caption{Benchmark on methods for the out-of-domain challenge.}
% \vspace{-8pt}
\label{out_of_domain_benchmark}
\centering
\renewcommand{\arraystretch}{1.05}
\setlength{\tabcolsep}{1pt}
\resizebox{\columnwidth}{!}{%
\rowcolors{12}{white}{gray!10}
\begin{tabular}{lccccccc}
\toprule
\multicolumn{1}{c}{\multirow{3}{*}[-0.5em]{\textbf{Method}}} & \multicolumn{7}{c}{\textbf{Source $\rightarrow$ Target Datasets}} \\
\multicolumn{1}{c}{} & Sim$\rightarrow$City & City$\rightarrow$FCity & \multicolumn{1}{l}{Kit$\rightarrow$City} & City$\rightarrow$Kit & City$\rightarrow$BDD & VOC$\rightarrow$Cli & VOC$\rightarrow$Wac \\ 
\cmidrule(l){2-2}\cmidrule(l){3-3}\cmidrule(lr){4-4}\cmidrule(lr){5-5}\cmidrule(lr){6-6}\cmidrule(lr){7-7}\cmidrule(lr){8-8}
\multicolumn{1}{c}{} & V16 & V16 & V16 & V16 & V16 & R101 & R101 \\ 
\midrule
Source & 34.6 & 23.4 & 37.4 & 53.5 & 23.4 & 27.8 & 44.6 \\
Target & 69.7 & 41.5 & - & - & 43.3 & 59.3 & 58.6 \\
DA-Faster~\cite{Chen0SDG18} & 39.0 & 27.6 & 38.5 & 64.1 & 24.0 & 19.8 & 46.0 \\
SWDA~\cite{SaitoUHS19} & 40.1 & 34.3 & 37.9 & 71.0 & 25.3 & 38.1 & 53.3 \\
D-adapt~\cite{JiangCWL22} & 50.3 & 41.3 & - & - & - & 49.0 & 57.5 \\
Selective DA~\cite{ZhuPYSL19} & 43.0 & 33.8 & 42.5 & - & - & - & - \\
CRDA~\cite{XuZJW20} & - & 37.4 & - & - & 26.9 & 38.3 & - \\
MCAR~\cite{ZhaoGSY20} & - & 38.8 & - & - & - & - & 56.0 \\
SAPNet~\cite{LiDZWLWZ20} & 44.9 & 40.9 & 43.4 & 75.2 & - & 42.2 & 55.2 \\
MeGA-CDA~\cite{VSGOSP21} & 44.8 & 41.8 & 43.0 & 75.5 & - & - & - \\
MAF~\cite{HeZ19} & 41.1 & 34.0 & 41.0 & 72.1 & - & 36.8 & - \\
MEAA~\cite{NguyenTS20} & 42.0 & 40.5 & - & - & - & 41.1 & 55.5 \\
ATF~\cite{HeZ20} & 42.8 & 38.7 & 42.1 & 73.5 & - & 42.1 & 54.9 \\
HTCN~\cite{ChenZD0D20} & 42.5 & 39.8 & - & - & - & 40.3 & - \\
CFFA~\cite{Zheng0LW20} & 43.8 & 38.6 & - & 73.6 & - & - & - \\
RPN-PA~\cite{ZhangWM21} & 45.7 & 40.5 & - & 75.1 & - & - & - \\
%DBGL~\cite{ChenLZ0DY21} & - & - & - & - & - & 41.6 & 53.8 \\
CDN~\cite{SuWZTCQW20} & 49.3 & 36.6 & 44.9 & - & - & - & - \\
ICCR-VDD~\cite{WuLHZ021} & - & 40.0 & - & - & - & - & 56.6 \\
PFD~\cite{WuHZY22} & - & 36.6 & - & - & - & 42.1 & 56.9 \\
ICCM~\cite{HouZFL21} & - & - & - & - & - & 46.7 & 57.4 \\
SIGMA~\cite{LiLY22} & 53.7 & 43.5 & 45.8 & - & - & - & - \\
DD-DML~\cite{KimJKCK19} & - & 34.6 & - & - & - & 41.8 & 52.0 \\
ST+C+RPL~\cite{RodriguezM19} & 44.2 & 29.7 & - & - & - & 44.8 & 57.3 \\
AFAN~\cite{WangLS21} & 45.5 & 39.6 & - & 74.7 & - & - & 50.6 \\
TSA-DA~\cite{YunHLKK21} & 42.9 & 37.9 & - & - & - & - & 57.4 \\
SC-UDA~\cite{YuWCKSYLLS022} & 52.4 & 38.7 & 46.4 & - & - & - & - \\
PDA~\cite{hsu2020progressive} & - & 36.9 & 43.9 & - & 24.3 & - & - \\
CST-DA~\cite{ZhaoLXL20} & 44.5 & 35.9 & 43.6 & - & - & - & - \\
CDG~\cite{LiH0Z21} & 48.8 & 42.3 & - & - & - & - & 59.7 \\
UMT~\cite{Deng0CD21} & 43.1 & 41.7 & - & - & - & 44.1 & 58.1 \\
TDD~\cite{HeWWWLLGWQ22} & 53.4 & 43.1 & 47.4 & - & \textbf{33.6} & - & - \\
AT~\cite{LiDMLCWHKV22} & - & 50.9 & - & - & - & \textbf{49.3} & \textbf{59.9} \\
KTNet~\cite{TianZWXP21} & 50.7 & 40.9 & 45.6 & - & - & - & - \\
%DC~\cite{LiuZWJY22} & 41.6 & - & - & - & - & 43.2 & 53.7 \\
TIA~\cite{Zhao022} & - & 42.3 & 44.0 & \textbf{75.9} & - & 46.3 & - \\
CMT~\cite{CaoJGW23} & - & \textbf{51.9} & - & - & - & 47.0 & - \\
CIGAR~\cite{LiuWHWX23} & \textbf{58.5} & 44.7 & \textbf{48.5} & - & - & 46.2 & - \\
FGRR~\cite{ChenLZHHDY23} & 44.5 & 40.8 & - & - & - & 43.3 & 55.7 \\
\bottomrule
\end{tabular}%
}
% \vspace{-10 pt}
\end{table}

%% file: tables/open_vocabulary_object_detection.tex
% \vspace{-5mm}
\begin{table}[t]
    \centering    \makeatletter\def\@captype{table}\makeatother\caption{Open-vocabulary object detection performance on MS COCO and LVIS datasets. \ddag~denotes the method trained using the generated pseudo-labels.}
    % \vspace{-8pt}
    \setlength{\tabcolsep}{2pt}
    \resizebox{\columnwidth}{!}{
    \rowcolors{5}{gray!10}{white}
    \begin{tabular}{lcccccccc} 
\toprule
\multicolumn{1}{c}{\multirow{2}{*}[-0.35em]{Method}} & \multirow{2}{*}[-0.35em]{Backbone} & \multicolumn{3}{c}{COCO Dataset} & \multicolumn{4}{c}{LVIS Dataset} \\ 
\cmidrule(l){3-5} \cmidrule(l){6-9}
\multicolumn{1}{c}{} &  & Known & UnKnown & All & \multicolumn{1}{l}{mAP$_{r}$} & \multicolumn{1}{l}{mAP$_{c}$} & \multicolumn{1}{l}{mAP$_{f}$} & \multicolumn{1}{l}{All} \\ 
\midrule
OVR-CNN~\cite{zareian2021open} & BERT & 46.0 & 22.8 & 39.9 & - & - & - & - \\
ViLD~\cite{guopen} & $\mathbb{C}$ & 59.5 & 27.6 & 51.3 & 16.7 & 26.5 & 34.2 & 27.8 \\
DetPro~\cite{du2022learning} & CLIP ($\mathbb{T}_B$) & - & - & - & 20.8 & 27.8 & 32.4 & 28.4 \\
HierKD~\cite{ma2022open} & CLIP ($\mathbb{T}_B$) & 51.3 & 20.3 & 43.2 & - & - & - & - \\
RegionCLIP~\cite{zhong2022regionclip} & $\mathbb{C}$ & 57.1 & 31.4 & 50.4 & 17.1 & 27.4 & 34.0 & 28.2 \\
RegionCLIP~\cite{zhong2022regionclip} & $\mathbb{C}$x4 & \textbf{61.6} & 39.3 & 55.7 & 22.0 & 32.1 & 36.9 & 32.3 \\
\ddag~VL-PLM~\cite{zhao2022exploiting} & CLIP ($\mathbb{C}$) & 60.2 & 34.4 & 53.5 & 17.2 & 23.7 & 35.1 & 27.0 \\
OV-DETR~\cite{zang2022opendetr} & CLIP ($\mathbb{T}_B$) & 61.0 & 29.4 & 52.7 & 17.4 & 25.0 & 32.5 & 26.6 \\
PB-OVD~\cite{gao2022open} & CLIP ($\mathbb{T}_B$) & - & 25.8 & - & - & - & - & - \\
\ddag~PB-OVD~\cite{gao2022open} & CLIP ($\mathbb{T}_B$) & 46.1 & 30.8 & 42.1 & - & - & - & - \\
OWL-ViT~\cite{zang2022open} & CLIP ($\mathbb{T}_L$) & - & - & - & 25.6 & - & - & 34.7 \\
PromptDet~\cite{feng2022promptdet} & CLIP ($\mathbb{C}$) & - & 26.6 & 50.6 & 21.4 & 23.3 & 29.3 & 25.3 \\
% PromptDet~\cite{feng2022promptdet} & CLIP ($\mathbb{C}$) & 57.1 & 26.6 & 50.6 & 21.4 & 23.3 & 29.3 & 25.3 \\
Rasheed \etal~\cite{bangalath2022bridging} & CLIP ($\mathbb{T}_B$) & 54.0 &   36.6 & 49.4 & - & - & - & - \\
\ddag~Rasheed \etal~\cite{bangalath2022bridging} & CLIP ($\mathbb{T}_B$) & 56.6 & 36.9 & 51.5 & 21.1 & 25.0 & 29.1 & 25.9 \\
F-VLM~\cite{kuo2023f} & CLIP ($\mathbb{C}$) & - & 28.0 & 39.6 & 18.6 & - & - & 24.2 \\
% F-VLM~\cite{kuo2023f} & CLIP ($\mathbb{C}$) & 54.6 & 26.2 & 37.4 & 15.5 & 25.3 & 31.1 & 22.5 \\
RO-ViT~\cite{kim2023region} & CLIP ($\mathbb{T}_L$) & - & 33.0 & 47.7 & 32.1 & - & - & 34.0 \\
UniDetector~\cite{wang2023detecting} & RegionCLIP ($\mathbb{C}$) & 56.8 & 35.2 & 51.2 & 18.0 & 19.2 & 21.2 & 19.8 \\
% CORA~\cite{wu2023cora} & CLIP ($\mathbb{C}$) & 54.3 & 34.2 & 33.8 & 15.6 & 25.2 & 31.8 & 26.3 \\
% CORA~\cite{wu2023cora} & CLIP ($\mathbb{C}$x64) & 59.2 & 42.3 & \textbf{55.1} & 21.3 &  31.4 & 35.4 & 31.1 \\
CORA~\cite{wu2023cora} & CLIP ($\mathbb{C}$) & 35.1 & 35.5 & 35.4 & - & - & - & - \\
CORA~\cite{wu2023cora} & CLIP ($\mathbb{C}$x64) & 60.9 & \textbf{43.1} & \textbf{56.2} & 28.1 & - & - & - \\
GLIP~\cite{li2022grounded} & Swin-L & - & - & - & 17.1 & 23.3 & 35.4 & 26.9 \\
% GLIP~\cite{li2022grounded} & Swin-T & 55.3 & 38.7 & 46.4 & 9.7 & 11.8 & 24.1 & 16.9 \\
% GLIP~\cite{li2022grounded} & Swin-L & 56.1 & \textbf{43.4} & 49.7 & 16.6 & 21.8 & 33.2 & 25.9 \\
DetCLIP~\cite{yao2022detclip} & Swin-T & - & - & - & 33.2 & 35.7 & 36.4 & 35.9  \\
DetCLIP~\cite{yao2022detclip} & Swin-L & - & - & - & 36.0 & 38.3 & 39.3 & 38.6  \\
DetCLIPv2~\cite{yao2023detclipv2} & Swin-T & - & - & - & 36.0 & 41.7 & 40.0 & 40.4 \\
DetCLIPv2~\cite{yao2023detclipv2} & Swin-L & - & - & - & \textbf{43.1} & \textbf{46.3} & \textbf{43.7} & \textbf{44.7} \\
% CORA~\cite{wu2023cora} & CLIP ($\mathbb{C}$x64) & - & - & - & \textbf{28.1} & - & - & - \\
\bottomrule
    \end{tabular}
    }
    \label{tab:open_vocabulary_object_detection}
% \vspace{-15pt}
\end{table}

%% file: tables/benchmark_Adversarial_Training.tex
\begin{table}[!t]
\caption{Benchmarks on solutions for robust learning challenge.}
% \vspace{-10pt}
\label{benchmark_Adversarial_Training}
\centering
\large

\newcolumntype{Z}{>{\centering\arraybackslash}p{\dimexpr.95\columnwidth/7}}
\setlength{\tabcolsep}{10pt}
\resizebox{\columnwidth}{!}{%
\begin{tabular}{lZZZZZZZZ}

\\
\multicolumn{9}{c}{\textbf{(a) Adversarial Training Paradigm}} \\
\toprule
\multicolumn{1}{c}{\multirow{2}{*}[-0.3em]{Method}} & \multicolumn{4}{c}{VOC Dataset} & \multicolumn{4}{c}{COCO Dataset} \\ 
\cmidrule(l){2-5} \cmidrule(l){6-9}
 & clean & $A_{cls}$ & $A_{loc}$ & CWA & clean & $A_{cls}$ & $A_{loc}$ & CWA \\ 
\midrule
\rowcolor{gray!10} SSD~\cite{liu2016ssd} & 77.5 & 1.8 & 4.5 & 1.2 & 42.0 & 0.4 & 1.8 & 0.1 \\
SSD-AT (cls) & 46.7 & 21..8 & \textbf{32.2} & - & - & - & - & - \\
\rowcolor{gray!10} SSD-AT (loc) & 51.9 & 23.7 & \textbf{26.5} & - & - & - & - & - \\
MTD~\cite{zhang2019towards} & 48.0 & 29.1 & 31.9 & 18.2 & 24.2 & 13.0 & 13.4 & 7.7 \\
\rowcolor{gray!10} CWAT~\cite{chen2021class} & 51.3 & 22.4 & 36.7 & 19.9 & 23.7 & 14.2 & 15.5 & 9.2 \\
RobustDet~\cite{dong2022adversarially} & 74.8 & 45.9 & 49.1 & 48.0 & 36.0 & 20.0 & 19.0 & 19.9 \\
\rowcolor{gray!10} AIAD~\cite{chengadversarial} & 77.2 & 51.1 & 47.9 & 49.0 & 37.7 & 20.1 & 19.7 & 20.3 \\

\bottomrule
\\
\multicolumn{9}{c}{\textbf{(b) Model Robust Inference}} \\
\toprule
\multicolumn{1}{c}{\multirow{2}{*}[-0.3em]{Method}} & \multicolumn{4}{c}{COCO Dataset} & \multicolumn{4}{c}{xView Dataset} \\ 
\cmidrule(l){2-5} \cmidrule(l){6-9}
 & clean & \( 75^2 \) & \( 100^2 \) & \( 125^2 \) & clean & \( 75^2 \) & \( 100^2 \) & \( 125^2 \) \\ 
\midrule
FPN~\cite{lin2017feature} & 49.0 & 19.8 & 14.4 & 9.9 & 27.2 & 8.4 & 7.1 & 5.3 \\
\rowcolor{gray!10}JPG~\cite{dziugaite2016study} & 45.6 & 39.7 & 37.2 & 33.3 & 23.3 & 19.3 & 17.8 & 15.9 \\
SP~\cite{xu2017feature} & 46.0 & 40.4 & 38.1 & 34.3 & 21.8 & 16.2 & 14.2 & 12.4 \\
\rowcolor{gray!10}LGS~\cite{naseer2019local} & 42.7 & 36.8 & 35.2 & 32.8 & 19.1 & 11.9 & 10.9 & 9.8 \\
SAC~\cite{liu2022segment} & 49.0 & 45.7 & 45.0 & 40.7 & 27.2 & 25.3 & 23.6 & 23.2 \\ 
\bottomrule

\end{tabular}
}
% \vspace{-10pt}
\end{table}

%% file: sections/8-future_outlook.tex
\section{Discussion and Conclusion}
\label{sec:outlook}
This section outlines future research directions and concludes, aiming to inspire and guide further development in the field.

\subsection{Observation of Technology Trend}

\ding{182} \textbf{Broadening detector generalization evaluation.} Addressing the out-of-domain challenge requires a thorough evaluation of detector generalizability, which includes evaluating diverse detector architectures such as YOLO~\cite{bochkovskiy2020yolov4}, SSD~\cite{liu2016ssd}, FCOS~\cite{tian2019fcos} and DETR~\cite{carion2020end}. To capture real-world complexity, it is essential to move beyond traditional datasets such as FoggyCityscapes~\cite{sakaridis2018semantic}, Sim10K~\cite{johnson2016driving}, and Pascal VOC~\cite{everingham2007pascal} by creating new datasets. In addition, evaluations should expand into underexplored areas such as medical imaging and text recognition, alongside conventional applications in autonomous driving and crowd surveillance.

\ding{183} \textbf{Incorporating knowledge for reliable detector.} For the out-of-category challenge, future efforts should aim at embedding human knowledge into detection models to enhance decision accuracy. This includes domain-specific insights~\cite{yao2022detclip}, common sense reasoning~\cite{ma2024codet}, contextual cues~\cite{ma2022open}, and broader linguistic and textual understanding~\cite{menon2023visual}. Such integration can significantly refine the model's judgments in intricate situations, offering richer contextual insights and explanatory depth.

\ding{184} \textbf{Developing practical attack and defense strategies.} In the face of a robust learning challenge, future research should focus on combining malicious data with the actual physical environment to improve the realistic applicability of attack scenarios and the completeness of security assessment. Additionally, it is crucial to develop lightweight and efficient defense mechanisms that dynamically adapt to evolving adversarial contexts, ensuring protection against emerging security threats.

\ding{185} \textbf{Optimizing costs during incremental learning.} In the face of the challenges of incremental learning, especially the significant investment of time and resources required for the labeling of new object categories. Therefore, researchers could explore ways to efficiently represent new objects using finite samples. The core of future research should focus on preventing overfitting problems during incremental learning and improving the efficiency and accuracy of learning new objects under small-sample conditions. Additionally, research should aim to enhance the model's self-adaptive capabilities through algorithm optimization, data processing enhancements, and innovations in model update mechanisms to advance incremental learning techniques.

\subsection{Technical Areas That Can Be Combined}
\ding{182} \textbf{Research on advanced fundamental models.} Currently, while numerous base models have been developed, they often fail to focus sufficiently on localized regions when dealing with details. Studies have shown that the larger the size and diversity of the pre-training dataset used by the base model, the better its ability to recognize and detect unknown objects in the object detection task. Therefore, future base model development not only needs to expand and diversify the pre-training dataset but also consider the potential connections between objects in depth and integrate certain a priori knowledge structures into the pre-training stage to promote deeper understanding and generalization capabilities of the model. Such an advanced base model will lead to more accurate and flexible solutions in the field of object detection.

\ding{183} \textbf{Innovative interpretable open-world object detection techniques.} Interpretable methods~\cite{chen2023sim2word,chen2024less,bao2021drive} have not been widely developed in the current field of object detection. Utilizing multimodal models in conjunction with large-scale language models like GPT-4 for object classification can provide an explanation of the results by generating descriptive scores, a strategy that can effectively reveal how the model recognizes unknown object classes, thereby increasing the reliability and trust of the detector. Furthermore, integrating human knowledge into the pre-training process of multimodal base models for object detection can facilitate the formation of more robust model representation capabilities~\cite{li2023disc}. Through such an approach, we can not only improve the model's recognition capability, but also provide users with a clear path of understanding, thus promoting open-world object detection technology to a higher level.

\ding{184} \textbf{Expansion of application scope through fusion of multi-dimensional targets.} The recent development of open-world object detectors such as those shown in ORE~\cite{joseph2021towards} signifies the continued expansion of the range of applications in this area. The capability of such systems, which is not limited to recognizing known categories, but also includes learning about unknown categories, is critical to broadening their applicability in a variety of application scenarios. In addition, Fujii \etal~\cite{fujii2022adversarially} further demonstrate the value of integrating these capabilities to cope with the complexity of different environments by showing cases in which the models are enhanced through adaptation and robustness.

%% file: sections/9-conclusion.tex
\subsection{Conclusion}
This paper for the first time provides a comprehensive review of the development of challenges and solutions of deep object detectors in the open environment. We analyze the limitations of key components in deep detectors and explore the four main challenges of open environment object detectors from the dimensions of data change and learning target change, summarizing their problems and counter strategies in detail. We also present a benchmark for these methods. This paper aims to catalyze the development of more reliable applications in real-world scenarios.

%% file: suppl_sections/1-related_survey.tex
\section{Overview of related surveys}
\input{tables/RelatedSuveryList}
In Tab.~\ref{RelatedSurveyList}, we present a collection of review papers from the past five years within the domain of deep object detection, aimed at providing readers with insights into the current research developments in this field. These review papers on depth detectors are organized into three categories: comprehensive overviews of the field, reviews focusing on specific areas within the field, and analyses of emerging topics.

The general field review provides an overview of the development of deep object detection. Taking time as a clue, Zou \etal~\cite{zou2019object} review the development of object detectors over the last 20 years, marking milestones in the historical development of detectors. Liu \etal~\cite{liu2020deep} provide an exhaustive study of the last five years of research on object detectors, covering the various components of detectors, including detection frameworks, object feature representations, object proposal generation, contextual information modeling, and training strategies. Deep learning, especially neural networks, greatly facilitates the development of deep object detectors. Therefore, Jiao \etal~\cite{jiao2019survey} analyze the trend of the development of object detection models from the perspective of deep learning, introducing one-stage detectors and two-stage detectors using the ANCHOR technique as a distinction. Similarly, Dhillon~\cite{dhillon2020convolutional} discusses deep learning structures in terms of different architectures of CNN models, such as AlexNet~\cite{krizhevsky2012imagenet}, ZFNet~\cite{zeiler2014visualizing}, and GoogleNet~\cite{szegedy2015going}, for different object detection tasks. However, deep learning models do not always maintain an advantage in detectors, and Kaur \etal~\cite{kaur2023comprehensive} dialectically analyze the advantages and disadvantages of deep learning models versus traditional detection methods and compare the detection performance of these two different genres. In addition, Padilla \etal~\cite{padilla2021comparative} focus on evaluation methods and tools in the field of object detection, presenting the latest evaluation methods, as well as object annotation formats, and analyzing their impact.

Specific Domain reviews focus on deep object detector approaches in specific subfields. Due to the wide range of applications of object detectors, such as face recognition, industrial inspection, etc., we cannot list all subfield-specific reviews related to object detection. Here, we mainly summarize the application articles with ``object detection'' in the title, covering small object detection, imbalance problems, and domain-specific detection. For small object detection, Cheng \etal~\cite{cheng2023towards} review the problems of poor visual appearance and noisy representations inherent in small objects, thoroughly review the development of small object detection, and construct two related datasets to attempt large-scale basic testing, similar to the review~\cite{tong2020recent}. Oksuz \etal~\cite{oksuz2020imbalance} summarize the imbalance problems in object detection, categorizing them as class imbalance, scale imbalance, spatial imbalance, and object imbalance. Domain-specific review categories include the review of low-altitude UAVs by Mittal \etal~\cite{mittal2020deep}.

\begin{figure*}[!t]
 \centering
  \begin{overpic}[width=\textwidth,tics=8]{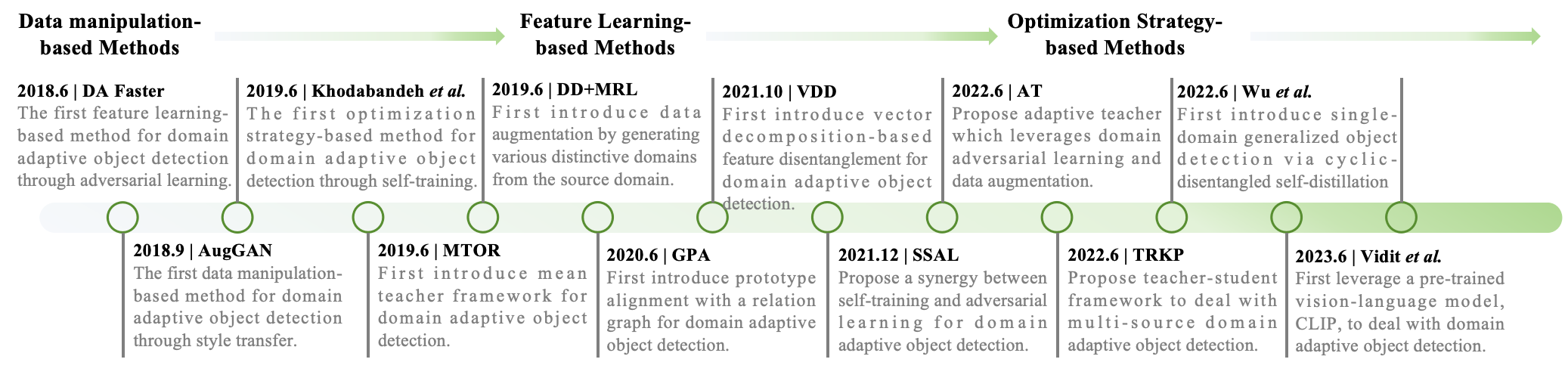}  
        % first line 
        \put (10.5,18.45) {\tiny \cite{Chen0SDG18}}
        \put (29.4,18.45) {\tiny \cite{KhodabandehVRM19}}
        \put (40.5,18.45) {\tiny \cite{KimJKCK19}}
        \put (53.2,18.45) {\tiny \cite{WuLHZ021}}
        \put (66.4,18.45) {\tiny \cite{LiDMLCWHKV22}}
        \put (83.8,18.45) {\tiny \cite{WuD22}}
        % bottom line
        \put (17.4,8.25) {\tiny \cite{HuangLCWHL18}}
        \put (31.8,8.25) {\tiny \cite{CaiPNTDY19}}
        \put (45.2,8.1) {\tiny \cite{XuWNTZ20}}
        \put (61,8.1) {\tiny \cite{MunirKSA21}}
        \put (75.4,8.0) {\tiny \cite{WuCHWLMGWW022}}
        \put (92,7.96) {\tiny \cite{ViditES23}}
    \end{overpic} 
 \caption{Timeline for solving out-of-domain solutions in object detection.}
 \label{fig:timeline_OOD}
\end{figure*}
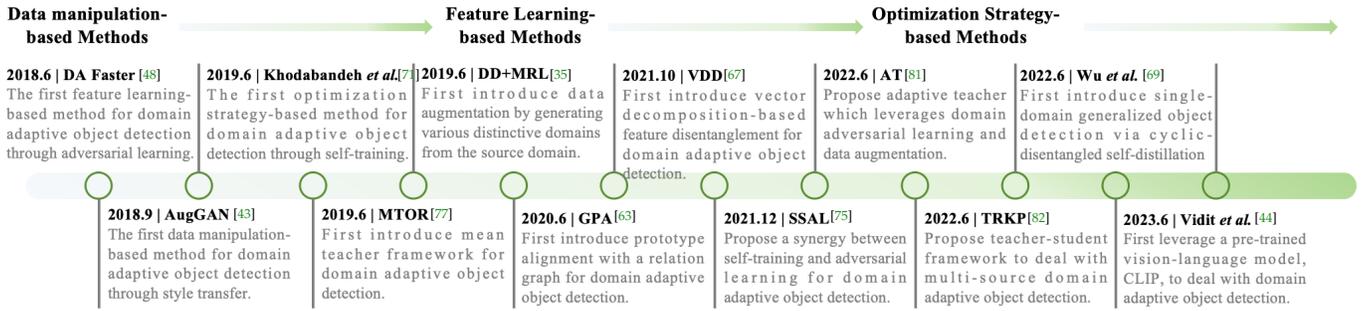

\input{tables/out_of_domain_summary}

The new issues review aims to review and summarize emerging issues in the field of object detection, and existing reviews are mainly related to unsupervised domain migration, continuous learning, and robustness. Oza \etal~\cite{oza2023unsupervised} review the problem of unsupervised domain adaptation in object detection, that is, degradation of model performance during the evaluation of label-scarce datasets. The authors provided a comprehensive comparison of the strengths and weaknesses of the method, the performance of the datasets, and the evaluation of model performance from three aspects. comprehensive comparison Menezes \etal~\cite{menezes2023continual} focus on the incremental classification task in the object detection class, systematically reviewing how continuous learning strategies can be applied to the object detection task, and the authors look at the performance performance of existing detectors in terms of new evaluation metrics. In terms of robustness, Amirkhani \etal~\cite{amirkhani2023survey} focus on a systematic review of adversarial attack and defense in the field of object detection, elucidating its strengths and weaknesses; in addition, the authors review the literature on adversarial robustness in the field of automated driving to further add to the importance of adversarial security research.

Although this paper focuses on a review of an emerging problem, we pay special attention to the problem of object detection in dynamically changing open environments. Distinguishing this paper from other review works, it is unique in the following ways: first, we not only review the progress of deep object detection techniques, but also delve into the limitations of existing detection architectures; second, in the context of open environments, this paper takes into account the changes of data and objects during the detection process, and such considerations are not only closely related to the applications in the key subfields but also more consistent with the real world context; third, we synthesize the existing challenges and summarize the problem of object detection in open environments into four main aspects and highlight their interconnections, providing an important summary of the current main challenges in open environments; finally, we discuss several key techniques, such as multimodal base models and interpretability techniques, and explore how they can be exploited to improve the performance of object detectors in the open environment in the future. performance of object detectors in the environment. We hope that these new insights will help professionals develop more widely applicable object detectors.

%% file: tables/RelatedSuveryList.tex
\begin{table*}[]
\caption{Summary of object detection literature reviews. This table provides a list of recent surveys and reviews in the field of object detection, detailing their focus areas, publications, and core contributions to the understanding and advancement of object detection technologies.}
\label{RelatedSurveyList}
% \centering
\footnotesize
\rowcolors{3}{gray!10}{white}
\resizebox{\textwidth}{!}{%
\setlength{\tabcolsep}{4pt}
\renewcommand{\arraystretch}{1.2}
\begin{tabular}{p{.4\textwidth}|p{.18\textwidth}|p{.42\textwidth}}
\toprule
Title & Publication & Description \\ \midrule
\multicolumn{3}{l}{\textbf{General Field Review}} \\
Object detection in 20 years: A survey~\cite{zou2019object} & Proc. IEEE 2023 & The survey looks back at 20 years of object detectors through a timeline. \\
A Survey of Deep Learning-Based Object Detection~\cite{jiao2019survey} & IEEE Access 2019 & The survey  provides an in-depth analysis of both one-stage and two-stage detectors. \\
Deep Learning for Generic Object Detection: A Survey~\cite{liu2020deep} & IJCV 2020 & The survey provides a comprehensive review of common object detection algorithms over the past five years. \\
A comprehensive review of object detection with deep learning~\cite{kaur2023comprehensive} & Digit. Signal Process 2022 & The survey explores in detail the advantages of deep learning and traditional methods in object detection. \\
A comparative analysis of object detection metrics with a companion open-source toolkit~\cite{padilla2021comparative} & Electronics 2021 & The survey summarizes existing detection methods from annotated data sets and evaluation perspectives. \\
Convolutional neural network: a review of models, methodologies and applications to object detection~\cite{dhillon2020convolutional} & Prog Artif Intell 2019 & The survey discusses the open applications of object detection from the perspective of deep learning. \\ \midrule
\multicolumn{3}{l}{\textbf{Specific Domain Review}} \\
Towards Large-Scale Small Object Detection: Survey and Benchmarks~\cite{cheng2023towards} & TPAMI 2023 & The survey provides a thorough review of small object detection. \\
Imbalance problems in object detection: A review~\cite{oksuz2020imbalance} & TPAMI 2020 & The survey describes the problem of unbalance in object detection. \\
Recent advances in small object detection based on deep learning: A review~\cite{tong2020recent} & Image Vis. Comput 2020 & The survey summarizes the methods, performance and potential directions of small object detection. \\
Deep learning-based object detection in low-altitude UAV datasets: A survey~\cite{mittal2020deep} & Image Vis. Comput 2020 & The survey reviews deep object detection for use in the UAV field. \\ \midrule
\multicolumn{3}{l}{\textbf{New Issues Review}} \\
Unsupervised Domain Adaptation of Object Detectors: A Survey~\cite{oza2023unsupervised} & TPAMI 2023 & The survey discusses object detectors within domain adaptation issues. \\
Continual Object Detection: A review of definitions, strategies, and challenges~\cite{menezes2023continual} & Neural Netw. 2023 & The survey summarises existing strategies and challenges for continual  learning in object detection. \\
A survey on adversarial attacks and defenses for object detection and their applications  in autonomous vehicles~\cite{amirkhani2023survey} & Vis. Comput. 2022 & The survey reviews the field of adversarial robustness for object detection. \\ \bottomrule
\end{tabular}%
}
\end{table*}

%% file: tables/out_of_domain_summary.tex
\vspace{-1mm}
\begin{table*}[!t]
\vspace{-1mm}
\caption{Detailed categorization of solutions to address out-of-domain challenges in object detection.}
\label{solutions for ood}
\renewcommand{\arraystretch}{1.2}
\rowcolors{6}{white}{gray!10}
\setlength{\tabcolsep}{16pt}
\resizebox{\textwidth}{!}{%
\begin{tabular}{lcccccccc}
\toprule
\multicolumn{1}{c}{\multirow{2}{*}{Method}} & \multirow{2}{*}{Paper List} & \multicolumn{2}{c}{\cellcolor[HTML]{EDF5EC}Data Manipulation} & \multicolumn{3}{c}{\cellcolor[HTML]{D7ECC9}Feature Learning} & \multicolumn{2}{c}{\cellcolor[HTML]{B0D993}Optimization Strategy} \\  \cmidrule(l){3-4} \cmidrule(l){5-7} \cmidrule(l){8-9}
\multicolumn{1}{c}{} &  & \begin{tabular}[c]{@{}c@{}}Data\\ Augmentation\end{tabular} & \begin{tabular}[c]{@{}c@{}}Style\\ Transfer\end{tabular} & \begin{tabular}[c]{@{}c@{}}Adversarial\\ Learning\end{tabular} & Prototype & \begin{tabular}[c]{@{}c@{}}Feature \\ Disentanglement\end{tabular} & \begin{tabular}[c]{@{}c@{}}Self-\\ training\end{tabular} & \begin{tabular}[c]{@{}c@{}}Mean \\ Teacher\end{tabular}\\ \midrule
Kim \etal~\cite{KimJKCK19} & CVPR 2019 & \CIRCLE &  & \RIGHTcircle &  &  &  &  \\
Khirodkar \etal~\cite{KhirodkarYK19} & WACV 2019 & \CIRCLE &  &  &  &  &  &  \\
Prakash \etal~\cite{PrakashBBACSSB19} & ICRA 2019 & \CIRCLE &  &  &  &  &  &  \\
Wang \etal~\cite{WangLS21} & TIP 2021 & \CIRCLE &  & \RIGHTcircle &  &  &  &  \\
Hsu \etal~\cite{hsu2020progressive} & WACV 2020 & \CIRCLE &  & \RIGHTcircle &  &  &  &  \\
Chen \etal~\cite{ChenZD0D20} & CVPR 2020 & \RIGHTcircle &  & \CIRCLE &  &  &  &  \\
Gao \etal~\cite{GaoLYW023} & CVPR 2023 & \CIRCLE &  &  &  &  &  &  \\
Huang \etal~\cite{HuangLCWHL18} & ECCV 2018 & \CIRCLE &  &  &  &  &  &  \\
Rodriguez \etal~\cite{RodriguezM19} & BMVC 2019 &  & \CIRCLE &  &  &  & \RIGHTcircle &  \\
Yun \etal~\cite{YunHLKK21} & RAL 2021 &  & \CIRCLE &  &  &  &  &  \\
Yu \etal~\cite{YuWCKSYLLS022} & BMVC 2022 &  & \CIRCLE &  &  &  & \RIGHTcircle &  \\
Vidit \etal~\cite{ViditES23} & CVPR 2023 &  & \CIRCLE &  &  &  &  &  \\
Chen \etal~\cite{Chen0SDG18} & CVPR 2018 &  &  & \CIRCLE &  &  &  &  \\
Saito \etal~\cite{SaitoUHS19} & CVPR 2019 &  &  & \CIRCLE &  &  &  &  \\
Zhu \etal~\cite{ZhuPYSL19} & CVPR 2019 &  &  & \CIRCLE &  &  &  &  \\
Xu \etal~\cite{XuZJW20} & CVPR 2020 &  &  & \CIRCLE &  &  &  &  \\
Li \etal~\cite{LiDZWLWZ20} & ECCV 2020 &  &  & \CIRCLE &  &  &  &  \\
He \etal~\cite{HeZ19} & ICCV 2019 &  &  & \CIRCLE &  &  &  &  \\
Nguyen \etal~\cite{NguyenTS20} & ACMMM 2020 &  &  & \CIRCLE &  &  &  &  \\
Fu \etal~\cite{FuXLD20} & arXiv 2020 &  &  & \CIRCLE &  &  &  &  \\
He \etal~\cite{HeZ20} & ECCV 2020 &  &  & \CIRCLE &  &  &  &  \\
Zhao \etal~\cite{ZhaoGSY20} & ECCV 2020 &  &  & \CIRCLE &  &  &  &  \\
Rezaeianaran \etal~\cite{RezaeianaranSAR21} & ICCV 2021 &  &  & \RIGHTcircle &  &  &  &  \\
Jiang \etal~\cite{JiangCWL22} & ICLR 2022 &  &  & \CIRCLE &  &  &  &  \\
Yao \etal~\cite{YaoZ0Y21} & ICCV 2021 &  &  & \CIRCLE &  &  &  &  \\
Wang \etal~\cite{Wang00HZ0T21} & ACMMM 2021 &  &  &  &  &  &  &  \\
Hou \etal~\cite{HouZFL21} & CVPR 2021 &  &  &  &  &  &  &  \\
Zhao \etal~\cite{Zhao022} & CVPR 2022 &  &  & \CIRCLE &  &  &  &  \\
Xu \etal~\cite{XuWNTZ20} & CVPR 2020 &  &  &  & \CIRCLE &  &  &  \\
Zheng \etal~\cite{Zheng0LW20} & CVPR 2020 &  &  &  & \CIRCLE &  &  &  \\
Zhang \etal~\cite{ZhangWM21} & CVPR 2021 &  &  &  & \CIRCLE &  &  &  \\
Su \etal~\cite{SuWZTCQW20} & ECCV 2020 &  &  & \RIGHTcircle &  & \CIRCLE &  &  \\
Wu \etal~\cite{WuLHZ021} & ICCV 2021 &  &  & \RIGHTcircle &  & \CIRCLE &  &  \\
Wu \etal~\cite{WuHZY22} & TPAMI 2022 &  &  & \RIGHTcircle &  & \CIRCLE &  &  \\
Wu \etal~\cite{WuD22} & CVPR 2022 &  &  &  &  & \CIRCLE &  &  \\
RoyChowdhury \etal~\cite{RoyChowdhuryCSJ19} & CVPR 2019 &  &  &  &  &  & \CIRCLE &  \\
Khodabandeh \etal~\cite{KhodabandehVRM19} & ICCV 2019 &  &  &  &  &  & \CIRCLE &  \\
Kim \etal~\cite{KimCKK19} & ICCV 2019 &  &  & \RIGHTcircle &  &  & \CIRCLE &  \\
Li \etal~\cite{LiCXYYPZ21} & AAAI 2021 &  &  &  &  &  & \CIRCLE &  \\
Li \etal~\cite{LiH0Z21} & AAAI 2021 &  &  &  &  &  & \CIRCLE &  \\
Munir \etal~\cite{MunirKSA21} & NIPS 2021 &  &  & \RIGHTcircle &  &  & \CIRCLE &  \\
Lu \etal~\cite{0006ZS23} & AAAI 2023 & \RIGHTcircle &  &  &  &  & \CIRCLE &  \\
Cai \etal~\cite{CaiPNTDY19} & CVPR 2019 &  &  &  &  &  &  & \CIRCLE \\
Deng \etal~\cite{Deng0CD21} & CVPR 2021 & \RIGHTcircle &  &  &  &  &  & \CIRCLE \\
He \etal~\cite{HeWWWLLGWQ22} & CVPR 2022 &  &  &  &  &  &  & \CIRCLE \\
Li \etal~\cite{LiDMLCWHKV22} & CVPR 2022 & \Circle &  & \RIGHTcircle &  &  &  & \CIRCLE \\
Wu \etal~\cite{WuCHWLMGWW022} & CVPR 2022 &  &  &  &  &  &  & \CIRCLE \\
Kennerley \etal~\cite{KennerleyWVT23} & CVPR 2023 & \RIGHTcircle &  &  &  &  &  & \CIRCLE \\ \bottomrule
\end{tabular}%
}
\end{table*}

%% file: suppl_sections/4-out-of-domain.tex
\section{Out-of-Domain Challenge}

\subsection{Timeline}

We depict the timeline of domain adaptive object detection (DAOD) methods in Fig.~\ref{fig:timeline_OOD}. Chen \etal~\cite{Chen0SDG18} are the first to propose the DAOD problem, employing adversarial learning at the feature level to learn domain-invariant features. Huang \etal~\cite{HuangLCWHL18} pioneer the use of style transfer at the data input layer to generate images similar to the target domain. Khodabandeh \etal~\cite{KhodabandehVRM19} are the first to suggest the use of self-training as an optimization strategy to enhance the generalization of the source domain-trained model to the target domain. Subsequently, at the data input layer, Kim \etal~\cite{KimJKCK19} are the first to introduce data augmentation to generate more diverse domains. At the feature level, Xu \etal~\cite{XuWNTZ20} are the first to propose the prototype alignment, while Wu \etal~\cite{WuLHZ021} are the first to introduce the concept of feature disentanglement. In terms of optimization strategies, Cai \etal~\cite{CaiPNTDY19} are the first to introduce the teacher-student optimization strategy. Current research has focused on integrating different types of methods. For example, Muniret \etal~\cite{MunirKSA21} combine self-training optimization strategies with adversarial feature learning methods. Li \etal~\cite{LiDMLCWHKV22} integrate teacher-student optimization strategies, adversarial feature learning, and data augmentation. Considering that the DAOD problem setup may not align well with practical application scenarios, Wu \etal~\cite{WuCHWLMGWW022} propose a teacher-student optimization strategy to address the multi-source DAOD problem. Wu \etal~\cite{WuD22} also introduce a cyclic disentangled self-distillation method to address domain generalization in object detection. With the rise of visual large language models, recent work (\eg, Vidit \etal~\cite{ViditES23}) begins to focus on utilizing large models and has achieved remarkable results, indicating a future research trend.

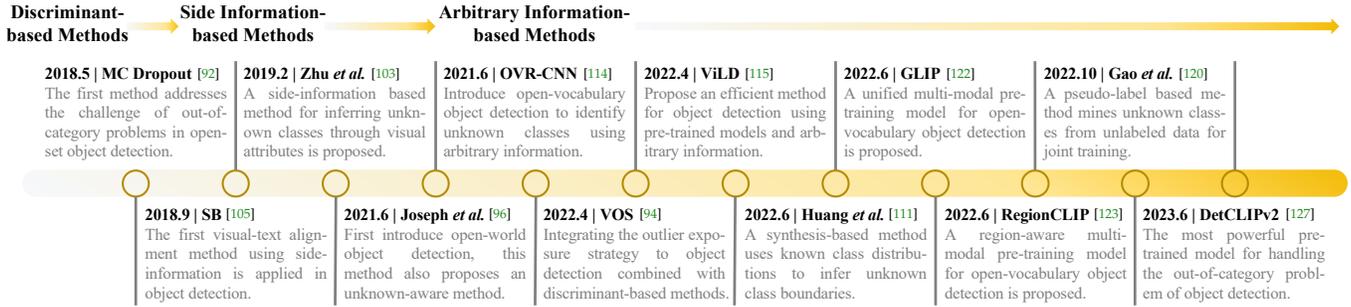
\begin{figure*}[!t]
 \centering
    \begin{overpic}[width=\textwidth,tics=8]{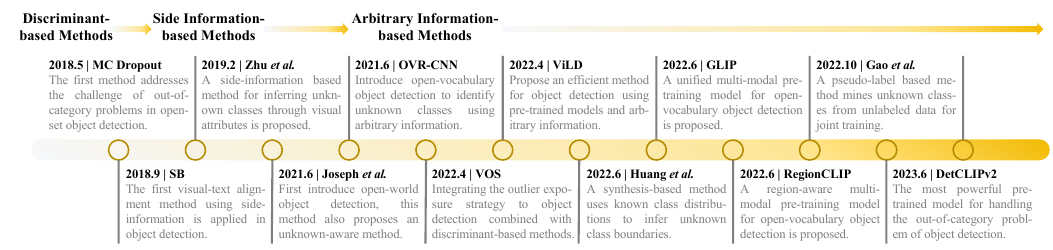}  
        % first line 
        \put (15.5,17.25) {\tiny \cite{miller2018dropout}}
        \put (28.2,17.25) {\tiny \cite{zhu2019zero}}
        \put (43.7,17.25) {\tiny \cite{zareian2021open}}
        \put (55.5,17.25) {\tiny \cite{guopen}}
        \put (70.2,17.25) {\tiny \cite{li2022grounded}}
        \put (87.2,17.25) {\tiny \cite{gao2022open}}
        % bottom line
        \put (17.6,6.95) {\tiny \cite{bansal2018zero}}
        \put (36.8,6.95) {\tiny \cite{joseph2021towards}}
        \put (47.8,6.95) {\tiny \cite{du2021vos}}
        \put (66,6.95) {\tiny \cite{huang2022robust}}
        \put (81,6.95) {\tiny \cite{zhong2022regionclip}}
        \put (95,6.95) {\tiny \cite{yao2023detclipv2}}
    \end{overpic}  
    \caption{Timeline for solving out-of-category challenges in object detection.}
    \label{fig:timeline-ooc}
\end{figure*}
\input{tables/ooc_summary}

\subsection{Detailed Categories}
In Tab.~\ref{solutions for ood}, we have outlined a detailed classification of existing methods. The solid in the figure represents the primary type of method, the hollow represents the secondary type, and the semi-hollow represents something in between. 
It can be observed that methods belonging to the data manipulation type are relatively few, mainly because it is challenging to form an end-to-end training framework. Considering it as a secondary type of assistance is a better choice. In the type of feature learning, the application of adversarial learning strategies is particularly widespread. There are fewer methods related to prototype alignment, mainly due to its simplicity and difficulty in serving as a universal method applied to different stages of object detection frameworks. Additionally, the sometimes challenging accurate characterization of distribution information by class prototypes leads to poor alignment effects. Feature disentanglement, having started relatively late, is also relatively scarce. In the optimization strategy type, self-training and teacher-student research are both prominent. Furthermore, it can be observed that adversarial learning, as an effective strategy, can be combined with a large number of methods.

%% file: tables/ooc_summary.tex
\begin{table*}
\caption{Detailed categorization of solutions addressing out-of-category challenges in object detection.}
\label{solutions-for-ooc}
\renewcommand{\arraystretch}{1.2}
\rowcolors{7}{gray!10}{white}
\setlength{\tabcolsep}{7pt}
\resizebox{\textwidth}{!}{%
    \begin{tabular}{lcccccccccc}
    \toprule
     & & \multicolumn{3}{c}{\cellcolor[HTML]{F9F1D3}Discriminant} & \multicolumn{3}{c}{\cellcolor[HTML]{FDE8A1}Side Information} & \multicolumn{3}{c}{\cellcolor[HTML]{F4BE0E}Arbitrary   Information} \\ \cmidrule(lr){3-5} \cmidrule(lr){6-8} \cmidrule(l){9-11}
    \multicolumn{1}{c}{\multirow{-2}{*}{Method}} & \multirow{-2}{*}{Paper List} & \begin{tabular}[c]{@{}c@{}}Sampling\\ \end{tabular} & \begin{tabular}[c]{@{}c@{}}Outlier\\ Exposure\end{tabular} & \begin{tabular}[c]{@{}c@{}}Unknown-aware\\ \end{tabular} & \begin{tabular}[c]{@{}c@{}}Attribute\\ \end{tabular} & \begin{tabular}[c]{@{}c@{}}Visual-text \\ Alignment\end{tabular} & \begin{tabular}[c]{@{}c@{}}Synthesis\\ \end{tabular} & \begin{tabular}[c]{@{}c@{}}Multi-modal \\ Alignment \end{tabular} & Pseudo-labeling & \begin{tabular}[c]{@{}c@{}}Pre-training \\ \end{tabular} \\ \midrule
    Monte Carlo Dropout~\cite{miller2018dropout} & ICRA 2018 & \CIRCLE  &   &   &    &    &    &    &    &   \\
    SB~\cite{bansal2018zero} & ECCV 2018 &    &    &    &    & \CIRCLE &    & \CIRCLE &    &    \\ 
    Miller \etal~\cite{miller2019evaluating} & ICRA 2019 & \CIRCLE &    &    &    &    &    &    &    &    \\ 
    Li \etal~\cite{li2019zero} & AAAI 2019 &    &    &    &    & \CIRCLE &    & \CIRCLE &    &    \\ 
    Zhu \etal~\cite{zhu2019zero} & TCSVT 2019 &    &    &    & \CIRCLE &    &    & \RIGHTcircle &    &    \\ 
    Mao \etal~\cite{mao2020zero} & TCSVT 2020 &    &    &    & \CIRCLE &    &    &   \RIGHTcircle &    &    \\
    Yan \etal~\cite{yan2020semantics} & TIP 2020 &    &    &    &    & \CIRCLE &    & \CIRCLE &    &    \\ 
    Joseph \etal~\cite{joseph2021towards} & CVPR 2021 &    &    & \CIRCLE &    &    &    &    &    &    \\ 
    OVR-CNN \etal~\cite{zareian2021open} & CVPR 2021 &    &    &    &    &    &    & \CIRCLE &    & \Circle \\ 
    VOS~\cite{du2021vos} & ICLR 2022 &    & \CIRCLE &    &    &    &    &    &    &    \\ 
    ViLD~\cite{guopen} & ICLR 2022 &    &    &    &    & \CIRCLE &    & \CIRCLE &    &    \\ 
    OpenDet~\cite{han2022expanding} & CVPR 2022 &    & \CIRCLE &    &    &    &    &    &    &    \\ 
    STUD~\cite{du2022unknown} & CVPR 2022 &    &    & \CIRCLE &    &    &    &    & \RIGHTcircle &    \\ 
    ProposalCLIP~\cite{shi2022proposalclip} & CVPR 2022 & \Circle &    & \CIRCLE &    &    &    &    & \RIGHTcircle &    \\ 
    OW-DETR~\cite{gupta2022ow} & CVPR 2022 &    &    & \CIRCLE &    &    &    &    & \RIGHTcircle &    \\ 
    Huang \etal~\cite{huang2022robust} & CVPR 2022 &    &    &    &    & \CIRCLE & \CIRCLE & \CIRCLE &    &    \\ 
    Ma \etal~\cite{ma2022open} & CVPR 2022 &    &    &    &    & \CIRCLE &    & \CIRCLE &    &    \\ 
    GLIP~\cite{li2022grounded} & CVPR 2022 &    &    &    &    & \CIRCLE &    & \CIRCLE &    & \CIRCLE \\ 
    RegionCLIP~\cite{zhong2022regionclip} & CVPR 2022 &    &    &    &    & \CIRCLE &    & \CIRCLE & \Circle & \CIRCLE \\ 
    DetPro~\cite{du2022learning} & CVPR 2022 &    &    &    &    & \CIRCLE &    &  \CIRCLE &    &    \\ 
    Liu \etal~\cite{liu2022open} & ECCV 2022 &    &    & \CIRCLE &    &    &    &    & \CIRCLE &    \\ 
    UC-OWOD~\cite{wu2022uc} & ECCV 2022 &    &    & \CIRCLE &    &    &    &    & \RIGHTcircle &    \\ 
    Gao \etal~\cite{gao2022open} & ECCV 2022 &    &    & \Circle &    & \CIRCLE &    & \CIRCLE &   \CIRCLE &    \\ 
    PromptDet~\cite{feng2022promptdet} & ECCV 2022 &    &    & \Circle &    & \CIRCLE &    & \CIRCLE & \CIRCLE &    \\ 
    DetCLIP~\cite{yao2022detclip} & NIPS 2022 &    &    &    & \Circle & \CIRCLE &    & \CIRCLE & \CIRCLE & \CIRCLE \\ 
    Nie \etal~\cite{nie2022node} & WACV 2022 &    &    &    &    & \CIRCLE &    & \CIRCLE &    &    \\ 
    Rahman \etal~\cite{rahman2020improved} & TNNLS 2022 &    &    &   & \Circle & \CIRCLE &    & \CIRCLE &    &    \\ 
    Yan \etal~\cite{yan2022semantics} & TPAMI 2022 &    &    &    &    & \CIRCLE &    & \CIRCLE &    &    \\ 
    F-VLM~\cite{kuo2023f} & ICLR 2023 &    &    &    &    & \CIRCLE &    & \CIRCLE &    &    \\ 
    Su \etal~\cite{su2023hsic} & ACM MM 2023 & \Circle &    & \CIRCLE &    &    &    &    &    &    \\ 
    Gupta \etal~\cite{gupta2023generative} & TPAMI 2023 &    &    &    &    & \CIRCLE & \CIRCLE & \CIRCLE &    &    \\ 
    Li \etal~\cite{li2023zero} & TIP 2023 &    &    &    &    & \CIRCLE & \CIRCLE &    \CIRCLE &    &    \\ 
    CORA~\cite{wu2023cora} & CVPR 2023 &    &    &    &    & \CIRCLE &   & \CIRCLE &    &    \\ 
    UniDetector~\cite{wang2023detecting} & CVPR 2023 &    &    & \RIGHTcircle &    & \CIRCLE &    & \CIRCLE & \Circle & \CIRCLE \\ 
    Ro-ViT~\cite{kim2023region} & CVPR 2023 & \Circle &    &    &    & \CIRCLE &    & \CIRCLE &    &    \\
    DetCLIPv2~\cite{yao2023detclipv2} & CVPR 2023 &    &    &    & \Circle & \CIRCLE &    & \CIRCLE & \CIRCLE & \CIRCLE \\ 
     \bottomrule
\end{tabular}
}
\end{table*}

%% file: suppl_sections/5-out-of-category.tex
\section{Out-of-Category Challenge}

\subsection{Timeline}

In Fig.~\ref{fig:timeline-ooc}, we present a timeline of object detection methods aimed at addressing out-of-category challenges. In the early stages, researchers use Monte Carlo dropout~\cite{miller2018dropout} techniques to infer the presence of unknown objects in object detection tasks. Bansal \etal introduce SB~\cite{bansal2018zero}, a concept for zero-shot object detection that leverages word embeddings to identify unknown categories. Subsequently, Zhu \etal~\cite{zhu2019zero} utilize visual attributes to deduce unknown categories. Joseph \etal~\cite{joseph2021towards} first propose the concept of open-world object detection, aiming to continuously discover and recognize unknown objects in the natural world. However, the information used by these methods has limitations. Zareian \etal~\cite{zareian2021open} introduce the concept of open-vocabulary object detection, aiming to leverage diverse information sources to identify unknown categories. Furthermore, several approaches, including ViLD~\cite{guopen} and Gao \etal~\cite{gao2022open}, suggest employing multi-modal pre-trained models to enhance object detectors' performance and their capability to identify unknown categories. Recent approaches focus on developing superior pre-training models by leveraging diverse multi-modal data, aiming to boost the target detector's performance, \eg, GLIP~\cite{li2022grounded}, RegionCLIP~\cite{zhong2022regionclip}, and DetCLIPv2~\cite{yao2023detclipv2}. This strategy is anticipated to be a key trend in future developments.

\subsection{Detailed Categories}

In Tab.~\ref{solutions-for-ooc}, we have outlined a detailed classification of existing methods, where solid represents the main types of methods, hollow represents secondary types, and semi-hollows represent something in between. Methods for addressing out-of-category challenges in object detection can be broadly classified into three main categories: discriminant-based, side-information-based, and arbitrary-information-based. Discriminant-based approaches encompass sampling-based, outlier exposure, and unknown-aware methods. Sampling-based methods, being among the earliest, have seen limited application, whereas unknown-aware based methods are more widely utilized. The side-information-based method primarily concentrates on visual-text alignment technologies, with certain approaches utilizing attributes or synthesis. The arbitrary information-based approach primarily relies on multi-modal alignment technology, with some methods also exploring the identification of unknown objects using pseudo-label technology. The current trend in addressing out-of-category issues involves pre-training more generalized open-vocabulary object detection models, some methods will combine other techniques, such as pseudo-labeling, to pre-train a better pre-trained model.

%% file: suppl_sections/6-malicious-data.tex
\section{Robust Learning Challenge}

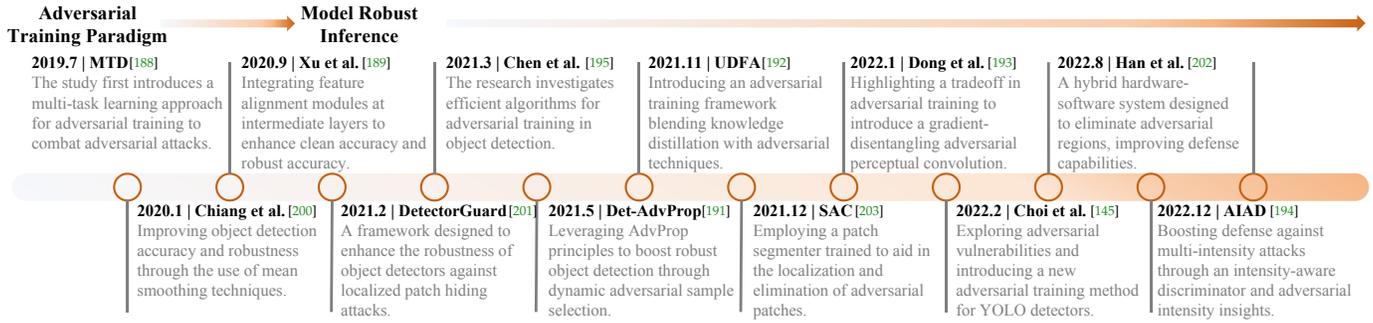
\begin{figure*}[!t]
 \centering
    \begin{overpic}[width=\textwidth,tics=8]{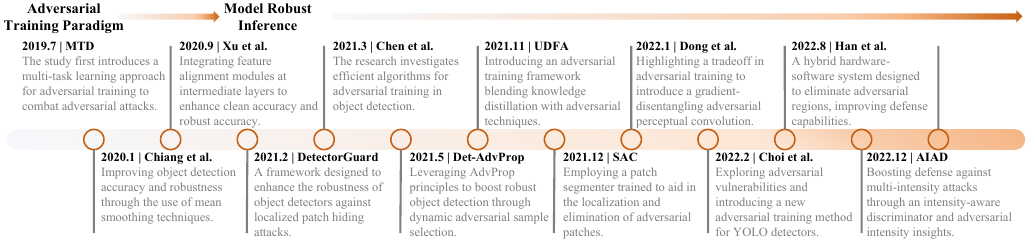}  
        % first line 
        \put (9.0,18.72) {\tiny \cite{zhang2019towards}}
        \put (26.0,18.72) {\tiny \cite{xu2021using}}
        \put (42.3,18.72) {\tiny \cite{chen2021class}}
        \put (55.2,18.72) {\tiny \cite{xu2022robust}}
        \put (71.7,18.72) {\tiny \cite{dong2022adversarially}}
        \put (86.3,18.72) {\tiny \cite{han2022real}}
        % bottom line
        \put (20.6,7.93) {\tiny \cite{chiang2020detection}}
        \put (36.7,7.93) {\tiny \cite{xiang2021detectorguard}}
        \put (50.8,7.93) {\tiny \cite{chen2021robust}}
        \put (62,7.93) {\tiny \cite{liu2022segment}}
        \put (79.2,7.93) {\tiny \cite{im2022adversarial}}
        \put (92.3,7.93) {\tiny \cite{chengadversarial}}
    \end{overpic}  
    \caption{Timeline for solving robust learning challenges in object detection.}
    \label{fig:timeline-malicious}
\end{figure*}
\input{tables/malicious_data_summary}

\subsection{Timeline}
Fig.~\ref{fig:timeline-malicious} shows the evolution timeline in robust learning within the domain of object detection. Initially, the domain sees the adaptation of adversarial training methods, specifically designed for classification tasks, to object detection through the multitask learning approach with the introduction of MTD~\cite{zhang2019towards}. Following this, Chiang \etal~\cite{chiang2020detection} improve the accuracy and robustness of the model through the application of average smoothing techniques, categorizing these advances as model robust inference. Soon after, Xu \etal~\cite{xu2021using} pioneer an adversarial training approach that incorporates an alignment module within the intermediate layer of the model, with the objective of increasing both precision and robustness. The introduction of DetectorGuard~\cite{xiang2021detectorguard} marks a significant step forward in improving the resilience of object detectors to local occlusion attacks. By 2021, the field will witness exponential growth in robust incremental learning methods, with Chen \etal~\cite{chen2021class} exploring efficient robust training algorithms for object detection. Det-AdvProp~\cite{chen2021robust} takes advantage of the AdvProp principle to fortify object detection robustness through dynamic adversarial sample selection. The UDFA framework~\cite{xu2022robust} brings knowledge refinement strategies into adversarial training, while SAC~\cite{liu2022segment} proposes a patch splitter technique, echoing DetectorGuard's original aim ~\cite{xiang2021detectorguard} to mitigate adversarial patch effects. The year 2022 sees further advancements, including Dong \etal~\cite{dong2022adversarially} introducing a balanced adversarial training method through gradient decoupling of adversarial perceptual convolutions. Choi \etal~\cite{im2022adversarial} unveil a novel adversarial training strategy that targets vulnerabilities in the YOLO detector. Moreover, Han \etal~\cite{han2022real} develop a hybrid hardware-software system dedicated to neutralizing adversarial regions, thus improving defense mechanisms. Recent efforts like AIAD~\cite{chengadversarial} focus on strengthening defenses against multi-intensity attacks, improving performance through intensity-aware discriminators and adversarial intensity insights. These milestones underscore the ongoing enhancements in adversarial attack defense within robust learning for object detection, hinting at future directions toward more efficient, generalized training methodologies, and robust inference mechanisms with enhanced utility.

\subsection{Solutions}

Tab.~\ref{malicious data summary} presents the diversity of technology applications and strategic evolution within the realm of object detection against malicious data attacks and defenses. Through the different markings of solid, semi-hollow, and hollow circles, the table reveals the dominance and relative importance of various approaches in research, providing an intuitive understanding of different techniques and their roles in the defense system. First, the techniques for adversarial attacks are evolving rapidly and their breadth of research and application far exceeds that of backdoor attacks. Adversarial attacks are usually more refined and difficult to detect, probably because they can take effect during the inference phase of the model, whereas backdoor attacks are usually implanted during the model training phase. Second, adversarial defense strategies also show positive development, reflecting the ability of researchers to respond quickly and update their defense mechanisms after discovering novel attacks. This suggests that despite the increasing sophistication of attack methods, defense techniques are also advancing, aiming to ensure the security and reliability of models. Finally, the development of adversarial-attack and defense techniques reveals a dynamic balance in the field, where defense strategies are upgraded accordingly as attack methods advance, creating an evolving security ecosystem. However, despite the continuous technological advances, we still need to be concerned about the ethical and security risks that may arise in research and applications to ensure that technological advances are not exploited for malicious purposes.

%% file: tables/malicious_data_summary.tex
% \vspace{-3mm}
\begin{table*}[!t]
\caption{Detailed categorization of solutions addressing robust learning challenges in object detection.}
\label{malicious data summary}
% \vspace{-3mm}
\centering

\resizebox{\textwidth}{!}{%
\setlength{\tabcolsep}{9pt}
\renewcommand{\arraystretch}{1.1}
\rowcolors{9}{gray!10}{white}
\begin{tabular}{lccccccccccccc}

\toprule
 &  & \multicolumn{2}{c}{\cellcolor[HTML]{5CCCCC}Backdoor Attack} & \multicolumn{4}{c}{\cellcolor[HTML]{5CCCCC}Adversarial Attack} & \multicolumn{2}{c}{\cellcolor[HTML]{5CCCCC}Adversarial Scenario} & \multicolumn{2}{c}{\cellcolor[HTML]{5CCCCC}Knowledge Type} & \multicolumn{2}{c}{\cellcolor[HTML]{F5B889}Defense Method} \\ \cmidrule(l){3-4} \cmidrule(l){5-8} \cmidrule(l){9-10}  \cmidrule(l){11-12} \cmidrule(l){13-14}
\multicolumn{1}{c}{\multirow{-2}{*}{Method}} & \multirow{-2}{*}{Paper List} & \begin{tabular}[c]{@{}c@{}}Clean-\\ label\end{tabular} & Poison & \begin{tabular}[c]{@{}c@{}}Pixel \\ Attack\end{tabular} & \begin{tabular}[c]{@{}c@{}}Planar \\ Attack\end{tabular} & \begin{tabular}[c]{@{}c@{}}Non-Planar \\ Attack\end{tabular} & \begin{tabular}[c]{@{}c@{}}Adversarial \\ Camouflage\end{tabular} & Digital & Physical & \begin{tabular}[c]{@{}c@{}}White-\\ box\end{tabular} & \begin{tabular}[c]{@{}c@{}}Black-\\ box\end{tabular} & \begin{tabular}[c]{@{}c@{}}Adversarial Training \\ Paradigm\end{tabular}& \begin{tabular}[c]{@{}c@{}}Model Robust \\
Inference\end{tabular} \\ \midrule
\multicolumn{14}{c}{\textbf{Challenge in Malicious Data}} \\

DAG~\cite{xie2017adversarial} & ICCV 2017 &  &  & \CIRCLE &  &  &  & \CIRCLE &  & \CIRCLE &  &  &  \\
Gurbaxani \etal~\cite{gurbaxani2018traits} & ARXIV 2018 &  &  & \CIRCLE &  &  &  & \CIRCLE &  & \CIRCLE & \CIRCLE &  &  \\
ShapeShifter~\cite{chen2019shapeshifter} & ECML PKDD 2018 &  &  &  &  &  & \CIRCLE &  & \CIRCLE & \CIRCLE &  &  &  \\
Wei \etal~\cite{wei2018transferable} & IJCAI 2019 &  &  & \CIRCLE &  &  &  & \CIRCLE &  & \RIGHTcircle & \CIRCLE &  &  \\
Liu \etal~\cite{liu2019perceptual} & AAAI 2019 &  &  &  & \CIRCLE &  &  & \CIRCLE &  & \RIGHTcircle & \CIRCLE &  &  \\
Lee \etal~\cite{lee2019physical} & ARXIV 2019 &  &  &  & \CIRCLE &  &  & \CIRCLE & \CIRCLE & \CIRCLE &  &  &  \\
Zhao \etal~\cite{zhao2019seeing} & SIGSAV 2019 &  &  &  &  & \CIRCLE &  &  & \CIRCLE & \RIGHTcircle & \CIRCLE &  &  \\
Zhang \etal~\cite{zhang2019camou} & ICLR 2019 &  &  &  &  &  & \CIRCLE &  & \CIRCLE &  & \CIRCLE &  &  \\
Liao \etal~\cite{liao2020category} & ARXIV 2020 &  &  & \CIRCLE &  &  &  & \CIRCLE &  & \CIRCLE & \Circle &  &  \\
Zhao \etal~\cite{zhao2020object} & ARXIV 2020 &  &  &  & \CIRCLE &  &  & \CIRCLE &  & \RIGHTcircle & \CIRCLE &  &  \\
Bao \etal~\cite{bao2020sparse} & CIKM 2020 &  &  &  & \CIRCLE &  &  & \CIRCLE &  & \RIGHTcircle & \CIRCLE &  &  \\
Wu \etal~\cite{wu2020dpattack} & CIKM 2020 &  &  &  & \CIRCLE &  &  & \CIRCLE &  & \RIGHTcircle & \CIRCLE &  &  \\
Saha \etal~\cite{saha2020role} & CVPR 2020 &  &  &  & \CIRCLE &  &  & \CIRCLE &  & \CIRCLE &  &  &  \\
Huang \etal~\cite{huang2020universal} & CVPR 2020 &  &  &  &  & \CIRCLE &  &  & \CIRCLE & \CIRCLE &  &  &  \\
Xu \etal~\cite{xu2020adversarial} & ECCV 2020 &  &  &  &  & \CIRCLE &  &  & \CIRCLE & \CIRCLE &  &  &  \\
Wu \etal~\cite{wu2020making} & ECCV 2020 &  &  &  &  & \CIRCLE &  &  & \CIRCLE & \RIGHTcircle & \CIRCLE &  &  \\
Hoory \etal~\cite{hoory2020dynamic} & ARXIV 2020 &  &  &  &  & \CIRCLE &  &  & \CIRCLE & \RIGHTcircle & \CIRCLE &  &  \\
Wu \etal~\cite{wu2020physical} & ARXIV 2020 &  &  &  &  &  & \CIRCLE &  & \CIRCLE &  & \CIRCLE &  &  \\
Lin \etal~\cite{lin2020composite} & SIGSAC 2020 &  & \CIRCLE &  &  &  &  & \CIRCLE &  &  & \CIRCLE &  &  \\
Wang \etal~\cite{wang2021dual} & CYB IEEE 2021 &  &  & \CIRCLE &  &  &  & \CIRCLE & \Circle & \CIRCLE &  &  &  \\
Nezami \etal~\cite{nezami2021pick} & CVIU 2021 &  &  & \CIRCLE &  &  &  & \CIRCLE &  & \CIRCLE &  &  &  \\
Liang \etal~\cite{liang2022parallel} & ICCV 2021 &  &  & \CIRCLE &  &  &  & \CIRCLE &  & \RIGHTcircle & \CIRCLE &  &  \\
Zhang \etal~\cite{zhang2021adversarial} & ICML(w) 2021 &  &  &  & \CIRCLE &  &  & \CIRCLE &  & \CIRCLE &  &  &  \\
Zhu \etal~\cite{zhu2021you} & ARXIV 2021 &  &  &  & \CIRCLE &  &  & \CIRCLE &  & \CIRCLE &  &  &  \\
Zolfi \etal~\cite{zolfi2021translucent} & CVPR 2021 &  &  &  & \CIRCLE &  &  & \CIRCLE &  & \RIGHTcircle & \CIRCLE &  &  \\
Zhou \etal~\cite{zhou2021multi} & IJCAI 2021 &  & \CIRCLE &  &  &  &  & \CIRCLE &  & \CIRCLE &  &  &  \\
Choi \etal~\cite{im2022adversarial} & ARXIV 2022 &  &  & \CIRCLE &  &  &  & \CIRCLE &  & \CIRCLE &  &  & \CIRCLE \\
Yin \etal~\cite{yin2022adc} & WACV 2022 &  &  & \CIRCLE &  &  &  &  &  & \CIRCLE & \Circle &  &  \\
Cai \etal~\cite{cai2022context} & AAAI 2022 &  &  & \CIRCLE &  &  &  & \CIRCLE &  & \RIGHTcircle & \CIRCLE &  &  \\
Cai \etal~\cite{cai2022zero} & CVPR 2022 &  &  & \CIRCLE &  &  &  & \CIRCLE &  & \Circle & \CIRCLE &  &  \\
Xia \etal~\cite{xia2022ssmi} & ARXIV 2022 &  &  & \CIRCLE &  &  &  & \CIRCLE &  & \CIRCLE &  &  &  \\
Liang \etal~\cite{liang2022large} & ECCV 2022 &  &  & \CIRCLE &  &  &  & \CIRCLE &  & \Circle & \CIRCLE &  &  \\
Pavlitskaya \etal~\cite{pavlitskaya2022suppress} & ICECCME 2022 &  &  &  & \CIRCLE &  &  & \CIRCLE &  & \CIRCLE &  &  &  \\
Pavlitskaya \etal~\cite{pavlitskaya2022feasibility} & IJCAI(w) 2022 &  &  &  & \CIRCLE &  &  & \CIRCLE &  & \CIRCLE &  &  &  \\
Hartnett \etal~\cite{hartnett2022empirical} & ARXIV 2022 &  &  &  & \CIRCLE &  &  & \CIRCLE &  & \RIGHTcircle & \CIRCLE &  &  \\
Labarbarie \etal~\cite{labarbarie2022benchmarking} & IJCAI(w) 2022 &  &  &  & \CIRCLE &  &  & \CIRCLE &  & \CIRCLE &  &  &  \\
Maesumi \etal~\cite{maesumi2021learning} & ARXIV 2022 &  &  &  &  & \CIRCLE &  &  & \CIRCLE & \RIGHTcircle & \CIRCLE &  &  \\
Du \etal~\cite{du2022physical} & WACV 2022 &  &  &  &  & \CIRCLE &  &  & \CIRCLE & \CIRCLE &  &  &  \\
Shapira \etal~\cite{shapira2022attacking} & ARXIV 2022 &  &  &  &  & \CIRCLE &  &  & \CIRCLE & \RIGHTcircle & \CIRCLE &  &  \\
Duan \etal~\cite{duan2021learning} & IJCAI 2022 &  &  &  &  &  & \CIRCLE &  & \CIRCLE & \CIRCLE &  &  &  \\
Zhang \etal~\cite{zhang2022transferable} & ARXIV 2022 &  &  &  &  &  & \CIRCLE &  & \CIRCLE & \RIGHTcircle & \CIRCLE &  &  \\
Hu \etal~\cite{hu2022adversarial} & CVPR 2022 &  &  &  &  &  & \CIRCLE &  & \CIRCLE & \CIRCLE &  &  &  \\
Chan \etal~\cite{chan2023baddet} & ECCV(w) 2022 &  & \CIRCLE &  &  &  &  & \CIRCLE &  &  & \CIRCLE &  &  \\
Ma \etal~\cite{ma2022macab} & ARXIV 2022 & \CIRCLE &  &  &  &  &  &  & \CIRCLE &  & \CIRCLE &  &  \\
Ma \etal~\cite{ma2022dangerous} & ARXIV 2022 &  & \CIRCLE &  &  &  &  &  & \CIRCLE &  & \CIRCLE &  &  \\
Shapira \etal~\cite{shapira2023phantom} & WACV 2023 &  &  & \CIRCLE &  &  &  & \CIRCLE &  & \CIRCLE &  &  &  \\
Huang \etal~\cite{huang2023t} & CVPR 2023 &  &  & \CIRCLE &  &  &  & \CIRCLE &  & \RIGHTcircle & \CIRCLE &  &  \\
Wu \etal~\cite{wu2022adversarial} & IEEE IV 2023 &  &  &  & \CIRCLE &  &  & \CIRCLE &  & \RIGHTcircle & \CIRCLE &  &  \\
Sun \etal~\cite{sun2023differential} & NN 2023 &  &  &  &  &  & \CIRCLE &  & \CIRCLE &  & \CIRCLE &  &  \\
Luo \etal~\cite{luo2022untargeted} & ICASSP 2023 &  & \CIRCLE &  &  &  &  & \CIRCLE & \RIGHTcircle &  & \CIRCLE &  &  \\ 
\bottomrule
\multicolumn{14}{c}{\textbf{Adversarial Defense}} \\
JPG~\cite{dziugaite2016study} & ARXIV 2016 &  &  & \CIRCLE &  &  &  &  &  &  &  &  & \CIRCLE \\
SP~\cite{xu2017feature} & ARXIV 2017 &  &  & \CIRCLE &  &  &  &  &  &  &  &  & \CIRCLE \\
MTD~\cite{zhang2019towards} & ICCV 2019 &  &  & \CIRCLE &  &  &  &  &  &  &  & \CIRCLE &  \\
LGS~\cite{naseer2019local} & WACV 2019 &  &  & \CIRCLE &  &  &  &  &  &  &  &  & \CIRCLE \\
Chiang \etal~\cite{chiang2020detection} & NeurIPS 2020 &  &  & \CIRCLE &  &  &  &  &  &  &  &  & \CIRCLE \\
DetectorGuard~\cite{xiang2021detectorguard} & CCS 2021 &  &  &  & \CIRCLE &  &  &  &  &  &  &  & \CIRCLE \\
Xu \etal~\cite{xu2021using} & ICIP 2021 &  &  & \CIRCLE &  &  &  &  &  &  &  & \CIRCLE &  \\
Det-AdvProp~\cite{chen2021robust} & CVPR 2021 &  &  & \CIRCLE &  &  &  &  &  &  &  & \CIRCLE &  \\
Chen \etal~\cite{chen2021class} & CVPR 2021 &  &  & \CIRCLE &  &  &  &  &  &  &  & \CIRCLE &  \\
APM \etal~\cite{chiang2021adversarial} & ACMMM 2021 &  &  & \CIRCLE &  &  &  &  &  &  &  &  & \CIRCLE \\
SAC~\cite{liu2022segment} & CVPR 2022 &  &  &  & \CIRCLE &  &  &  &  &  &  &  & \CIRCLE \\
Choi \etal~\cite{im2022adversarial} & IV 2022 &  &  & \CIRCLE &  &  &  &  &  &  &  & \CIRCLE &  \\
UDFA~\cite{xu2022robust} & ICIP 2022 &  &  & \CIRCLE &  &  &  &  &  &  &  & \CIRCLE &  \\
Dong \etal~\cite{dong2022adversarially} & ECCV 2022 &  &  & \CIRCLE &  &  &  &  &  &  &  & \CIRCLE &  \\
AIAD~\cite{chengadversarial} & SSRN 2022 &  &  & \CIRCLE &  &  &  &  &  &  &  & \CIRCLE &  \\
Han \etal~\cite{han2022real} & TCAD 2023 &  &  &  & \CIRCLE &  &  &  &  &  &  &  & \CIRCLE \\
Z-Mask~\cite{rossolini2023defending} & AAAI 2023 &  &  &  & \CIRCLE &  &  &  &  &  &  &  & \CIRCLE \\
Li \etal~\cite{li2023importance} & ARXIV 2023 &  &  & \CIRCLE &  &  &  &  &  &  &  & \CIRCLE &  \\
\bottomrule
\end{tabular}%
}
\end{table*}

%% file: suppl_sections/7-incremental.tex
\section{Incremental Learning Challenge}

\begin{figure*}[t]
 \centering
    \begin{overpic}[width=\textwidth]{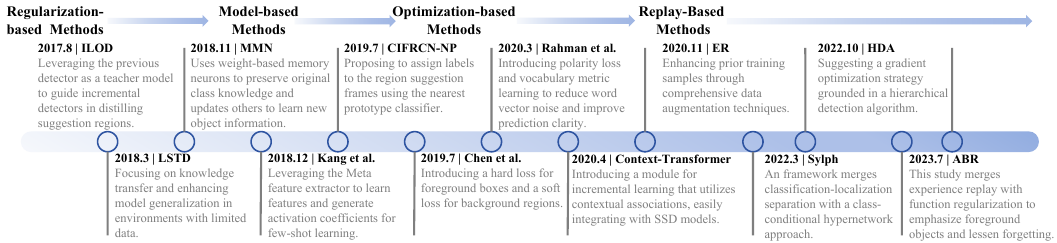}  
        % first line 
        \put (10.6,18.12) {\tiny \cite{shmelkov2017incremental}}
        \put (25.6,18.12) {\tiny \cite{li2018incremental}}
        \put (43.1,18.12) {\tiny \cite{hao2019end}}
        \put (58.2,18.12) {\tiny \cite{rahman2020any}}
        \put (69.0,18.12) {\tiny \cite{shieh2020continual}}
        \put (84.6,18.12) {\tiny \cite{she2022fast}}
        % bottom line
        \put (18.0,7.65) {\tiny \cite{chen2018lstd}}
        \put (35.4,7.65) {\tiny \cite{kang2019few}}
        \put (49.25,7.65) {\tiny \cite{chen2019new}}
        \put (69,7.65) {\tiny \cite{yang2020context}}
        \put (79.2,7.65) {\tiny \cite{yin2022sylph}}
        \put (92.6,7.65) {\tiny \cite{yuyang2023augmented}}
    \end{overpic}  
    \caption{Timeline for solving incrimental learning challenges in object detection.}
    \label{fig:timeline-incrimental}
\end{figure*}

\input{tables/solutions_for_increment_data}
\subsection{Timeline}
Fig.~\ref{fig:timeline-incrimental} illustrates the ongoing evolution of incremental learning in target detection, highlighting significant advances. Currently, regularization-based strategies, highlighted by ILOD~\cite{shmelkov2017incremental} with its distillation framework, balance new and pre-existing knowledge. Following this, LSTD~\cite{chen2018lstd} introduces a model-based approach that uses model decomposition to facilitate knowledge transfer from the older model to the newer one using minimal annotations for new categories, thus increasing adaptability. In development, optimization-based methods come to the fore, with CTFRCN-NP~\cite{hao2019end} leading with a closest prototype classifier that improves label assignment and significantly increases the discriminative capacity of the model. Soon after, Kang \etal~\cite{kang2019few} introduce meta-learning, utilizing a meta-feature extractor that helps to learn from small samples. In addition, playback-based methodologies such as ER~\cite{shieh2020continual} and Context-Transformer~\cite{yang2020context} offer fresh insights into incremental learning. ER improves learning through extensive data augmentation, while Context-Transformer, integrated with SSD models, strengthens contextual learning, providing strategies for effectively preventing catastrophic forgetting.

In 2022, HDA~\cite{she2022fast} has unveiled a gradient optimization strategy within hierarchical detection algorithms, marrying optimization and model structure design to handle more complex detection tasks. By 2023, ABR~\cite{yuyang2023augmented} merges experience playback with functional regularization, focusing more on foreground objects and minimizing forgetting, indicating a shift towards improving the model's adaptability to novel information. These progressions highlight a transition from isolated strategies to an amalgamation of various techniques for incremental learning in target detection. This trend towards multi-strategy integration reflects the drive to counteract forgetting, enhance flexibility, and optimize performance. Signals the field's move towards selecting the most suitable incremental learning strategies based on the distinct needs of the model and specific application scenarios, promising more efficient, accurate, and reliable target detection models.

\subsection{Solutions}

Tab.~\ref{solutions for incremental data} presents an in-depth overview of the incremental learning landscape in target detection, outlining key methodologies and their significance in various research endeavors. Solid circles in the table highlight primary techniques emphasized by researchers, whereas semi-hollow and hollow circles represent secondary or supportive methods. The table uncovers an initial reliance on replay and regularization methods in early incremental learning techniques, such as ILOD~\cite{shmelkov2017incremental} and MMN~\cite{li2018incremental}, which prioritized empirical playback and functional regularization to prevent old knowledge from being overshadowed by new information, ensuring stability and continuity in model learning. As incremental learning evolved, model-based and optimization-based strategies gained traction, signifying a shift toward advanced network architecture adjustments and meta-learning approaches. For instance, Kang \etal~\cite{kang2019few}'s proposition of using Meta feature extractors for feature learning not only enhances the generalization ability of the model, but also provides a more flexible way of dealing with new categories of data.

The current technological trajectory reveals a preference for hybrid strategies that combine various methods to minimize forgetting, combining diverse playback and regularization techniques with gradient projection and meta-learning to refine model structure and efficacy. This shift reflects the research community's pursuit of more intricate solutions to accommodate expanding datasets and increasingly complex scenarios.

In general, the trend of incremental learning for target detection has transitioned from straightforward memory retention tactics to elaborate network optimization and functional integration. Future research might investigate deeper model introspection and adaptive tuning methods, enabling models to autonomously adjust to fluctuating data streams. Additionally, leveraging unsupervised and semi-supervised learning could further strengthen model robustness and adaptability, pushing the boundaries of current methodologies.

%% file: tables/solutions_for_increment_data.tex
\begin{table*}[t!]
\caption{Detailed categorization of solutions addressing incremental learning challenges in object detection.}
\label{solutions for incremental data}
\renewcommand{\arraystretch}{1.2}
\rowcolors{7}{gray!10}{white}
\setlength{\tabcolsep}{7pt}
\resizebox{\textwidth}{!}{%
\begin{tabular}{lccccccccccc}
\toprule
 &  & \multicolumn{3}{c}{\cellcolor[HTML]{EDF1FA}Replay} & \multicolumn{4}{c}{\cellcolor[HTML]{D2DDF2}Model} & \multicolumn{2}{c}{\cellcolor[HTML]{ADC0E8}Regularization} & \cellcolor[HTML]{94AFE1}Optimization \\
 \cmidrule(l){3-5} \cmidrule(l){6-9} \cmidrule(l){10-11} \cmidrule(l){12-12}
\multicolumn{1}{c}{Method} & Paper List & \begin{tabular}[c]{@{}c@{}}Experience\\ Replay\end{tabular} & \begin{tabular}[c]{@{}c@{}}Feature\\ Replay\end{tabular}  & \begin{tabular}[c]{@{}c@{}}Generation\\ Replay\end{tabular} & \begin{tabular}[c]{@{}c@{}}Parameter\\ Isolation\end{tabular} & \begin{tabular}[c]{@{}c@{}}Model \\ Decomposition\end{tabular} & \begin{tabular}[c]{@{}c@{}}Modular\\ Network\end{tabular} & Representation & \begin{tabular}[c]{@{}c@{}}Weight\\ Regularization\end{tabular} & \begin{tabular}[c]{@{}c@{}}Structural\\ Regularization\end{tabular}  &  \\ \midrule

ILOD~\cite{shmelkov2017incremental} & ICCV 2017 &  &  &  & \RIGHTcircle &  &  &  &  & \CIRCLE &  \\
MMN~\cite{li2018incremental} & INFOCOM 2018 &  &  &  & \CIRCLE &  &  &  &  &  &  \\
LSTD~\cite{chen2018lstd} & AAAI 2018 &  &  &  &  & \CIRCLE &  &  & \RIGHTcircle &  &  \\
CIFRCN-NP~\cite{hao2019end} & ICME 2019 &  & \CIRCLE &  &  &  &  &  &  & \CIRCLE &  \\
Kang \etal~\cite{kang2019few} & ICCV 2019 &  &  &  &  & \CIRCLE &  &  &  &  & \RIGHTcircle \\
RILOD~\cite{li2019rilod} & SEC 2019 &  &  &  &  &  &  &  &  & \CIRCLE &  \\
Chen \etal~\cite{chen2019new} & IJCNN 2019 &  &  &  &  &  &  &  & \CIRCLE & \CIRCLE &  \\
ER~\cite{shieh2020continual} & SENSORS 2020 & \CIRCLE &  &  &  &  &  &  & \CIRCLE &  &  \\
RODEO~\cite{acharya2020rodeo} & BMVC 2020 & \CIRCLE &  &  &  &  &  &  &  &  &  \\
ONCE~\cite{perez2020incremental} & CVPR 2020 &  &  &  &  & \CIRCLE &  &  &  &  & \RIGHTcircle \\
Context-Transformer~\cite{yang2020context} & AAAI 2020 &  &  &  &  &  & \CIRCLE &  &  &  &  \\
Rahman \etal~\cite{rahman2020any} & ACCV 2020 &  &  &  &  &  &  & \CIRCLE & \CIRCLE &  &  \\
Faster ILOD~\cite{peng2020faster} & PR 2020 &  &  &  &  &  &  &  &  & \CIRCLE &  \\
ORE~\cite{joseph2021towards} & CVPR 2021 & \CIRCLE &  &  &  &  &  &  &  &  &  \\
iMTFA~\cite{ganea2021incremental} & CVPR 2021 &  & \CIRCLE &  &  &  &  &  & \RIGHTcircle &  &  \\
DualFusion~\cite{tambwekar2021few} & ICCV 2021 &  &  &  &  & \CIRCLE &  & \CIRCLE &  &  &  \\
Expert Detector~\cite{zhang2021incremental} & ICOIAS 2021 &  &  &  &  &  & \CIRCLE &  &  &  &  \\
Teo \etal~\cite{li2021towards} & ARXIV 2021 &  & \CIRCLE &  & \CIRCLE &  &  &  & \CIRCLE &  & \CIRCLE \\
IncDet~\cite{liu2020incdet} & TNNLS 2021 &  &  &  &  &  &  &  & \CIRCLE & \CIRCLE &  \\
Dong \etal~\cite{dong2021bridging} & NIPS 2021 &  &  &  &  &  &  &  &  & \CIRCLE &  \\
LEAST~\cite{li2021class} & ARXIV 2021 & \CIRCLE &  &  & \CIRCLE &  &  &  &  & \CIRCLE &  \\
SID~\cite{peng2021sid} & CVIU 2021 &  &  &  &  &  &  &  &  & \CIRCLE &  \\
PPAS~\cite{zhou2020lifelong} & ARXIV 2021 &  & \CIRCLE &  &  &  &  &  &  & \CIRCLE &  \\
Joseph \etal~\cite{joseph2021incremental} & TPAMI 2021 &  & \RIGHTcircle &  &  &  &  &  &  &  & \CIRCLE \\
Shieh \etal~\cite{shieh2022utilizing} & MVA 2022 & \CIRCLE &  &  &  & \CIRCLE &  &  & \CIRCLE &  &  \\
RT-Net~\cite{cui2023rt} & APIN 2022 &  & \CIRCLE & \CIRCLE &  &  & \CIRCLE &  &  & \CIRCLE &  \\
OW-DETR~\cite{gupta2022ow} & CVPR 2022 &  &  &  &  &  & \CIRCLE & \CIRCLE &  &  &  \\
Rosetta~\cite{yang2022continual} & CVPR 2022 &  & \CIRCLE &  &  &  & \CIRCLE &  &  &  &  \\
Incremental-DETR~\cite{dong2023incremental} & ARXIV 2022 &  &  &  & \RIGHTcircle &  &  & \CIRCLE &  &  &  \\
BRS~\cite{cui2022balanced} & ICASSP 2022 & \CIRCLE &  &  &  &  &  &  & \CIRCLE &  &  \\
MVCD~\cite{yang2022multi} & PR 2022 &  &  &  &  &  &  &  &  & \CIRCLE &  \\
MMA~\cite{cermelli2022modeling} & CVPR 2022 &  &  &  &  &  &  &  & \CIRCLE & \CIRCLE &  \\
ERD~\cite{feng2022overcoming} & CVPR 2022 &  &  &  &  &  &  &  & \CIRCLE & \CIRCLE &  \\
iFSOD~\cite{cheng2021meta} & TCSVT 2022 &  &  &  &  &  &  &  &  &  & \CIRCLE \\
iFS-RCNN~\cite{nguyen2022ifs} & CVPR 2022 &  &  &  &  &  &  &  & \CIRCLE &  &  \\
Sylph~\cite{yin2022sylph} & CVPR 2022 &  &  & \CIRCLE &  & \CIRCLE &  &  & \CIRCLE &  & \CIRCLE \\
HDA~\cite{she2022fast} & IROS 2022 &  &  &  & \CIRCLE &  &  &  &  &  & \CIRCLE \\
Verwimp \etal~\cite{verwimp2022re} & BMVC 2022 &  &  &  &  &  &  &  & \CIRCLE & \CIRCLE &  \\
OSR~\cite{yang2023one} & AAAI 2023 & \CIRCLE &  &  &  &  &  &  &  & \CIRCLE &  \\
DIODE~\cite{peng2023diode} & PR 2023 &  &  &  & \CIRCLE &  &  &  & \CIRCLE &  &  \\
CL-DETR~\cite{liu2023continual} & CVPR 2023 & \CIRCLE &  &  &  &  &  &  &  & \CIRCLE &  \\
ABR~\cite{yuyang2023augmented} & ICCV 2023 & \CIRCLE &  &  &  &  &  &  &  & \RIGHTcircle &  \\
\bottomrule
\end{tabular}
}
\end{table*}

%% file: suppl_sections/8-benchmark.tex
\section{Benchmarks and Evaluation}
\label{sec:benchmark_suppl}

\subsection{Out-of-Domain Solution Benchmarks}

In Tab. \ref{solutions for ood}, we have included three datasets to evaluate the performance of the DAOD methods using ResNet-101 as the backbone. It can be observed that there are few existing methods conducting experiments on these datasets using ResNet-101 as the backbone, and adversarial learning-based DAOD methods perform better.

\subsection{Out-of-Category Solution Benchmarks}

In this part, we benchmark the methods for out-of-category challenges.

% \subsubsection{Open-Set Object Detection}

% \textbf{Experimental Setting} In the experimental setup of the OpenDet method, the PASCAL VOC dataset is employed as known category data for training, while the subset of the MS-COCO dataset is chosen to represent unknown categories for testing. As for the evaluation metric, the Wilderness Impact (WI)~\cite{dhamija2020overlooked} is used to measure the percentage of unknown categories misclassified into known categories, the Absolute Open-Set Error (AOSE)~\cite{miller2018dropout} is used to count the number of misclassified unknown category objects, and mAP is used to measure the ability to detect known objects and unknown objects.

\begin{figure}[ht]
 \centering
  \includegraphics[width=0.48\textwidth]{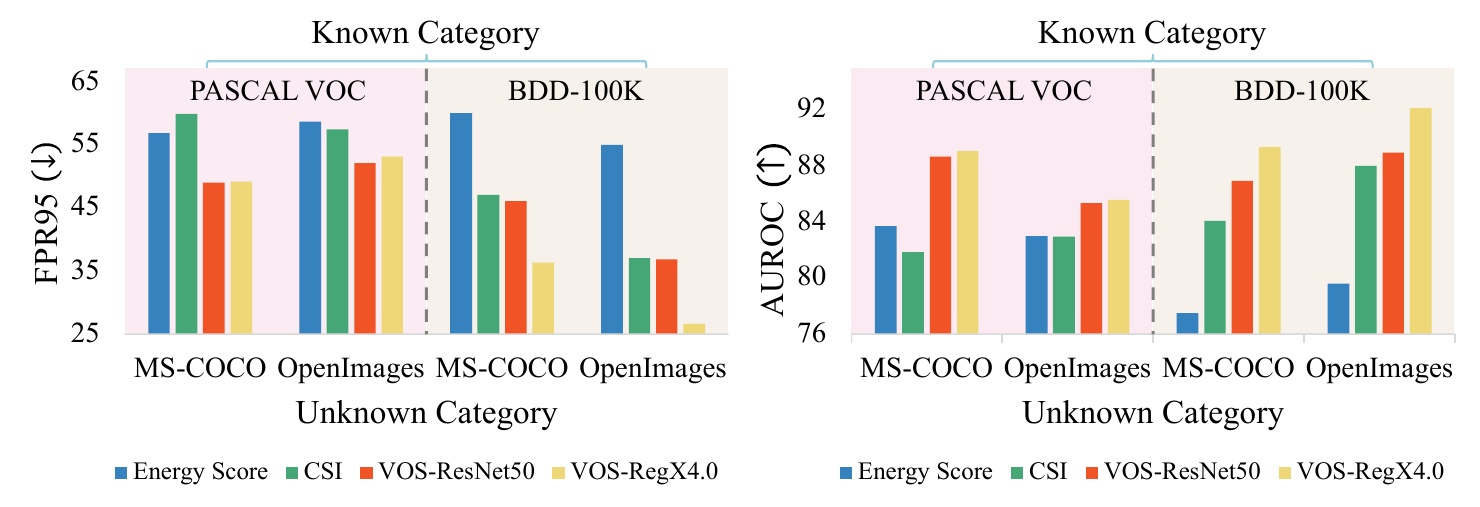}
 \caption{Open-set object detection performance on the VOS benchmark.}
 \label{fig:Open-Set OD VOS}
\end{figure}

% \textbf{Results of Open-Set Object Detection} 
% Tab.~\ref{osod_vos} and Tab.~\ref{osod_opendet} shown the results of OSOD methods. From Tab.~\ref{osod_vos} we find that using different known category data as training has an impact on the recognition of unknown category objects. In addition, when the open scene is more challenging (the unknown category data changes from the MS-COCO dataset to the OpenImages dataset), the ability to detect unknown objects will also decrease. In summary, we have drawn the following conclusions: \ding{182} The selection of known classes and their training scenarios directly impacts the detector's capability to identify unknown classes in open scenarios; \ding{183} As the number and variety of unknown classes in a scene increase, the detector's proficiency in detecting these unknown objects becomes limited.
The VOS~\cite{du2021vos} method establishes a benchmark for open-set object detection (OSOD), providing a standard for evaluating performance in this area. The common backbones used for the benchmark include ResNet-50~\cite{he2016deep} and RegNetX-4.0GF (denote as RegX4.0)~\cite{radosavovic2020designing}. The RegNetX-4.0GF model is part of the RegNet series, which seeks to identify efficient deep network architectures via systematic exploration of network design. The '4.0GF' in its name denotes the model's computational complexity, approximately 4.0 Giga FLOPs (billion floating-point operations per second). The evaluation spans four datasets: \ding{182} The PASCAL VOC~\cite{everingham2010pascal} dataset comprises data across 20 known categories for training. \ding{183} The Berkeley DeepDrive (BDD-100K)~\cite{yu2020bdd100k} dataset includes data for training across 10 known categories. The FPR95 ($\downarrow$) and AUROC ($\uparrow$) of unknown categories (not contain known categories) on \ding{184} the MS-COCO~\cite{lin2014microsoft} dataset and \ding{185} the OpenImages~\cite{kuznetsova2020open} dataset. From Fig.~\ref{fig:Open-Set OD VOS} we can identify: \ding{182} Compared to the Energy Score~\cite{liu2020energy} and CSI~\cite{tack2020csi} methods, the VOS approach, which employs an outlier exposure strategy, significantly enhances the detection of unknown classes. It has demonstrated considerable improvements in both the FPR95 and AUROC metrics. \ding{183} When employing the more powerful RegX4.0 backbone, it enhances the detection of unknown categories, particularly on the MS COCO and OpenImages datasets when the BDD-100K dataset is utilized for training known categories. It is important to note that the choice of backbone significantly influences the detector's proficiency in identifying unknown categories.

\begin{figure}[!t]
 \centering
  \begin{overpic}[width=0.48\textwidth,tics=8]{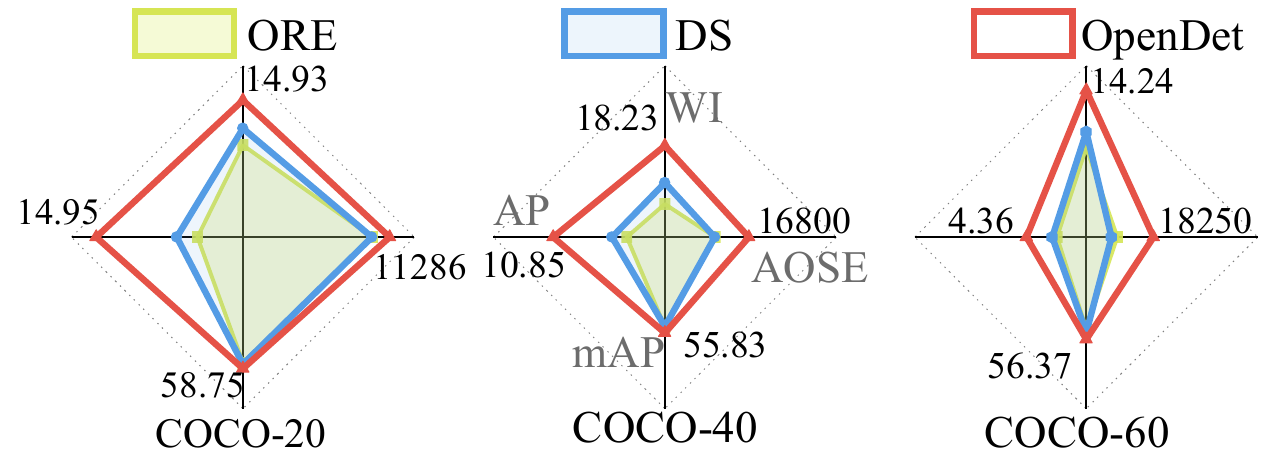}  
        % first line 
        \put (26,32.4){ \cite{joseph2021towards}}
        \put (57,32.4){ \cite{miller2018dropout}}
        \put (97,32.4){ \cite{han2022expanding}}
    \end{overpic}
 \caption{Open-set object detection performance on the OpenDet benchmark.}
 \label{fig:Open-Set OD}
\end{figure}

OpenDet~\cite{han2022expanding} has also developed another benchmark specifically designed for the OSOD task. The backbone used for the benchmark is ResNet-50~\cite{he2016deep}. It uses 20 PASCAL VOC~\cite{everingham2010pascal} categories as known data and $\{20, 40, 60\}$ non-VOC categories from MS COCO~\cite{lin2014microsoft} as unknown categories for testing. Evaluation metrics include Wilderness Impact (WI)~\cite{dhamija2020overlooked} for misclassification rates of unknown versus known categories, Absolute Open-Set Error (AOSE)~\cite{miller2018dropout} for the count of misclassified unknown categories, $mAP$ for detecting known categories, and $AP$ for detecting unknown categories. From Fig.~\ref{fig:Open-Set OD} we can identify: \ding{182} As the number of unknown categories escalates ($20\to 80$), there is a corresponding decline in the detection performance (measured by AP and AOSE) for these categories. \ding{183} Owing to the adoption of the density estimation method for unknown-aware estimation, the OpenDet method demonstrates significant advantages in estimating unknown classes compared to the ORE~\cite{joseph2021towards} and DS~\cite{miller2018dropout} methods.

\input{tables/OOD_Sup}

A benchmark for zero-shot object detection (ZSOD) encompasses the PASCAL VOC (2007+2012)~\cite{everingham2010pascal} benchmark. This benchmark utilizes the training sets from both PASCAL VOC's 2007 and 2012 editions for model training, while the validation set serves to evaluate the model's performance. It comprises a total of 20 categories, with 16 designated as known categories and the remaining 4 as unknown categories. The common backbones used for this benchmark includes RetinaNet~\cite{lin2017focal}, and ResNet-$\{50,101\}$~\cite{he2016deep}. The evaluation encompasses two distinct settings: \ding{182} The Zero-shot detection (ZSD) setting, the detector exclusively identifies unknown categories during testing and evaluates the mAP of known and unknown categories from the validation set separately. \ding{183} The generalized zero-shot detection (GZSD) setting, the detector is tasked with simultaneously identifying both known and unknown categories. It evaluates the mAP of these categories based on the validation set and calculates the harmonic mean (HM) of the performance metrics for both known and unknown categories. 
\input{tables/out_of_category_summary}

According to the results in Tab.~\ref{zsod_voc}, we can identify: \ding{182} Under the ZSD setting, the State-of-the-Art (SOTA) method demonstrates superior inference performance for unknown categories. \ding{183} Under the GZSD setting, which is closer to the open scenario, where simultaneous identification of known and unknown categories is required, the detector performance experiences a decline. It shows that existing methods also have difficulties in distinguishing known categories and unknown categories and still need to be improved.
\input{tables/ooc_sum_2}
Another benchmark for ZSOD encompasses the MS COCO~\cite{lin2014microsoft} benchmark. This benchmark uses two category divisions: 48 known/17 unknown (denotes as 48/17) and 65 known/15 unknown (denotes as 65/15). The common backbones used for this benchmark includes RetinaNet~\cite{lin2017focal}, and ResNet-$\{50,101\}$~\cite{he2016deep}. The evaluation encompasses both ZSD and GZSD settings. It employs two metrics: Recall@100, which focuses on the top 100 targets with the highest confidence, and the average mean precision (mAP), which assesses detection performance across known, unknown, and all categories. Tab.~\ref{coco_zsd} and Tab.~\ref{tab:generalized_zero_shot_detection} detail the settings for ZSD and GZSD settings, we can identify: \ding{182} Model performance varies with different category divisions (48/17 and 65/15), highlighting the significant impact of category division. \ding{183} Synthesis-based method excels in detecting unknown categories, suggesting enhanced model adaptability to unknown classes through feature generation.
\input{tables/ooc_sum_3}

\subsection{Incremental Learning Solution Benchmarks}

Frequently employed datasets for incremental object detection encompass Pascal VOC (2007)~\cite{everingham2007pascal}, a comprehensive object detection dataset comprising 20 distinct classes, and MS-COCO (2014)~\cite{lin2014microsoft}, comprising 80 distinct classes. In order to ensure a fair comparison, this section outlines the experimental setup and benchmarking methodologies for two distinct scenarios: incremental object detection and incremental few-shot object detection. These scenarios are further categorized based on the quantity of data involved.
\subsubsection{Incremental Object Detection}
To assess the incremental capabilities of the model, the experimental setting incorporates two types of incremental approaches: single-step incremental and multi-step incremental.  Hence, it is possible to partition the pertinent training dataset into two or more distinct groups in order to conduct incremental experiments. To ensure comprehensive coverage of test methodologies and facilitate a fair comparison, we employ the single-step incremental approach from Pascal VOC (2007) and the multi-step incremental approach from MS-COCO (2014) as representative benchmarks. The Pascal VOC (2007) dataset can be classified into three distinct configurations based on the alphabetical order of the categories, namely 19 and 1, 15 and 5, and 10 and 10. The MS-COCO dataset is divided into four groups based on the experimental setting in the paper\cite{hao2019end}. In this setting, the detection model acquires a new group of object categories at each incremental stage.

Tab.~\ref{single-step incremental} demonstrates the fundamental performance of the State-of-the-Art (SOTA) approach in the setting of the single-step incremental object detection task across three distinct scenarios. The table showcases the percentiles of the mean average precision (mAP) of the model for classes 1-10 and classes 10-20. Based on the analysis of Tab.~\ref{single-step incremental}, several conclusions can be drawn \ding{182} It is evident that the object detection model experiences a significant forgetting phenomenon, such as Fine-tuning, when incremental techniques are not employed. \ding{183} The incremental results of the model are significantly influenced by the number of incremental categories. However, establishing a direct relationship between the number of incremental categories and the difficulty of the model's learning display is not feasible, as the specific categories included in the incremental process also impact the model's learning outcomes. \ding{184} The combination of incremental methods and replay techniques, exemplified by the ABR algorithm, demonstrates promising advancements in achieving improved results. 4) Single-step incremental methods, such as Joint Training, exhibit a certain upper limit in terms of learning. Notably, the metrics indicate that incremental detection algorithms have nearly reached this upper limit with the single-step incremental setting of Pascal VOC (2007), as demonstrated by the ABR algorithm.
\input{tables/single-step_incremental_object_detection}
% \begin{figure}[!t]
%  \centering
%   \includegraphics[width=0.48\textwidth]{images/incrimental/multi-step_incremental.pdf}
%  \caption{Multi-step Incrimental.}
%  \label{multi-step_incremental}
% \end{figure}

% \input{tables/multi-step incremental object detection}
% Tab.~\ref{multi-step incremental object detection} presents the fundamental performance evaluation of the state-of-the-art (SOTA) technique on the multi-step incremental object detection task. The incremental object groups, denoted as A and B, consist of 20 distinct categories each. The symbol ``+'' denotes the utilization of distinct incremental object groups in this method, distinguishing it from other approaches, as described in the experimental setup of ORE. Upon analyzing Tab.~\ref{multi-step incremental object detection}, several conclusions can be drawn. \ding{182} The learning difficulty of a model in a multi-step incremental setup is higher compared to a single-step incremental setup as the number of iterations of incremental steps increases (\eg, A → B). \ding{183} The difficulty of model learning is influenced to a greater extent by different groups of incremental objects (\eg, MCVD \vs OW-DETR+). \ding{184} An improved design of the detection model not only enhances the algorithm's detection results, but also enhances its incremental effect (\eg, OW-DETR+ algorithm based on transformer structure).

\subsubsection{Incremental Few-shot Object Detection}
\input{tables/incremental_few-shot_object_detection}
In recent times, there has been a growing interest in incremental object detection that takes into account the incremental capability of a model when trained on limited novel class samples. This refers to the model effectively learning new classes using only a small number of data.  The above setting holds greater relevance to real-world scenarios and will greatly facilitate the widespread use of incremental object detection. To comprehensively summarize various methodologies, we primarily provide three conventional configurations, namely 1-shot, 5-shot, and 10-shot,  within a single-step incremental in the MS-COCO (2014) dataset. The model's selection of incremental new categories for the COCO dataset aligns with the 20 categories established in Pascal VOC (2007). It is noteworthy to mention that there exists a certain degree of resemblance between the aforementioned scenarios and few-shot object detection~\cite{wang2020frustratingly, antonelli2022few, wu2021universal}. However, the benchmark conducted in this section primarily focuses on highlighting the model's capacity for continuous learning, rather than its ability to effectively model with limited samples.

Tab.~\ref{incremental few-shot object detection} presents the incremental results of SOTA methods using varying quantities of new category samples. In Tab.~\ref{incremental few-shot object detection}, the ``BASE'' class refers to 60 categories, while the ``NOVEL'' class pertains to 20 categories within the COCO dataset. pon conducting an analysis of Tab.~\ref{incremental few-shot object detection}, it is possible to derive the following conclusions: \ding{182} The efficacy of the detection model's incremental ability experiences a substantial decline as the number of novel samples decreases. This can be observed by comparing the Fine-turning results presented in Tab.~\ref{single-step incremental} and Tab.~\ref{incremental few-shot object detection}. \ding{183} Maintaining the model's scalability in recognizing few-shot samples of new categories while preserving the model's proficiency in identifying base classes poses a significant challenge.  Various methods exhibit poor performance in the detection of new categories. \ding{184} Within a specific range, increasing the number of new category samples can considerably enhance the model's incremental ability.   This can be observed by comparing the results of the novel object detection of different methods at 1 shot and 10 shots. In general, the field of incremental few-shot object detection is still in its early stages of development and requires significant advancements before it can be effectively applicable. Moreover, the task of scaling up incremental detection models from single incremental steps to multiple incremental steps presents even more formidable obstacles~\cite{yin2022sylph}. 

%% file: tables/OOD_Sup.tex
\begin{table}[b!]
\caption{Performance comparison of various object detectors for the out-of-domain challenge.}
\label{solutions for ood}
\renewcommand{\arraystretch}{1.2}
\rowcolors{7}{white}{gray!10}
\setlength{\tabcolsep}{12pt}
\resizebox{.48\textwidth}{!}{

\begin{tabular}{lcccc}
\toprule
\multicolumn{1}{c}{\multirow{3}{*}{Method}} & \multirow{3}{*}{Paper List} & \multicolumn{3}{c}{Datasets} \\
\multicolumn{1}{c}{} &  & VOC-Com & Sim-City & City-FCity \\ 
\cmidrule(lr){3-3} \cmidrule(lr){4-4} \cmidrule(lr){5-5}
\multicolumn{1}{c}{} &  & R101 & R101 & R101 \\ 
\midrule
\multicolumn{2}{c}{source only} & 19.7 & 41.8 & 25.6 \\
\multicolumn{2}{c}{Oracle} & 44.6 & 70.4 & 43.2 \\
DA-Faster~\cite{Chen0SDG18} & CVPR 2018 & 21.2 & - & - \\
SWDA~\cite{SaitoUHS19} & CVPR~2019 & 29.4 & - & - \\
D-adapt~\cite{JiangCWL22} & ICLR 2022 & \textbf{40.5} & \textbf{51.9} & \textbf{42.2} \\
Selective DA~\cite{ZhuPYSL19} & CVPR 2019 & - & - & - \\
CRDA~\cite{XuZJW20} & CVPR 2020 & - & - & - \\
MCAR~\cite{ZhaoGSY20} & ECCV 2020 & 33.5 & - & - \\
SAPNet~\cite{LiDZWLWZ20} & ECCV 2020 & - & - & - \\
MeGA-CDA~\cite{VSGOSP21} & CVPR 2021 & - & - & - \\
MAF~\cite{HeZ19} & ICCV 2019 & - & - & - \\
MEAA~\cite{NguyenTS20} & MM 2020 & - & - & - \\
ATF~\cite{HeZ20} & ECCV 2020 & - & - & - \\
HTCN~\cite{ChenZD0D20} & CVPR 2020 & - & - & - \\
CFFA~\cite{Zheng0LW20} & CVPR 2020 & - & - & - \\
RPN-PA~\cite{ZhangWM21} & CVPR 2021 & - & - & - \\
DBGL~\cite{ChenLZ0DY21} & ICCV 2021 & 29.7 & - & - \\
CDN~\cite{SuWZTCQW20} & ECCV 2020 & - & - & - \\
ICCR-VDD~\cite{WuLHZ021} & ICCV 2021 & - & - & - \\
PFD~\cite{WuHZY22} & TPAMI 2021 & - & - & 38.6 \\
ICCM~\cite{HouZFL21} & CVPR 2021 & 37.1 & - & - \\
SIGMA~\cite{LiLY22} & CVPR 2022 & - & - & - \\
DD-DML~\cite{KimJKCK19} & CVPR 2019 & 34.5 & - & - \\
ST+C+RPL~\cite{RodriguezM19} & BMCV 2019 & 39.4 & - & - \\
AFAN~\cite{WangLS21} & TIP 2021 & - & - & - \\
TSA-DA~\cite{YunHLKK21} & RAL 2021 & - & - & - \\
SC-UDA~\cite{YuWCKSYLLS022} & WACV 2022 & - & - & - \\
PDA~\cite{hsu2020progressive} & WACV 2020 & - & - & - \\
CST-DA~\cite{ZhaoLXL20} & ECCV 2020 & - & - & - \\
CDG~\cite{LiH0Z21} & AAAI 2021 & - & - & - \\
UMT~\cite{Deng0CD21} & CVPR 2021 & - & - & - \\
TDD~\cite{HeWWWLLGWQ22} & CVPR 2022 & - & - & - \\
AT~\cite{LiDMLCWHKV22} & CVPR 2022 & - & - & - \\
KTNet~\cite{TianZWXP21} & ICCV 2021 & - & - & - \\
DC~\cite{LiuZWJY22} & TCSVT 2022 & 38.7 & - & - \\
TIA~\cite{Zhao022} & CVPR 2022 & - & - & - \\
\bottomrule

\end{tabular}

}
\end{table}

%% file: tables/out_of_category_summary.tex
\begin{table}[!t]
    \renewcommand{\arraystretch}{1.2}
    \rowcolors{4}{gray!10}{white}
    \centering
    \makeatletter\def\@captype{table}\makeatother\caption{Zero-shot detection and generalized zero-shot detection performance on Pascal VOC dataset. \ddag~denotes the method using image synthesis, HM denotes the harmonic mean of known and unknown categories.}
    \resizebox{0.48 \textwidth}{!}{
    \begin{tabular}{lccccc}
    \toprule
    \multicolumn{1}{c}{\multirow{2}{*}{Method}} & \multicolumn{2}{c}{ZSD} & \multicolumn{3}{c}{GZSD}      \\ \cmidrule(r){2-3} \cmidrule(l){4-6}
                            & Known      & Unknown  & Known & Unknown & HM \\ \midrule
    TL-ZSD~\cite{rahman2019transductive}     & - & \textbf{66.6}  & - & - & -  \\
    BLC~\cite{zheng2020background} & 75.1  & 55.2  & 58.2  & 22.9  & 32.9  \\
    SAN~\cite{rahman2020zero} & 69.6  & 59.1  & 48.0  & 37.0  & 41.8 \\
    PL~\cite{rahman2020improved} & 63.5  & 62.1  & - & - & - \\
    \ddag~Huang \textit{\etal}~\cite{huang2022robust} & - & 65.6  & 47.1  & \textbf{49.1}  & 48.1 \\
    ContrastZSD~\cite{yan2022semantics} & \textbf{76.7}      & 65.7        & \textbf{63.2} & 46.5   & \textbf{53.6}          \\ \bottomrule
    \end{tabular}}
    \label{zsod_voc}
\end{table}

%% file: tables/ooc_sum_2.tex
\begin{table}[!b]
    \renewcommand{\arraystretch}{1.2}
    \rowcolors{4}{gray!10}{white}
    \centering \makeatletter\def\@captype{table}\makeatother\caption{Zero-shot detection performance on MS COCO dataset. \ddag~denotes the method using image synthesis.}
    \resizebox{0.5 \textwidth}{!}{
    \begin{tabular}{lccccc}
    \toprule
    \multicolumn{1}{c}{\multirow{2}{*}{Method}} & \multirow{2}{*}{Split} & \multicolumn{3}{c}{Recall@100} & mAP     \\ \cmidrule(r){3-5} \cmidrule(l){6-6} 
                &   & IoU=0.4  & IoU=0.5  & IoU=0.6  & IoU=0.5 \\ \midrule
    SB~\cite{bansal2018zero} & 48/17 & 34.5    & 24.4    & 12.6  & 0.7 \\
    DSES~\cite{bansal2018zero} & 48/17   & 40.2  & 27.2  & 13.6  & 0.5 \\
    Li \textit{\etal}~\cite{li2019zero} & 48/17 & 45.5  & 34.3  & 18.1  & - \\
    GTNet~\cite{zhao2020gtnet} & 48/17    & 46.2  & 43.4  & 34.9  & - \\
    BLC~\cite{zheng2020background} & 48/17   & 51.3  & 48.8  & 45.0  & 10.6 \\
    \ddag DELO~\cite{zhu2020don} & 48/17 & - & 33.5 & -   & 7.6\\
    SPGP-Occurrence~\cite{yan2020semantics} & 48/17 & 45.6  & 35.4  & 19.4  & -  \\
    SAN~\cite{rahman2020zero} & 48/17   & - & 12.3 & - & 5.1 \\
    VL-SZSD~\cite{zheng2021visual} & 48/17 & 52.2  & 49.6  & 45.2  & 11.3\\
    PL~\cite{rahman2020improved} & 48/17   & -  & 43.6  & -  & 10.0 \\
    Nie \textit{\etal}~\cite{nie2022node} & 48/17 & \textbf{58.5} & \textbf{55.0}   & \textbf{50.3}  & 11.4 \\
    \ddag Huang \textit{\etal}~\cite{huang2022robust} & 48/17   & 58.1  & 53.5  & 47.9  & \textbf{13.4} \\
    ContrastZSD~\cite{yan2022semantics}  & 48/17                  & 56.1     & 52.4     & 47.2     & 12.5    \\ \midrule
    TL-ZSD~\cite{rahman2019transductive} & 65/15 & - & 48.2  & - & 14.6 \\
    BLC~\cite{zheng2020background} & 65/15   & 57.2  & 54.7  & 51.2  & 14.7 \\
    VL-SZSD~\cite{zheng2021visual} & 65/15 & 58.9  & 56.3  & 51.6  & 14.0  \\
    PL~\cite{rahman2020improved} & 65/15   & 39.8  & 37.7  & 35.5  & 12.4 \\
    Nie \textit{\etal}~\cite{nie2022node} & 65/15 & \textbf{65.3} & \textbf{62.7}   & \textbf{58.3}  & 14.9 \\
    \ddag Huang \textit{\etal}~\cite{huang2022robust} & 65/15 & \textbf{65.3}  & 62.3  & 55.9  & \textbf{19.8} \\
    ContrastZSD~\cite{yan2022semantics} & 65/15                  & 62.3     & 59.5     & 55.1     & 18.6    \\ 
    \bottomrule
    \end{tabular}}
    \label{coco_zsd}
\end{table}

%% file: tables/ooc_sum_3.tex
\begin{table}[!t]
\centering
\makeatletter\def\@captype{table}\makeatother\caption{Generalized zero-shot detection performance on MS COCO dataset. \ddag~denotes the method using image synthesis, HM denotes the harmonic mean of known and unknown categories.}
\resizebox{\columnwidth}{!}{
\rowcolors{5}{gray!10}{white}
\begin{tabular}{lcccccccc}
\toprule
\multicolumn{1}{c}{\multirow{2}{*}[-0.35em]{Method}} & \multirow{2}{*}[-0.35em]{Split} & \multicolumn{3}{c}{Recall@100} & \multicolumn{3}{c}{mAP} \\ 
\cmidrule(l){3-5} \cmidrule(l){6-8}
 &  & Known & Unknown & HM & Known & Unknown & HM \\ 
\midrule
GTNet~\cite{zhao2020gtnet} & 48/17 & 42.5 & 30.4 & - & - & - & - \\
BLC~\cite{zheng2020background} & 48/17 & 57.6 & 46.4 & 51.4 & 42.1 & 4.5 & 8.2 \\
SAN~\cite{rahman2020zero} & 48/17 & 20.4 & 12.4 & 15.5 & 13.9 & 2.6 & 4.3 \\
VL-SZSD~\cite{zheng2021visual} & 48/17 & \textbf{71.0} & 48.3 & 57.5 & \textbf{47.3} & 4.6 & 8.5 \\
PL~\cite{rahman2020improved} & 48/17 & 38.2 & 26.3 & 31.2 & 35.9 & 4.1 & 7.4 \\
Nie \textit{\etal}~\cite{nie2022node} & 48/17 & 66.7 & 54.5 & \textbf{60.0} & 43.9 & 4.7 & 8.5 \\
\ddag Huang \textit{\etal}~\cite{huang2022robust} & 48/17 & 59.7 & \textbf{58.8} & 59.2 & 42.3 & \textbf{13.4} & \textbf{20.4} \\
ContrastZSD~\cite{yan2022semantics} & 48/17 & 65.7 & 52.4 & 58.3 & 45.1 & 6.3 & 11.1 \\ 
\midrule
TL-ZSD~\cite{rahman2019transductive} & 65/15 & 54.1 & 37.2 & 44.1 & 28.8 & 14.1 & 18.9 \\
BLC~\cite{zheng2020background} & 65/15 & 56.4 & 51.7 & 53.9 & 36.0 & 13.1 & 19.2 \\
VL-SZSD~\cite{zheng2021visual} & 65/15 & \textbf{68.9} & 54.3 & 60.8 & 39.5 & 13.2 & 19.8 \\
PL~\cite{rahman2020improved} & 65/15 & 36.4 & 37.2 & 36.8 & 34.1 & 12.4 & 18.2 \\
Nie \textit{\etal}~\cite{nie2022node} & 65/15 & 65.3 & 60.5 & \textbf{62.8} & 38.1 & 13.9 & 20.4 \\
\ddag Huang \textit{\etal}~\cite{huang2022robust} & 65/15 & 58.6 & \textbf{61.8} & 60.2 & 37.4 & \textbf{19.8} & \textbf{26.0} \\
ContrastZSD~\cite{yan2022semantics} & 65/15 & 62.9 & 58.6 & 60.7 & \textbf{40.2} & 16.5 & 23.4 \\
\bottomrule
    \end{tabular}}
    \label{tab:generalized_zero_shot_detection}
\end{table}

%% file: tables/single-step_incremental_object_detection.tex
\begin{table}[]
\caption{Results of single-step incremental comparisons on Pascal VOC (2007).}
\label{single-step incremental}
\large
\resizebox{\columnwidth}{!}{%
\setlength{\tabcolsep}{8pt}
\rowcolors{6}{white}{gray!10}
\begin{tabular}{lccccccccccc}
\toprule
\multicolumn{1}{c}{\multirow{2}{*}[-0.25em]{Method}} & \multicolumn{3}{c}{19-1} & \multicolumn{3}{c}{15-5} & \multicolumn{3}{c}{10-10} \\  \cmidrule(l){2-4} \cmidrule(l){5-7} \cmidrule(l){8-10} 
\multicolumn{1}{c}{} & 1-19 & 20 & 1-20 & 1-15 & 16-20 & 1-20 & 1-10 & 11-20 & 1-20 \\ 
\midrule
Joint Training & 75.3 & 73.6 & 75.2 & 76.8 & 70.4 & 75.2 & 74.7 & 75.7 & 75.2 \\
Fine Tuning & 12.0 & 62.8 & 14.5 & 14.2 & 59.2 & 25.4 & 9.5 & 62.5 & 36.0 \\
ER~\cite{shieh2020continual} & 67.8 & 44.7 & 66.7 & - & - & - & 55.4 & 67.3 & 61.4 \\
ORE~\cite{joseph2021towards} & 69.4 & 60.1 & 68.9 & 71.8 & 58.7~ & 68.5 & 60.4 & 68.8 & 64.6 \\
ABR~\cite{yuyang2023augmented} & 71.0 & 69.7 & 70.9 & 73.0~ & 65.1 & 71.0 & 71.2 & 72.8 & 72.0 \\
ILOD~\cite{peng2020faster} & 68.5 & 62.7 & 68.3 & 68.3 & 58.4 & 65.9 & 63.2 & 63.1 & 63.2 \\
RILOD~\cite{li2019rilod} & 66.3 & 40.4 & 65.0 & - & - & - & 67.5 & 68.3 & 67.9 \\
Faster ILOD~\cite{peng2020faster} & 68.9 & 61.1 & 68.5 & 71.6 & 56.9 & 67.9 & 69.8 & 54.5 & 62.1 \\
PPAS~\cite{zhou2020lifelong} & 70.5 & 53.0 & 69.2 & - & - & - & 63.5 & 60.0~ & 61.8 \\
IncDet~\cite{liu2020incdet} & - & - & - & 72.7 & 63.5 & 70.4 & 69.7 & 71.8 & 70.8 \\
Dong \etal~\cite{dong2021bridging} & 73.5 & 65.8 & 73.1 & 72.7 & 58.4 & 69.1 & 69.2 & 68.3 & 68.7 \\
Joseph \etal~\cite{joseph2021incremental} & 70.9 & 57.6~ & 70.2 & 71.7 & 55.9~ & 67.8 & 68.4 & 64.3~ & 66.3 \\
OW-DETR~\cite{gupta2022ow} & 70.2 & 62.0 & 69.8 & 72.2 & 59.8 & 69.1 & 63.5 & ~67.9 & 65.7 \\
ROSETTA~\cite{yang2022continual} & 69.7 & 68.3 & 69.6 & 71.7 & 61.6 & 69.2 & 67.6 & 66.0 & 66.8 \\
MVCD~\cite{yang2022multi} & 70.2 & 60.6 & ~69.7 & 69.4~ & 57.9 & 66.5 & 66.2~~ & 66.0 & 66.1 \\
MMA~\cite{cermelli2022modeling} & 71.1 & 63.4 & 70.7 & 73.0~ & 60.5 & 69.9 & 69.3 & 63.9 & 66.6 \\
\bottomrule

\end{tabular}%
}
\end{table}

%% file: tables/incremental_few-shot_object_detection.tex
\begin{table}[b]
\caption{Results of incremental few-shot comparisons on MS-COCO (2014).}
\label{incremental few-shot object detection}
\resizebox{\columnwidth}{!}{%
\setlength{\tabcolsep}{6pt}
\rowcolors{6}{white}{gray!10}
\begin{tabular}{lccccccccc} 
\toprule
\multicolumn{1}{c}{\multirow{2}{*}{Method}} & \multicolumn{3}{c}{1 shot} & \multicolumn{3}{c}{5 shots} & \multicolumn{3}{c}{10 shots} \\ 
\cmidrule(l){2-4}\cmidrule(l){5-7}\cmidrule(l){8-10}
\multicolumn{1}{c}{} & 1-60 & 61-80 & 1-80 & 1-60 & 61-80 & 1-80 &1-60 & 61-80 & 1-80 \\ 
\midrule
Fine Tuning & 1.1 & 0.0 & 0.8 & 2.6 & 0.2 & 2.0 & 2.8 & 0.6 & 2.3 \\
Kang \etal~\cite{kang2019few} & 2.5 & 0.1 & 1.9 & 3.3 & 0.8 & 2.7 & 3.7 & 1.5 & 3.2 \\
ONCE~\cite{perez2020incremental} & 17.9 & 0.7 & 13.6 & 17.9 & 1.0 & 13.7 & 17.9 & 1.2 & 13.7 \\
IMTFA~\cite{ganea2021incremental} & 27.8 & 3.2 & 21.7 & 24.1 & 6.0 & 19.6 & 23.3 & 7.0 & 19.2 \\
Teo \etal~\cite{li2021towards} & - & - & - & - & - & - & 28.1 & 8.5 & 23.2 \\
DualFusion~\cite{tambwekar2021few} & - & - & - & - & - & - & 10.8 & 9.9 & 10.6 \\
LEAST~\cite{li2021class} & 29.5 & 4.2 & 23.2 & 31.3 & 9.3 & 25.8 & 31.3 & 12.8 & 26.7 \\
Inc-DETR~\cite{dong2023incremental} & 29.4 & 1.9 & 22.5 & 30.5 & 8.3 & 25.0 & 27.3 & 14.4 & 24.1 \\
HDA~\cite{she2022fast} & 39.2 & 3.0 & 30.2 & 39.2 & 7.1 & 31.2 & 39.2 & 9.1 & 31.7 \\
iFSOD~\cite{cheng2021meta} & 30.7 & 1.5 & 23.4 & 33.3 & 2.5 & 25.6 & 31.4 & 2.6 & 24.2 \\
iFS-RCNN~\cite{nguyen2022ifs} & 40.1 & 4.5 & 31.2 & 40.1 & 9.9 & 32.6 & 40.1 & 12.6 & 33.2 \\
\bottomrule
\end{tabular}%
}
\end{table}